\pdfoutput=1
\documentclass{article}

\usepackage{microtype}

\usepackage{multirow}

\usepackage[final]{neurips_2024}
\usepackage{multirow}

\usepackage{caption}
\usepackage{subcaption}
\usepackage[utf8]{inputenc} %
\usepackage[T1]{fontenc}    %
\usepackage{lmodern}

\usepackage{url}            %
\usepackage{booktabs}       %
\usepackage{amsfonts}       %
\usepackage{nicefrac}       %
\usepackage{
microtype}      %
\usepackage{amsmath}
\usepackage{amssymb}
\usepackage{mathtools}
\usepackage{slashed}
\usepackage{braket}
\usepackage{enumitem}
\usepackage[svgnames]{xcolor}
\usepackage[colorlinks,citecolor=DarkGreen,linkcolor=FireBrick,urlcolor=FireBrick,linktocpage=true,unicode]{hyperref}

\allowdisplaybreaks
 \usepackage[T1]{fontenc}
\usepackage{graphicx}
\usepackage{tikz}
\usepackage{tikz-cd}
\usetikzlibrary{positioning, shapes.geometric, calc, arrows.meta}
\usetikzlibrary{decorations.pathreplacing}
\usepackage{times}
\usepackage{courier}
\usepackage{bm}
\usepackage{physics}
\usepackage{xcolor}
\usepackage{natbib}
\usepackage{mdframed}
\usepackage{nicefrac}
\usepackage{booktabs}
\usepackage{lipsum}
\usepackage{titlesec}
\usepackage{wrapfig,lipsum,booktabs}
\usepackage{authblk}
\usepackage{blindtext}
\usepackage{algorithm}
\usepackage{algorithmicx}
\usepackage{algpseudocode}
\usepackage{titletoc}
\usepackage{todonotes}
\usepackage{authblk}
\usepackage{titletoc}
\usepackage{setspace}
\usepackage{cancel}
\usepackage{dsfont} %
\usepackage{multicol}
\newcommand{\one}{\mathds{1}}

\newcommand{\RR}{\mathbb{R}}
\newcommand{\EE}{\mathbb{E}}

\newcommand{\vzero}{\mathbf{0}}

\newcommand{\wt}{\tilde}

\usepackage[nameinlink,capitalize,noabbrev]{cleveref}

\usepackage{CJK}

\usepackage{array}
\usepackage{makecell}

\def\half{{\frac{1}{2}}}

\newcommand{\bea}{\begin{eqnarray}}
\newcommand{\eea}{\end{eqnarray}}

\newcommand{\cL}{\mathcal{L}}

\newcommand{\cT}{\mathcal{T}}
\newcommand{\cO}{\mathcal{O}}

\usepackage{slashed}

\def\({\left(}
\def\){\right)}
\def\[{\left[}
\def\]{\right]}

\usepackage{dashbox}
\usepackage{xcolor}
\usepackage{colortbl}
\usepackage{arydshln}
\definecolor{lightyellow}{rgb}{1.0, 0.95, 0.7}
\definecolor{Blue}{rgb}{0, 0, 0.8}
\definecolor{blue}{rgb}{0,0,1}
\definecolor{darkgreen}{rgb}{0,0.40,0}
\definecolor{firebrick}{rgb}{0.698,0.133,0.133}

\definecolor{colorA}{rgb}{1,0,0}
\definecolor{colorB}{rgb}{0,0.3,1}
\definecolor{colorC}{rgb}{0.9,0.8,0.2}
\definecolor{colorD}{rgb}{0,0.65,0}
\definecolor{lesslightgray}{rgb}{0.5,0.5,0.5}
\definecolor{light-gray}{gray}{0.95}

\let\tilde\widetilde
\let\hat\widehat

\newcommand{\calC}{\mathcal{C}}

\newcommand{\calF}{\mathcal{F}}
\newcommand{\calG}{\mathcal{G}}

\newcommand{\calL}{\mathcal{L}}

\newcommand{\calN}{\mathcal{N}}
\newcommand{\calO}{\mathcal{O}}

\newcommand{\calS}{\mathcal{S}}
\newcommand{\calT}{\mathcal{T}}

\newcommand{\bX}{\mathbf{X}}

\newcommand{\diag}{\mathop{\rm{diag}}}

\newcommand{\argmin}{\mathop{\mathrm{argmin}}}  
\newcommand{\argmax}{\mathop{\mathrm{argmax}}}

\newcommand{\Softmax}{\mathop{\rm{Softmax}}}
\newcommand{\Hardmax}{\mathop{\rm{Hardmax}}}

\renewcommand{\d}{\mathrm{d}}

\DeclareMathOperator{\vect}{vec}
\DeclareMathOperator{\A}{\mathsf{A}}

\newcommand{\sT}{ \mathsf{T} }

\newcommand{\poly}{\mathrm{poly}}

\def\R{\mathbb{R}}

\let\cite\citep 

\usepackage{amsthm}
\makeatletter
\def\th@remark{%
  \thm@headfont{\bfseries}%
  \normalfont %
  \thm@preskip\topsep \divide\thm@preskip\tw@
  \thm@postskip\thm@preskip
}
\makeatother
\usepackage[many]{tcolorbox}
\theoremstyle{definition}
\newtheorem{theorem}{Theorem}[section]
\tcolorboxenvironment{theorem}{
  breakable,
  colback=black!10,
  colframe=white,%
  width=\dimexpr\linewidth+10pt\relax,%
  enlarge left by=-5pt,%
  enlarge right by=-5pt,%
  boxsep=5pt,%
  boxrule=0pt,
  left=0pt,right=0pt,top=0pt,bottom=0pt,
  sharp corners,
  before skip=\topsep,
  after skip=\topsep
}
\newtheorem{lemma}{Lemma}[section]
\tcolorboxenvironment{lemma}{
  breakable,
  colback=black!10,
  colframe=white,%
  width=\dimexpr\linewidth+10pt\relax,%
  enlarge left by=-5pt,%
  enlarge right by=-5pt,%
  boxsep=5pt,%
  boxrule=0pt,
  left=0pt,right=0pt,top=0pt,bottom=0pt,
  sharp corners,
  before skip=\topsep,
  after skip=\topsep
}
\newtheorem{corollary}{Corollary}[theorem]
\tcolorboxenvironment{corollary}{
  breakable,
  colback=black!10,
  colframe=white,%
  width=\dimexpr\linewidth+10pt\relax,%
  enlarge left by=-5pt,%
  enlarge right by=-5pt,%
  boxsep=5pt,%
  boxrule=0pt,
  left=0pt,right=0pt,top=0pt,bottom=0pt,
  sharp corners,
  before skip=\topsep,
  after skip=\topsep
}
\newtheorem{proposition}{Proposition}[section]
\tcolorboxenvironment{proposition}{
  breakable,
  colback=black!10,
  colframe=white,%
  width=\dimexpr\linewidth+10pt\relax,%
  enlarge left by=-5pt,%
  enlarge right by=-5pt,%
  boxsep=5pt,%
  boxrule=0pt,
  left=0pt,right=0pt,top=0pt,bottom=0pt,
  sharp corners,
  before skip=\topsep,
  after skip=\topsep
}
\theoremstyle{definition}
\newtheorem{definition}{Definition}[section]
\tcolorboxenvironment{definition}{
  breakable,
  colback=black!10,
  colframe=white,%
  width=\dimexpr\linewidth+10pt\relax,%
  enlarge left by=-5pt,%
  enlarge right by=-5pt,%
  boxsep=5pt,%
  boxrule=0pt,
  left=0pt,right=0pt,top=0pt,bottom=0pt,
  sharp corners,
  before skip=\topsep,
  after skip=\topsep
}
\newtheorem{remark}{Remark}[section]
\newtheorem{assumption}{Assumption}[section]
\tcolorboxenvironment{assumption}{
  breakable,
  colback=black!10,
  colframe=white,%
  width=\dimexpr\linewidth+10pt\relax,%
  enlarge left by=-5pt,%
  enlarge right by=-5pt,%
  boxsep=5pt,%
  boxrule=0pt,
  left=0pt,right=0pt,top=0pt,bottom=0pt,
  sharp corners,
  before skip=\topsep,
  after skip=\topsep
}
\newtheorem{problem}{Problem}
\tcolorboxenvironment{problem}{
  breakable,
  colback=black!10,
  colframe=white,%
  width=\dimexpr\linewidth+10pt\relax,%
  enlarge left by=-5pt,%
  enlarge right by=-5pt,%
  boxsep=5pt,%
  boxrule=0pt,
  left=0pt,right=0pt,top=0pt,bottom=0pt,
  sharp corners,
  before skip=\topsep,
  after skip=\topsep
}
\newtheorem{question}{Question}
\tcolorboxenvironment{question}{
  breakable,
  colback=black!10,
  colframe=white,%
  width=\dimexpr\linewidth+10pt\relax,%
  enlarge left by=-5pt,%
  enlarge right by=-5pt,%
  boxsep=5pt,%
  boxrule=0pt,
  left=0pt,right=0pt,top=0pt,bottom=0pt,
  sharp corners,
  before skip=\topsep,
  after skip=\topsep
}
\crefname{question}{Question}{Questions}

\crefname{theorem}{Theorem}{Theorems}
\crefname{proposition}{Proposition}{Propositions}
\crefname{lemma}{Lemma}{Lemmas}
\crefname{corollary}{Corollary}{Corollaries}
\crefname{definition}{Definition}{Definitions}
\crefname{assumption}{Assumption}{Assumptions}
\crefname{remark}{Remark}{Remarks}
\crefname{problem}{Problem}{Problems}
\crefname{property}{Property}{property}
\newtheorem{hypothesis}{Hypothesis}
\tcolorboxenvironment{hypothesis}{
  breakable,
  colback=black!10,
  colframe=white,%
  width=\dimexpr\linewidth+10pt\relax,%
  enlarge left by=-5pt,%
  enlarge right by=-5pt,%
  boxsep=5pt,%
  boxrule=0pt,
  left=0pt,right=0pt,top=0pt,bottom=0pt,
  sharp corners,
  before skip=\topsep,
  after skip=\topsep
}
\crefname{hypothesis}{Hypothesis}{Hypothesises}

\numberwithin{equation}{section}
\numberwithin{theorem}{section}
\numberwithin{proposition}{section}
\numberwithin{definition}{section}
\numberwithin{lemma}{section}
\numberwithin{assumption}{section}
\numberwithin{remark}{section}

\usepackage{lipsum}

\newcommand*{\annot}[1]{\tag*{\footnotesize{\textcolor{black!50}{\big(#1\big)}}}}

\makeatletter
\let\save@mathaccent\mathaccent
\newcommand*\if@single[3]{%
    \setbox0\hbox{${\mathaccent"0362{#1}}^H$}%
    \setbox2\hbox{${\mathaccent"0362{\kern0pt#1}}^H$}%
    \ifdim\ht0=\ht2 #3\else #2\fi
}
\newcommand*\rel@kern[1]{\kern#1\dimexpr\macc@kerna}
\newcommand*\widebar[1]{\@ifnextchar^{{\wide@bar{#1}{0}}}{\wide@bar{#1}{1}}}
\newcommand*\wide@bar[2]{\if@single{#1}{\wide@bar@{#1}{#2}{1}}{\wide@bar@{#1}{#2}{2}}}
\newcommand*\wide@bar@[3]{%
    \begingroup
    \def\mathaccent##1##2{%
        \let\mathaccent\save@mathaccent
        \if#32 \let\macc@nucleus\first@char \fi
        \setbox\z@\hbox{$\macc@style{\macc@nucleus}_{}$}%
        \setbox\tw@\hbox{$\macc@style{\macc@nucleus}{}_{}$}%
        \dimen@\wd\tw@
        \advance\dimen@-\wd\z@
        \divide\dimen@ 3
        \@tempdima\wd\tw@
        \advance\@tempdima-\scriptspace
        \divide\@tempdima 10
        \advance\dimen@-\@tempdima
        \ifdim\dimen@>\z@ \dimen@0pt\fi
        \rel@kern{0.6}\kern-\dimen@
        \if#31
        \overline{\rel@kern{-0.6}\kern\dimen@\macc@nucleus\rel@kern{0.4}\kern\dimen@}%
        \advance\dimen@0.4\dimexpr\macc@kerna
        \let\final@kern#2%
        \ifdim\dimen@<\z@ \let\final@kern1\fi
        \if\final@kern1 \kern-\dimen@\fi
        \else
        \overline{\rel@kern{-0.6}\kern\dimen@#1}%
        \fi
    }%
    \macc@depth\@ne
    \let\math@bgroup\@empty \let\math@egroup\macc@set@skewchar
    \mathsurround\z@ \frozen@everymath{\mathgroup\macc@group\relax}%
    \macc@set@skewchar\relax
    \let\mathaccentV\macc@nested@a
    \if#31
    \macc@nested@a\relax111{#1}%
    \else
    \def\gobble@till@marker##1\endmarker{}%
    \futurelet\first@char\gobble@till@marker#1\endmarker
    \ifcat\noexpand\first@char A\else
    \def\first@char{}%
    \fi
    \macc@nested@a\relax111{\first@char}%
    \fi
    \endgroup
    }
\makeatother
\let\bar\widebar

\makeatletter
\newcommand*{\redefinesymbolwitharg}[1]{%
  \expandafter\let\csname ltx#1\expandafter\endcsname\csname #1\endcsname
  \@namedef{#1}{\@ifnextchar{^}{\@nameuse{#1@}}{\@nameuse{#1@}^{}}}%
  \expandafter\def\csname #1@\endcsname^##1##2{%
     \csname ltx#1\endcsname\ifx!##1!\else^{##1}\fi\mathopen{}\mathclose\bgroup\left(##2\aftergroup\egroup\right)
     }%
}
\makeatother
\redefinesymbolwitharg{sin}
\redefinesymbolwitharg{cos}

\titlespacing\section{0pt}{4pt plus 4pt minus 2pt}{-2pt plus 2pt minus 2pt}
\titlespacing\subsection{0pt}{2pt plus 4pt minus 2pt}{-2pt plus 2pt minus 2pt}
\titlespacing\subsubsection{0pt}{2pt plus 4pt minus 2pt}{-2pt plus 2pt minus 2pt}
\titlespacing\paragraph{0pt}{2pt plus 4pt minus 2pt}{1em}

\usepackage{titletoc}

\def\Snospace~{\S{}}

\newcommand{\sref}[2]{\hyperref[#2]{#1 \ref{#2}}}

\title{On Statistical Rates and  Provably Efficient Criteria\\ of Latent Diffusion Transformers (DiTs)

}

\author{
    {\bf
   Jerry Yao-Chieh Hu\thanks{Equal contribution.
   Future updates are on \href{https://arxiv.org/abs/2407.01079}{arXiv}.
   }$^{\:\;\dagger\ddag}$\quad 
   Weimin Wu$^{*\dagger\ddag}$ \quad 
    Zhao Song$^\S$ \quad
   Han Liu$^{\dagger\ddag\natural}$}
   \\ \vspace{0.5em}
{\small
 $^\dagger$Center for Foundation Models and Generative AI, $^\ddag$Department of Computer Science, $^\natural$Department of Statistics and Data Science, Northwestern University, Evanston, IL 60208, USA\\ 
 $^\S$Simons Institute for the Theory of Computing, UC Berkeley, Berkeley, CA 94720, USA
}\\
\vspace{.5em}
   {\footnotesize
   \texttt{\{\href{mailto:jhu@u.northwestern.edu}{jhu},\href{mailto:wwm@u.northwestern.edu}{wwm}\}@u.northwestern.edu;
   }\\
   \texttt{\href{mailto:magic.linuxkde@gmail.com}{magic.linuxkde@gmail.com}; \href{mailto:hanliu@northwestern.edu}{hanliu@northwestern.edu}}}
}

\setlist[itemize]{leftmargin=1em, before=\vspace{-0.3em}, after=\vspace{-0.3em}, itemsep=0.1em}
\setlist[enumerate]{leftmargin=1.4em, 
before=\vspace{-0.3em}, after=\vspace{-0.3em}, 
itemsep=0.1em}

\begin{document}
\setlength{\abovedisplayskip}{3pt}
\setlength{\abovedisplayshortskip}{3pt}
\setlength{\belowdisplayskip}{3pt}
\setlength{\belowdisplayshortskip}{3pt}

\maketitle
\begin{abstract}
We investigate the statistical and computational limits of latent \underline{Di}ffusion \underline{T}ransformers (DiTs) under the low-dimensional linear latent space assumption.
Statistically, we study the universal approximation and sample complexity of the DiTs score function, as well as the distribution recovery property of the initial data. 
Specifically, under mild data assumptions, we derive an approximation error bound for the score network of latent DiTs, which is sub-linear in the latent space dimension. 
Additionally, we derive the corresponding sample complexity bound and show that the data distribution generated from the estimated score function converges toward a proximate area of the original one. 
Computationally, we characterize the hardness of both forward inference and backward computation of latent DiTs, assuming the \underline{S}trong \underline{E}xponential \underline{T}ime \underline{H}ypothesis (SETH). 
For forward inference, we identify efficient criteria for all possible latent DiTs inference algorithms and showcase our theory by pushing the efficiency toward almost-linear time inference. 
For backward computation, we leverage the low-rank structure within the gradient computation of DiTs training for possible algorithmic speedup. 
Specifically, we show that such speedup achieves almost-linear time latent DiTs training by casting the DiTs gradient as a series of chained low-rank approximations with bounded error.
Under the low-dimensional assumption, we show that the statistical rates and the computational efficiency are all dominated by the dimension of the subspace, suggesting that latent DiTs have the potential to bypass the challenges associated with the high dimensionality of initial data.

\end{abstract}

\section{Introduction}
\label{sec:intro}
We investigate the statistical and computational limits of latent diffusion transformers (DiTs), assuming the data is supported on an unknown low-dimensional linear subspace. 
This analysis is not only practical but also timely. 
On one hand, DiTs have demonstrated revolutionary success in generative AI and digital creation by using Transformers as score networks \cite{esser2024scaling,ma2024sit,chen2023pixart,mo2024dit,peebles2023scalable}. 
On the other hand, they require significant computational resources \cite{liu2024sora}, making them challenging to train outside of specialized industrial labs.
Therefore, it is natural to ask whether it is possible to make them lighter and faster without sacrificing performance. 
Answering these questions requires a fundamental understanding of the DiT architecture. 
This work provides a timely theoretical analysis of the fundamental limits of DiT architecture, aided by the analytical feasibility provided by the low-dimensional data assumption.

Empirically, 
Latent Diffusion is a go-to design for effectiveness and computational efficiency \cite{rombach2022high,liu2021robust,pope2021intrinsic,su2018learning}. 
Theoretically,
it is capable of hosting the assumption of low-dimensional data structure (see \cref{assumption:1} for formal definition) for detailed analytical characterization \cite{chen2023score,de2022convergence}.
In essence, diffusion models with low-dimensional data structures manifest a natural lower-dimensional diffusion process through an encoder/decoder within a robust and informative latent representation feature space \cite{rombach2022high,pope2021intrinsic}. 
Such lower-dimensional diffusion improves computational efficiency by reducing data complexity without sacrificing essential information \cite{liu2021robust}.
With this assumption, \citet{chen2023score} decompose the score function of U-Net based diffusion models into on-support and orthogonal components. This decomposition allows for the characterization of the distinct behaviors of the two components: the on-support component facilitates latent distribution learning, while the orthogonal component facilitates subspace recovery.

In our work, we utilize low-dimensional data structure assumption to explore statistical and computational limits of latent DiTs.
Our analysis includes the characterizations of statistical rates and provably efficient criteria.
Statistically, 
we pose two questions and provide a theory to characterize the statistical rates of latent DiT under the assumption of a low-dimensional data:
\begin{question}\label{Q1}
What is the approximation limit of using transformers to approximate the DiT score function, particularly in the low-dimensional data subspace?
\end{question}
\begin{question}\label{Q2}
How accurate is the estimation limit for such a score estimator in practical training scenarios?
With the score estimator, how well can diffusion transformers recover the data distribution? 
\end{question}
Computationally, the primary challenge of DiT lies in the transformer blocks' quadratic complexity. This computational burden applies to both inference and training, even with latent diffusion. 
Thus, it is essential to design algorithms and methods to circumvent this 
 $\Omega(L^2)$ where $L$ is the latent DiT sequence length. 
However, there are no formal results to support and characterize such algorithms. 
To address this gap, we pose the following questions and provide a fundamental theory to fully characterize the complexity of latent DiT under the low-dimensional linear subspace data assumption:
\begin{question}\label{Q3}
    Is it possible to improve the $\Omega(L^2)$ time complexity with a bounded approximation error for both forward and backward passes? 
    What is the computational limit for such an improvement?
\end{question}

\paragraph{Contributions.} 
We study the fundamental limits of latent DiT.
Our contributions are threefold:
\begin{itemize}
    \item 
    \textbf{Score Approximation.} 
    We address \cref{Q1} by characterizing the approximation limit of matching the DiT score function with a transformer-based score estimator. 
    Specifically, under mild data assumptions, we derive an approximation error bound for the score network, sub-linear in the latent space dimension (\cref{theorem:1}). 
    These results not only explain the expressiveness of latent DiT (under mild assumptions) but also provide guidance for the structural configuration of the score network for practical implementations (\cref{theorem:1}).

    \item
    \textbf{Score and Distribution Estimation.}
    We address \cref{Q2} by exploring the limitations of score and distribution estimations of latent DiTs in practical training scenarios. 
    Specifically, 
    we provide a sample complexity bound for score estimation (\cref{corollary:score_est}), 
    using norm-based covering number bound of transformer architecture.
    Additionally, we show that the learned score estimator is able to recover the initial data distribution (\cref{corollary:dist_est}).

    \item  
    \textbf{Provably Efficient Criteria and Existence of Almost Linear Time Algorithms.}
    We address \cref{Q3} by providing provably efficient criteria for latent DiTs in both forward inference and backward computation/training. 
    For forward inference, we characterize all possible efficient DiT algorithms using a norm-based efficiency threshold for both conditional and unconditional generation (\cref{thm:forward_efficiency_threshold}). 
    Efficient algorithms, including almost-linear time algorithms (\cref{prop:nearly_linear_inf}), are possible only below this threshold. 
    For backward computation, we prove the existence of almost-linear time DiT training algorithms (\cref{thm:main_comp_back}) by utilizing the inherent low-rank structure in DiT gradients through a chained low-rank approximation.
\end{itemize}
Interestingly, both our statistical and computational results are dominated by the subspace dimension under the low-dimensional assumption, suggesting that latent DiT can potentially bypass the challenges associated with the high dimensionality of initial data.

\paragraph{Organization.}
\cref{sec:background} includes background on score decomposition and Transformer-based score networks. 
\cref{sec:method} includes DiTs' statistical rates. 
\cref{sec:comp} includes DiTs' provably efficient criteria. 
\cref{sec:conclusion} includes concluding remarks.
We defer discussions of related works to \cref{sec:related_works}.

\paragraph{Notations.}
We use lower case letters to denote vectors, e.g., $z \in \RR^{D}$. 
$\norm{z}_2$ and $\norm{z}_{\infty}$ denote its Euclidean norm and Infinite norm respectively. 
We use upper case letters to denote matrix, e.g., $Z \in \RR^{d\times L}$. 
$\norm{Z}_2$, $\norm{Z}_{\rm op}$, and $\norm{Z}_F$ denote the $2$-norm, operator norm and Frobenius norm respectively. 
$\norm{Z}_{p,q}$ denotes the $p,q$-norm where the $p$-norm is over columns and $q$-norm is over rows.
Given a function $f$, let $\norm{f(x)}_{L^2}\coloneqq(\int \norm{f(x)}_2^2\dd x )^{1/2}$, and $\norm{f(\cdot)}_{Lip} = \sup_{x\neq y}(\norm{f(x)-f(y)}_2/\norm{x-y}_2)$.
With a distribution $P$, we denote $\norm{f}_{L^2(P)}=(\int_{P} \norm{f(x)}_2^2\dd x )^{1/2}$ as the $L^2(P)$ norm. 
Let $f_{\sharp} P$ be a pushforward measure, i.e., for any measurable $\Omega$, $(f_{\sharp}P)(\Omega) = P(f^{-1}(\Omega))$.
We use $\psi$ for (conditional) Gaussian density functions.

\section{Background}
\label{sec:background}

This section reviews the ideas we built on, including an overview of diffusion models (\cref{sec:score_matching_DM}), the score decomposition under the linear latent space assumption (\cref{subsec:score_decom}), and the transformer backbone in DiT (\cref{sec:latent_score}).

\subsection{Score-Matching Denoising Diffusion Models}
\label{sec:score_matching_DM}
We briefly review forward process, backward process and score matching in diffusion models.

\paragraph{Forward and Backward Process.}
In the \textbf{forward} process, Diffusion models gradually add noise to the original data $x_0\in \mathbb{R}^{D}$, and $x_0 \sim P_{0}$.
Let $x_t$ denote the noisy data at time stamp $t$, with marginal distribution and destiny as $P_t$ and $p_t$.
The conditional distribution $P(x_t|x_0)$ follows $N(\beta(t)x_0, \sigma(t)I_D)$, where $\beta(t)={\exp}(-\int_0^t w(s) \mathrm{d}  s/2)$, $\sigma(t)=1-\beta^2(t)$, and $w(t)>0$ is a nondecreasing weighting function. 
In practice, the forward process terminates at a large enough $T$ such that $P_T$ is close to $N(0,I_D)$.
In the \textbf{backward} process, we obtain $y_t$ by reversing the forward process.
The generation of $y_t$ depends on the score function $\nabla \log p_t(\cdot)$.
However, this is unknown in practice, we use a score estimator $s_W(\cdot,t)$ to replace $\nabla\log p_t(\cdot)$, where $s_W(\cdot,t)$ is usually a neural network with parameters $W$.
See \cref{appendix:subsec_diff} for the details.

\paragraph{Score Matching.}
To estimate the score function, we use the following loss
\begin{align*}
\min_{W} \int_{T_0}^T \gamma(t) \EE_{x_t \sim P_t} \left[\norm{s_W(x_t, t) - \nabla \log p_t(x_t)}_2^2 \right] \dd  t,
\end{align*}
where $\gamma(t)$ is the weight function, and $T_0$ is a small value to stabilize training and prevent score function from blowing up \cite{Vahdat2021}.
However, it is hard to compute $\nabla \log p_t(\cdot)$ with available data samples. 
Therefore, we minimize the equivalent denoising score matching objective
\begin{align}    \label{eq:denoising_score_matching}
    \min_{W} \int_{T_0}^T \gamma(t) \EE_{x_0 \sim P_0} \left[\EE_{x_t|x_0}\left[\left\| s_W(x_t , t) - \nabla_{x_t} \log \psi_t(x_t \mid x_0) \right\|_2^2 \right] \right]\dd t,
\end{align}
where $\psi_t(x_t | x_0)$ is the transition kernel, then $\nabla_{x_t} \log \psi_t(x_t | x_0) = \left(\beta(t)x_0 -x_t\right)/\sigma(t)$.

To train the parameters $W$ in the score estimator $s_W(\cdot,t)$, we use the empirical version of \eqref{eq:denoising_score_matching}. 
We select $n$ i.i.d. data samples $\{x_{0,i}\}_{i=1}^{n} \sim P_0$, and sample time $t_i$ $(1 \leq i \leq n)$ uniformly from interval $[T_0, T]$.
Given $x_{0,i}$, we sample $x_{t_i}$ from $N(\beta(t_i)x_{0,i}, \sigma(t_i)I_D)$.  
The empirical loss is
\begingroup
\setlength{\abovedisplayskip}{1pt}
\setlength{\abovedisplayshortskip}{1pt}
\setlength{\belowdisplayskip}{1pt}
\setlength{\belowdisplayshortskip}{1pt}
\begin{align}\label{eqn:empirical_loss}
    \hat{\cL}(W) & = \frac{1}{n} \sum_{i=1}^n \norm{s_W(x_{t_i}, t_i)-x_{0,i}}_2^2
    .
\end{align}
\endgroup
For convenience of notation, we denote population loss $\cL(W) = \EE_{P_0} [\hat{\cL}(W)]$.

\subsection{Score Decomposition in Linear  Latent Space}
\label{subsec:score_decom}
In this part, we review the score decomposition in \cite{chen2023score}.
We consider that the $D$-dimensional input data $x$ supported on a $d_0$-dimensional subspace, where $d_0 \le D$.
\begin{assumption}[Low-Dimensional Linear Latent Space]
\label{assumption:1}
Let $x$ denote the initial data at $t=0$.
$x$ has a latent representation via $x = B h$, where $B \in \mathbb{R}^{D \times d_0}$ is an unknown matrix with orthonormal columns. 
The latent variable $h \in \mathbb{R}^{d_0}$ follows the distribution $P_h$ with a density function $p_h$.
\end{assumption}
\begin{remark}
    By ``linear latent space,''
    we mean that each entry of a given latent vector is a linear combination of the corresponding input, i.e., $x=Bh$. 
    This is also known as the ``low-dimensional data'' assumption in literature  \cite{chen2023score}.
\end{remark}

Let $\Bar{x}$ and $\Bar{h}$ denote the perturbed data and its associated latent variable at $t>0$, respectively.
Based on the low-dimensional data structure assumption, we have the following score decomposition theory: on-support score $s_+(B^\top \Bar{x}, t)$ and orthogonal score $s_-(\Bar{x}, t)$.
\begin{lemma}[Score Decomposition, Lemma 1 of \cite{chen2023score}]
\label{lemma:subspace_score}
Let data $x = Bh$ follow \cref{assumption:1}. 
The decomposition of score function $\nabla \log p_t(\Bar{x})$ is
\begin{align}\label{eq:score_decom}
~\nabla \log p_t(\Bar{x}) = \underbrace{B\nabla \log p_t^{h}(\Bar{h})}_{s_+(\Bar{h}, t)}   \underbrace{-  \left(I_D - BB^\top \right) \Bar{x}/\sigma(t)}_{s_-(\Bar{x}, t)}, \quad
\Bar{h}=B^\top \Bar{x},
\end{align}
where
$p_t^{h}(\Bar{h}) \coloneqq  \int  \psi_t(\Bar{h}|h)p_h(h) \dd h$, 
$\psi_t( \cdot | h)$ is the Gaussian density function of $N(\beta(t)h, \sigma(t)I_{d_0})$, $\beta(t) = e^{-t/2}$ and $\sigma(t) = 1 - e^{-t}$.
We restate the proof in \cref{pf:subspace_score} for completeness.
\end{lemma}
Additionally, our theoretical analysis is based on two following assumptions as in \cite{chen2023score}.
\begin{assumption}[Tail Behavior of $P_h$]\label{assumption:2}
    The density function $p_h > 0$ is twice continuously differentiable. 
    Moreover, there exist positive constants $A_0, A_1, A_2$ such that when $\norm{h}_2 \geq A_0$, the density function $p_h(h) \leq (2\pi)^{-d_0/2} A_1 {\exp} (-A_2  \| h \|_2^2 / 2)$.
\end{assumption}

\begin{assumption}[$L_{s_+}$-Lipschitz of $s_+(\Bar{h}, t)$]\label{assumption:3}
    The on-support score function $s_+(\Bar{h}, t)$ is $L_{s_+}$-Lipschitz in $\Bar{h} \in \mathbb{R}^{d_0}$ for any $t \in [0, T]$.
\end{assumption}

\subsection{Score Network and Transformers}
\label{sec:latent_score}
In this part, we introduce the score network architecture and Transformers.
Transformers are the backbone of the score network in DiT.
By \cref{assumption:1}, $\Bar{h}=B^\top  \Bar{x} \in\R^{d_0}$ with $d_0<D$.

\paragraph{(Latent) Score Network.}
Following \cite{chen2023score},
we rearrange \eqref{eq:score_decom} into
\begin{align}
    \label{eq:score_docom_rearange}
    \nabla \log p_t(\Bar{x}) = B ( \underbrace{\sigma(t)\nabla \log p_t^{h} ( B^\top \Bar{x})
    +B^{\top} \Bar{x}}_{\coloneqq q(B^\top \Bar{x},t):\;\R^{d_0} \times [T_0, T] \;\to\; \R^{d_0}})/\sigma(t) -  \Bar{x}/\sigma(t).
\end{align}
We use $W_B\in \RR^{D\times d_0}$ to approximate $B\in \RR^{D\times d_0}$, and a neural network $f(W_B^\top \Bar{x}, t)$ to approximate $q(B^\top \Bar{x},t)$. 
We adopt the following score network class for diffusion in latent space (i.e., in $\Bar{h}\in\R^{d_0}$)
\begin{align}\label{DiT_net_class}
    \mathcal{S}=
    \left\{s_W(\Bar{x},t)=W_Bf(W_B^T \Bar{x},t)/\sigma(t)-\Bar{x}/\sigma(t), ~ W=\{W_B, f\} 
    \right\},
\end{align}
where the columns in $W_B $ are orthogonal, $f: \RR^{d_0} \times [T_0, T] \rightarrow \RR^{d_0}$ is a neural network.
In this work, we focus on the diffusion transformers (DiTs), i.e., using Transformer for $f$ \cite{peebles2023scalable}.

\paragraph{Transformers.}
A Transformer block consists of a self-attention layer and a feed-forward layer, with both layers having skip connection. 
We use $\tau^{r,m,l}: \mathbb{R}^{d\times L}\rightarrow \mathbb{R}^{d\times L}$ to denote a Transformer block.
Here $r$ and $m$ are the number of heads and head size in self-attention layer, and $l$ is the hidden dimension in feed-forward layer.
Let $X\in\R^{d\times L}$ be the model input, then we have the model output
\begin{align}
    \label{eq:Attn}
    ~ {\rm Attn}(X) 
    &= X + 
    \sum\nolimits_{i=1}^r W_O^i W_V^i X \cdot \Softmax \left(\left(W_K^i X\right)^\sT W_Q^i X\right),
    \\
    \label{eq:FF}
    ~ {\rm FF}\circ{\rm Attn}(X) 
    &= {\rm Attn}(X) + W_2 \cdot {\rm ReLU} (W_1 \cdot {\rm Attn}(X) + b_1 \one_L^\sT) + b_2 \one_L^\sT,
\end{align}
where $W_K^i, W_Q^i, W_V^i \in \R^{m \times d}, W_O^i \in \R^{d \times m}, W_1 \in \R^{l \times d}, W_2 \in \R^{d \times l}, b_1 \in \RR^{l}, b_2 \in \R^{d}$.

In our work, we use Transformer networks with positional encoding $E\in\R^{d\times L}$.
We define the Transformer networks as the composition of Transformer blocks
\begin{align*}
    \calT_{P}^{r,m,l}=\{f_{\cT}:\mathbb{R}^{d\times L}\rightarrow{\mathbb{R}^{d\times L}}\mid f_{\cT}\text{ is a composition of blocks }\tau^{r,m,l}\text{'s}\}.
\end{align*}
For example, the following is a Transformer network consisting $K$ blocks and positional encoding
\begin{align}\label{eq:trans_net}
f_{\cT}(X)= {\rm FF}^{(K)} \circ {\rm Attn}^{(K)} \circ  \cdots {\rm FF}^{(1)} \circ  {\rm Attn}^{(1)} (X+E).
\end{align}

\section{Statistical Rates of Latent DiTs with Subspace Data Assumption}
\label{sec:method}
In this section, we analyze the statistical rates of latent DiTs. \cref{subsec:dit_class} introduces the class of latent DiT score networks. 
In \cref{subsec:score_approximation}, we prove the approximation limit of matching the DiT score function with the score network class, and characterize the structural configuration of the score network when a specified approximation error is required.
Following this, in \cref{subsec:score_dist_est}, utilizing the characterized structural configuration, we prove the score and distribution estimation for latent DiTs.

\subsection{DiT Score Network Class}
\label{subsec:dit_class}
Here, we provide the details about DiT score network class used in our analysis.
In \eqref{DiT_net_class}, $f$ is a network with Transformer as the backbone, and $(\Bar{h},t) \in \RR^{d_0} \times [T_0, T]$ denotes the input data.
Following \cite{peebles2023scalable}, DiT uses time point $t$ to calculate the scale and shift value in the Transformer backbone, and it transforms an input picture into a sequential version.
To achieve the transformation, we introduce a reshape layer.
\begin{definition}[DiT Reshape Layer $R(\cdot)$]
\label{def:reshape_layer}
     Let $R(\cdot) : \mathbb{R}^{d_0} \to \mathbb{R}^{d \times L}$ be a reshape layer that transforms the $d_0$-dimensional input into a $d \times L$ matrix. 
     Specifically, for any $d_0=i\times i$ image input, $R(\cdot)$ converts it into a sequence representation with feature dimension $d \coloneqq p^2$ (where $p \geq 2$) and sequence length $L \coloneqq \left(i/p\right)^2$.
     Besides, we define the corresponding reverse reshape (flatten) layer $R^{-1} (\cdot): \mathbb{R}^{d \times L} \to \mathbb{R}^{d_0}$ as the inverse of $R(\cdot)$.
     By $d_0= dL$, $R,R^{-1}$ are associative w.r.t. their input. 
\end{definition}

To simplify the self-attention block in \eqref{eq:Attn}, let $W_{OV}^i = W_O^i W_V^i$ and $W_{KQ}^i = (W_K^i)^\sT W_Q^i$.

\begin{definition}[Transformer Network Class $\calT_p^{r,m,l}$]\label{def:Trans_net_class}
\label{def:trans}
    We define the Transformer network class as
    \begin{align*}
        \calT_p^{r,m,l} & (K, C_{\cT}, C_{OV}^{2,\infty}, C_{OV}, C_{KQ}^{2,\infty}, C_{KQ}, C_{F}^{2,\infty}, C_{F}, C_E, L_{\cT}),~
        \text{satisfying the constraints}
    \end{align*}
    
    \begin{itemize}[after=\vspace{-.2em}]
        \item Model architecture with $K$ blocks: $f_{\cT}(X)= {\rm FF}^{(K)} \circ {\rm Attn}^{(K)} \circ  \cdots {\rm FF}^{(1)} \circ  {\rm Attn}^{(1)} (X)$;
        \item Model output bound: $\sup_{X}\norm{f_{\cT}(X)}_2\leq C_{\cT}$;
        \item Parameter bound in ${\rm Attn^{(i)}}$:  $\norm{(W_{OV}^i)^\top}_{2,\infty}\leq C_{OV}^{2,\infty}$, $\norm{(W_{OV}^i)^\top}_2 \leq C_{OV}$, $\norm{W_{KQ}^i}_{2,\infty}\leq C_{KQ}^{2,\infty}$, $\norm{W_{KQ}^i}_{2}\leq C_{KQ}$, 
        $\norm{E^\top}_{2,\infty}\leq C_E, \forall i\in[K]$;
        \item Parameter bound in ${\rm FF^{(i)}}$:      $\norm{W_{j}^i}_{2,\infty}\leq C_{F}^{2,\infty}, \norm{W_{j}^i}_{2}\leq C_{F}, \forall j\in[2], i\in[K] $;
        \item Lipschitz of $f_{\cT}$: $\norm{f_{\cT}(X_1)-f_{\cT}(X_2)}_F\leq L_{\cT}\norm{X_1-X_2}_F, \forall X_1,X_2 \in \RR^{d\times L}$.
    \end{itemize}
\end{definition}

\begin{figure}[t!]

\centering
\definecolor{blackborder}{RGB}{0,0,0}  %
\definecolor{whitebg}{RGB}{255,255,255} %

\resizebox{\textwidth}{!}{%
\begin{tikzpicture}[
    encoder/.style={trapezium, trapezium angle=60, draw=blackborder, fill=whitebg, thick, minimum height=1.2cm, minimum width=1.0cm, align=center, rotate=270},
    decoder/.style={trapezium, trapezium angle=60, draw=blackborder, fill=whitebg, thick, minimum height=1.2cm, minimum width=1.0cm, align=center, rotate=90},
    network/.style={rectangle, draw=blackborder, fill=whitebg, thick, minimum height=2cm, minimum width=3cm, align=center},
    greyblock/.style={rectangle, draw=blackborder, fill=whitebg, thick, minimum height=2cm, minimum width=1cm, align=center},
    yellowblock/.style={rectangle, draw=blackborder, fill=whitebg, thick, minimum height=1.5cm, minimum width=1.2cm, align=center},
    concatblock/.style={rectangle, draw=blackborder, fill=whitebg, thick, minimum height=2.0cm, minimum width=1.2cm, align=center},
    ffblock/.style={rectangle, draw=blackborder, fill=whitebg, thick, minimum height=1.5cm, minimum width=1cm, align=center},
    attnblock/.style={rectangle, draw=blackborder, fill=whitebg, thick, minimum height=1.5cm, minimum width=1cm, align=center},
    sum/.style={circle, draw=blackborder, fill=whitebg, thick, minimum size=0.4cm},
    shortcut/.style={dashed,  -{Latex[scale=1.5]}},
    myarrow/.style={thick, ->, -{Latex[scale=1.5]}},
    node distance=0.8cm, auto, scale=0.75, transform shape,
    decorate,decoration={brace,amplitude=10pt,raise=3pt},
    smallrect/.style={rectangle, draw=black, fill=whitebg, minimum width=0.01cm, minimum height=0.01cm},
    highlightrect/.style={rectangle, draw=black, fill=whitebg, minimum width=0.01cm, minimum height=0.01cm},
    scale=0.85
]

\node (input) at (-4.1,0) {};
\node (y) at (-1.8,-1) {};

\node[encoder] at (-1.5,0) (encoder) {\rotatebox{90}{$ W_B^{\top} $}};
\node at (encoder.north) [above, xshift=-0.5cm, yshift=+1.2cm, text=black] {\shortstack{Latent Encoder}};

\node[decoder] at (12.5,0) (decoder) {\rotatebox{-90}{$W_B$}};
\node at (decoder.north) [above, xshift=0.5cm, yshift=+1.2cm, text=black] {\shortstack{Latent Decoder}};

\node[yellowblock] at (1.5,0) (reshape) {{$ R(\cdot) $}};
\node at (reshape.north) [above, xshift=0.0cm, yshift=-0.0cm, text=black] {\shortstack{Reshape Layer}};

\node[network] at (5.5,0) (network) {$f_{\calT} \in \calT^{r,m,l}$};
\node at (network.south) [above, xshift=0.0cm, yshift=+2cm] {\shortstack{Transformer Network}};

\node[yellowblock] at (9.5,0) (reshapei) {{$ R^{-1}(\cdot) $}};

\node at (reshapei.north) [above, xshift=0.0cm, yshift=-0.0cm, text=black] {\shortstack{Reversed\\ Reshape Layer}};

\node (sum) at (15,0) {{\LARGE $\oplus$}};
\node at (16.5,0) (output) {};

\draw[myarrow] (input) -- (encoder) node[midway, below, yshift=-.2cm, xshift=-0.1cm] {$x \in \R^{D}$};

\draw[myarrow] (encoder) -- (reshape) node[midway, below, yshift=-.2cm,xshift=-0.1cm] {$x \in \R^{d_0}$};

\draw[myarrow] (reshape) -- (network) node[midway, below, xshift=0.0cm, yshift=-.3cm] {$ \R^{d \times L} $};

\draw[myarrow] (reshapei) -- (decoder) node[midway, below, yshift=-.2cm] {$ \R^{d_0} $};

\draw[myarrow] (decoder) -- (sum) node[midway, below, yshift=-.2cm] {$ \R^{D} $};
\draw[myarrow] (decoder) -- (sum) node[midway, below, yshift=+0.8cm] {$ 1/\sigma(t) $};

\draw[myarrow] (network) -- (reshapei) node[midway, below, yshift=-.2cm] {$ \R^{d \times L} $};

\draw[shortcut] (-4,1) -- ++(0,1) -- ++(19,0) node[midway, below] {$-{1}/{\sigma_{t}^{2}}$} -- (sum.north);

\draw[decorate,decoration={brace,amplitude=5pt,mirror,raise=5pt},thick]
(reshape.south west) -- (reshapei.south east) 
node[midway,below,yshift=-0.5cm] 
{$f = R^{-1} \circ f_\calT \circ R \quad : \quad \R^{d_0} \to \R^{d \times L}$};

\draw[myarrow] (sum) -- (output) node[midway, below, yshift=-.2cm] {$s_W(x, t)$};

\end{tikzpicture}
}
\vspace{-1.5em}
\caption{\small \textbf{Overview of DiT Score Network Architecture} $s_W(\cdot, t)$. 
$W_B^T$ denotes the linear layer from the input data space to the linear latent space.
$f(\cdot) = {R^{-1}\circ f_{\calT} \circ R}(\cdot)$ denotes the transformer network $f_{\calT}(\cdot)$ with reshaping layer $R(\cdot)$, where $f_{\calT}(\cdot) \in \calT_p^{r,m,l}$.
$W_B$ denotes the linear layer from the linear latent space to the input data space.
$\sigma(t)$ denote the variance of the conditional distribution $P(x_t \mid x_0)$.}
\label{fig:pipeline}
\vspace{-1.5em}
\end{figure}
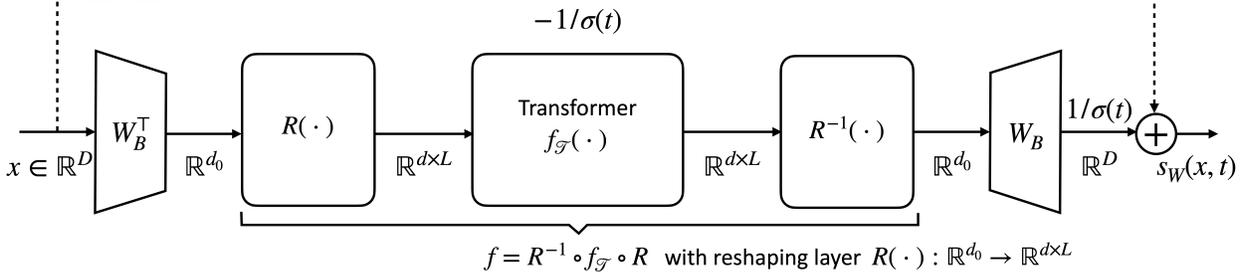

\begin{definition}[DiT Score Network Class $\mathcal{S}_{\calT_p^{r,m,l}}$ (\cref{fig:pipeline})]
\label{def:DiT_function_class}
    We denote $\mathcal{S}_{\calT_p^{r,m,l}}$ as the DiT score network class in \eqref{DiT_net_class}, replacing $f$ with ${R^{-1}\circ f_{\calT} \circ R}$, and $f_{\calT}$ is from the Transformer class $\calT_p^{r,m,l}$.
\end{definition}

\subsection{Score Approximation of DiT}
\label{subsec:score_approximation}
Here, we explore the approximation limit of latent DiT score network class $\mathcal{S}_{\calT_p^{r,m,l}}$ under linear latent space assumption.
Recall that $P_t$ is the distribution of $x_t$, $\sigma(t)$ is the variance of $P(x_t|x_0)$, $d_0$ is the dimension of latent space, $L$ is the sequence length of transformer input, $T$ is the stopping time in forward process, $T_0$ is the early stopping time in backward process, and $L_{s_+}$ is the Lipschitz coefficient of on-support score function.
Then we have the following \cref{theorem:1}.

\begin{theorem}[Score Approximation of DiT]
\label{theorem:1}
For any approximation error $\epsilon>0$ and any data distribution $P_0$ under \cref{assumption:1,assumption:2,assumption:3}, there exists a DiT score network $s_{\hat{W}}$ from $\mathcal{S}_{\cT_p^{2,1,4}}$ (defined in \cref{def:trans}), where 
$\hat{W}=\{\hat{W}_B, \hat{f}_{\cT}\}$, such that for any $t\in [T_0,T]$, we have:
\begin{align*}
\norm{s_{\hat{W}}(\cdot,t)-\nabla\log p_t(\cdot)}_{L^2(P_t)}\leq \epsilon\cdot \sqrt{d_0}/\sigma(t),
\end{align*}
where $\sigma(t)=1-e^{-t}$, and the upper bound of hyperparameters in $\mathcal{S}_{\cT_p^{2,1,4}}$ are
\begin{align*}
    &~K= \calO(\epsilon^{-2L}), ~ C_{\cT}=\calO\left(d_{0} L_{s_{+}}  \sqrt{d_0\log (d_0/T_0)+\log (1/\epsilon)}\right),\\
    &~ C_{OV}^{2,\infty}=  (1/\epsilon)^{\calO(1)}, ~ C_{OV}=  (1/\epsilon)^{\calO(1)}, ~ C_{KQ}^{2,\infty}=  (1/\epsilon)^{\calO(1)}, ~ C_{KQ}=  (1/\epsilon)^{\calO(1)},
    \\
    &~C_E=\calO(L^{3/2}), ~ C_{F}^{2,\infty}=  (1/\epsilon)^{\calO(1)}, ~ C_{F}=  (1/\epsilon)^{\calO(1)}, ~ L_{\cT}=\calO\left(d_{0}L_{s_{+}}\right). 
\end{align*}
\end{theorem}

\begin{proof}[Proof Sketch]
Our proof is built on the key observation that there is a tail behavior of the low-dimensional latent variable distribution $P_h$ (\cref{assumption:2}).
Recall that $\nabla \log p_t(\Bar{x}) = B q(\Bar{h}, t)/\sigma(t) -  \Bar{x}/\sigma(t)$, where $\Bar{h}=B^\top \Bar{x}$ (defined in \eqref{eq:score_docom_rearange}).
By taking $\hat{W}_{B}=B$, our aim reduces to construct a transformer network to approximate $q(\Bar{h}, t)$.
To achieve this, we firstly approximate $q(\Bar{h}, t)$ with a compact-supported continuous function, based on the tail behavior of $P_h$.
Then we construct a transformer to approximate the compact-supported continuous function using the universal approximation capacity of transformer \cite{yun2019transformers}.
See \cref{subsec:pf_thm_1} for a detailed proof.
\end{proof}

Intuitively, \cref{theorem:1} indicates the capability of the transformer-based score network to approximate the score function with precise guarantees. 
Furthermore, \cref{theorem:1} provides empirical guidance for the design choices of the score network when a specified approximation error is required.

\begin{remark}[Comparing with Existing Works]
    Theoretical analysis of DiTs is limited. Previous works that do not specify the model architecture assume that the score estimator is well-approximated \cite{benton2023linear,wibisono2024optimal}. 
    To the best of our knowledge, this work is the first to present an approximation theory for DiTs, offering the estimation theory in \cref{corollary:score_est,corollary:dist_est} based on the estimated score network, rather than a perfectly trained one.
\end{remark}

\begin{remark}[Latent Dimension Dependency]
    \cref{theorem:1} suggests that the approximation capacity and Transformer network size primarily depend on the latent variable dimension $d_0=d \times L$.
    This indicates that DiTs can potentially bypass the challenges associated with the high dimensionality of initial data by transforming input data into a low-dimensional latent variable.
\end{remark}

\subsection{Score Estimation and Distribution Estimation}
\label{subsec:score_dist_est}

Besides score approximation capability, 
\cref{theorem:1} also characterizes the structural configuration of the score network for any specific precision, e.g., $K,C_E,C_F$, etc.
This characterization enables further analysis of the performance of score network in practical scenarios.
In \cref{corollary:score_est},
we provide a sample complexity bound for score estimation. 
In \cref{corollary:dist_est},
show that the learned score estimator is able to recover the initial data distribution.

\paragraph{Score Estimation.}
To derive a sample complexity for score estimation using $\mathcal{S}_{\cT_p^{2,1,4}}$, we rewrite the score matching objective in \eqref{eqn:empirical_loss} as 
$\hat{W} \in \argmin_{s_{W} \in \mathcal{S}_{\cT_p^{2,1,4}}} \hat{\cL}(s_{W}), ~ \hat{W}=\{\hat{W}_B, \hat{f}_{\cT}\}$.

\cref{corollary:score_est} shows that as sample size $n \rightarrow \infty$, $s_{W}(\cdot,t)$ convergences to $\nabla \log p_t(\cdot)$.

\begin{theorem}[Score Estimation of DiT]
\label{corollary:score_est}
    Under \cref{assumption:1,assumption:2,assumption:3}, we choose $\mathcal{S}_{\cT_p^{2,1,4}}$ as in  \cref{theorem:1} using $\epsilon \in (0,1)$ and $L> 1$,
    With probability $1-1/\poly(n)$, we have
    \begin{align}
        \label{eq:score_estimation}
       &~\frac{1}{T-T_0}\int_{T_0}^{T}\norm{ s_{\hat{W}}(\cdot,t)-\nabla\log p_t(\cdot)}_{L^2(P_t)} \dd t 
        =\tilde{\calO}
        \( \frac{1}{n^{1/3} T_0 T}\cdot 2^{(1/\epsilon)^{2L}} + \frac{1}{n^{1/3} T_0 T} + \frac{1}{ T_0 T} \epsilon^2 \),
    \end{align}
    where $\tilde{\calO}$ hides the factors related to $D, d_0, d, L_{s_+}$, and $ \log n$.   
\end{theorem}
\begin{proof}
    See \cref{subsec:pf_thm_2} for a detailed proof.
\end{proof}
Intuitively, \cref{corollary:score_est} shows a sample complexity bound for score estimation in practice.
\begin{remark}[Comparing with Existing Works]
    \cite{zhu2024sample} provides a sample complexity for simple ReLU-based diffusion models under the assumption of an accurate score estimator. 
    To the best of our knowledge, we are the first to provide a sample complexity for DiTs, based on the learned score network in \cref{theorem:1} and the quantization (piece-wise approximation) approach for transformer universality \cite{yun2019transformers}.
    Furthermore, our first term shows a convergence rate of $1/T$, outperforming \cite{chen2023score}, in which the first term is independent of $T$.
\end{remark}

\begin{remark}[Double Exponential Factor and Inconsistent Convergence]
\label{rk:neg_result}
\cref{corollary:score_est}  reports an explicit result on sample complexity bounds for score estimation of latent DiTs: a double exponential factor $ 2^{(1/\epsilon)^{2L}}$ in the first term.
We remark that this arises from 
    the required depth $K$ is $\calO(\epsilon^{-2L})$, and the norm of required weight parameters is $(1/\epsilon)^{\calO(1)}$ as shown in \cref{theorem:1}, assuming the universality of transformers requires dense layers \cite{yun2019transformers}.
    This double exponential factor causes inconsistent convergence with respect to sample size $n$, as its large value prevents setting $\epsilon$ as a function of $n$ to balance the first and second terms in \eqref{eq:score_estimation}.
    This motivates us to rethink transformer universality and explore new proof techniques for DiTs, which we leave for future work.
\end{remark}

\begin{definition}
For later convenience, we define $\xi(n,\epsilon,L):=\frac{1}{n^{1/3}}\cdot 2^{(1/\epsilon)^{2L}} + \frac{1}{n^{1/3}} + \epsilon^2 $.
\end{definition}

\paragraph{Distribution Estimation.}
In practice, DiTs generate data using the discretized version with step size $\mu$, see \cref{appendix:subsec_diff} for details.
Let $\hat{P}_{T_0}$ be the distribution generated by $s_{\hat{W}}$ using the discretized backward process in \cref{corollary:score_est}.
Let $P_{T_0}^h$ and $p_{T_0}^h$ be the distribution and density function of on-support latent variable $\Bar{h}$ at $T_0$.
We have the following results for distribution estimation.

\begin{corollary}[Distribution Estimation of DiT, Modified From Theorem 3 of \cite{chen2023score}]
\label{corollary:dist_est}
    Let $T =\calO(\log n), T_0 =\calO(\min\{c_0,1/L_{s_+}\})$, where $c_0$ is the minimum eigenvalue of $\EE_{P_h}[h h^\top]$.
    With the estimated DiT score network $s_{\hat{W}}$ in \cref{corollary:score_est}, we have the following with probability $1-1/\poly(n)$.
    \begin{itemize}[leftmargin=2em]
        \item [(i)]
        The accuracy to recover the subspace $B$ is
        $\norm{W_B W_B^\top - BB^\top}_F^2=\tilde\calO\(\xi(n,\epsilon,L)/c_0\)$.
        \item [(ii)]
        With the conditions 
        ${\sf KL}(P_h || N (0, I_{d_0})) < \infty$,
        there exists an orthogonal matrix $U \in \RR^{d \times d}$ such that we have the following upper bound for the total variation distance 
        \begin{align}
        \label{eq:tv_item2}
                {\sf TV} (P_{T_0}^{h}, (W_BU)^\top_{\sharp} \hat{P}_{T_0})= \tilde\calO(\sqrt{\xi(n,\epsilon,L) \cdot \log n}),
        \end{align}
        where $\tilde{\calO}$ hides the factor about $D, d_0, d, L_{s_+}, \log n$, and $T-T_0$. and $(W_BU)^\top_{\sharp} \hat{P}_{T_0}$ denotes the pushforward distribution.

        \item [(iii)] 
        For the generated data distribution $\hat{P}_{T_0}$, the orthogonal pushforward $(I-W_BW_B^\top)_{\sharp} \hat{P}_{T_0}$ is ${N}(0, \Sigma)$, where $\Sigma \preceq aT_0 I$ for a constant $a > 0$.
    \end{itemize}
\end{corollary}

\begin{proof}
    See \cref{subsec:pf_thm_3} for a detailed proof.
\end{proof}

Intuitively, \cref{corollary:dist_est} shows the estimation results in 3 parts:
(i) the accuracy of recovering the subspace $B$; 
(ii) the estimation error between $\hat{P}_{T_0}$ and $P_{T_0}^h$; 
and (iii) the vanishing behavior of $\hat{P}_{T_0}$ in the orthogonal space.
These indicate that the learned score estimator is capable of recovering the initial data distribution.
Notably, \cref{corollary:dist_est} is agnostic to the specifics of $\xi(n,\epsilon,L)$. 

\begin{remark}[Comparing with Existing Works]
    \citet{oko2023diffusion} analyze the distribution estimation under the assumption that the initial density is supported on $[-1,1]^{D}$ and smooth in the boundary.
    Our \cref{assumption:2} demonstrates greater practical relevance. 
    This suggests that our method of distribution estimation aligns more closely with empirical realities.
\end{remark}

\begin{remark}[Subspace Recovery Accuracy]
    (i) of \cref{corollary:dist_est} confirms that the subspace is learned by DiTs. 
    The error is proportional to the sample complexity for score estimation and depends on the minimum eigenvalue of the covariance of $P_h$. 
\end{remark}

\section{Provably Efficient Criteria}
\label{sec:comp}
Here, we analyze the computational limits of latent DiTs under low-dimensional linear subspace data assumption (i.e., \cref{assumption:1}).
The hardness of DiT models ties to both forward and backward passes of the score network in \cref{def:DiT_function_class}.
We characterize them  separately.

\subsection{Computational Limits of Backward Computation}
\label{sec:comp_back}
Following \cref{sec:background},
suppose we have $n$ i.i.d. data samples $\{x_{0,i}\}_{i=1}^{n} \sim P_{d}$, and time $t_{i_0}$ $(1 \leq i \leq n)$ uniformly sampled from  $[T_0, T]$.
For each data $x_{0,i}\in\R^{D}$, we sample $x_{t_{i_0}}\in\R^{D}$ from $N(\beta(t_{i_0})x_{0,i}, \sigma(t_{i_0})I_D)$.  
Let $(W_AR^{-1}(\cdot))^\dagger$ be the inverse transformation of $W_AR^{-1}(\cdot)$, and denote $Y_{0,i}\coloneqq (W_AR^{-1})^\dagger(x_{0,i})\in \R^{d\times L}$.
We rewrite the empirical denoising score-matching loss  \eqref{eqn:empirical_loss} as
\begin{align}
    \label{eqn:DiT_empirical_loss}
    \frac{1}{n} \sum_{i=1}^n \Big\| W_AR^{-1}(f_\calT(R(\underbrace{W_A^\top x_{t_{i_0}}}_{d_0\times 1})))-x_{0,i}\Big\|_F^2 
    = 
    \frac{1}{n} \sum_{i=1}^n 
    \Big\|
    \underbrace{W_A}_{D\times d_0}\underbrace{R^{-1}\big(\overbrace{f_\calT(R(W_A^\top x_{t_{i_0}}})}_{d_0\times 1}^{d\times L})-\underbrace{Y_{0,i}}_{d\times L}\big)\Big\|_F^2.
\end{align}
For efficiency, it suffices to focus on just transformer attention heads of the DiT score network due to their
dominating quadratic time complexity in both passes.
Thus, we consider only a single layer attention for $f_\calT$, 
to simplify our analysis.
Further, we consider the following simplifications:
\begin{itemize}[leftmargin=2.2em]

    \item [(S0)] \label{item:S0}
    To prove the hardness of \eqref{eqn:DiT_empirical_loss} for both full gradient descent and stochastic mini-batch gradient descent methods, 
    it suffices to consider training on a single data point.
    
    \item [(S1)] \label{item:S1}
    For the convenience of our analysis, 
    we consider the following expression for attention mechanism.
    Let $X,Y\in\R^{d\times L}$.
    Let $W_K,W_Q,W_V\in \R^{s\times d}$ be attention weights such that
    $Q=W_Q X\in\R^{d\times L}$, $K=W_K X\in\R^{s\times L}$ and $V=W_V X\in\R^{s\times L}$.
    We write attention mechanism of hidden size $s$ and sequence length $L$ as
\begin{align}
\label{eqn:attention}
{\rm Att}(X)
=\underbrace{( W_OW_VX)}_{V\text{ multiplication}} \underbrace{D^{-1}\exp( X^\sT W_K^\sT W_Q X )}_{K\text{-}Q \text{ multiplication}} 
\in\R^{d\times L},
\end{align}
with  $D\coloneqq \diag(\exp(XW_Q W_K^\sT X^\sT)\one_L)$.
Here, $\exp(\cdot)$ is entry-wise exponential function, 
i.e., $\exp(A)_{i,j}=\exp(A_{i,j})$ for any matrix $A$
, $\diag(\cdot)$ 
converts a vector into a diagonal matrix with the vector's entries on the diagonal, and $\one_L$ is the length-$L$ all ones vector.

    \item [(S2)] \label{item:S2} Since $V$ multiplication is linear in weight while $K$-$Q$ multiplication is exponential in weights,
    we only need to focus on the gradient update of $K$-$Q$ multiplication.
    Therefore, for efficiency analysis of gradient, it is equivalent to analyzing a reduced problem with fixed $ W_OW_VX=\text{const.}$.

    \item [(S3)]
    To focus on the DiT, we consider the low-dimensional linear encoder $W_A$ to be pretrained and to not participate in gradient computation. 
    This aligns with common practice \cite{rombach2022high} and is justified by the trivial computation cost due to the linearity of $W_A$\footnote{The gradient computation is linear in $W_A$, and hence the computation w.r.t. $W_A$ is cheap and upper-bounded by $L\cdot \poly(d)$ time in a straightforward way.}.

    \item [(S4)] \label{item:S4}
    To further simplify, we introduce $A_1,A_2,A_3\in\R^{s\times L}$ and $W\in\R^{d\times d}$ via
    \begin{align}\label{eqn:loss_simplified}
    &~ \Big\|
    W_AR^{-1}\big(f_\calT(\underbrace{R(W_A^\top x_{t_{i_0}})}_{\coloneqq X\in\R^{d\times L}})-\underbrace{Y_{0,i}}_{\coloneqq Y\in\R^{d\times L}}\big)\Big\|_F^2
    \annot{By \hyperref[item:S0]{(S0)}, \hyperref[item:S1]{(S1)} and \hyperref[item:S2]{(S2)}}
    \\
    =&~
    \Big\|
    W_AR^{-1}\big(\underbrace{W_OW_V}_{\coloneqq W_{OV}\in \R^{d\times d}}\underbrace{X}_{\coloneqq A_3\in \R^{d\times L}} D^{-1}\exp\big( \underbrace{X^\sT}_{\coloneqq A_1^\top \in \R^{L\times d}} \underbrace{W_K^\sT W_Q}_{\coloneqq W\in \R^{d\times d}} \underbrace{X}_{\coloneqq A_2\in \R^{d\times L}} \big) -Y\big)\Big\|_F^2.
    \end{align}
    Notably, $A_1,A_2,A_3, X, Y$ are constants w.r.t. training above loss with gradient updates.
\end{itemize}

Therefore, we simplify the objective of training DiT into
\begin{definition}[Training Generic DiT Loss]
\label{def:generic_dit_loss}
Given $A_1,A_2,A_3,Y\in \R^{d\times L}$ and $W_{OV},W\in\R^{d\times d}$ following \hyperref[item:S4]{(S4)}, 
Training a DiT with $\ell_2$ loss on a single data point $X,Y \in \R^{d\times L}$ is formulated as
\begin{align}\label{eqn:generic_dit_loss}
    \min_{W}~\calL_0(W) 
    = \min_{W}~\half 
    \Big\|
    W_AR^{-1}\big( W_{OV} A_3D^{-1}\exp(A_1^\top WA_2)-Y\big)\Big\|_F^2
    .
\end{align}
Here  
    $D :=  \diag(  \exp ( A_1^\top W A_2 ) {\one}_n ) \in \R^{L \times L}$.  
\end{definition}
\begin{remark}[Conditional and Unconditional Generation]\label{remark:generic}
    $\calL_0$ is generic.
    If $A_1\neq A_2\in\R^{d\times L}$, \cref{def:generic_dit_loss} reduces to cross-attention in DiT score net (for conditional generation).
    If $A_1=A_2\in\R^{d\times L}$, \cref{def:generic_dit_loss} reduces to self-attention in DiT score net (for unconditional vanilla generation).
\end{remark}
We introduce the next problem to characterize all possible gradient computations of optimizing \eqref{eqn:generic_dit_loss}.
\begin{problem}[Approximate DiT Gradient Computation ($\textsc{ADiTGC}(L,d,\Gamma,\epsilon)$)]
\label{prob:ADiTGC}
Given  $A_1,A_2,A_3,Y\in\R^{d\times L}$.
Let $\epsilon>0$.
Assume all numerical values are in $\cO( \log(L) )$-bits encoding.
Let loss function $\calL_0$ follow \cref{def:generic_dit_loss}.
The problem of approximating gradient computation of optimizing empirical DiT loss \eqref{eqn:generic_dit_loss} is to find an approximated gradient matrix $\Tilde{G}^{(W)}\in\R^{d\times d}$ such that
$\big\|\underline{\Tilde{G}}^{(W)}-\pdv{\calL}{\underline{W}}\big\|_{\max}\le 1/\poly (L)$.
Here, $\norm{A}_{\max}\coloneqq\max_{i,j} \abs{A_{ij}}$ for any matrix $A$. 
\end{problem}
In this work, we aim to investigate the computational limits of all possible efficient algorithms of $\textsc{ADiTGC}$ with $\epsilon=1/\mathrm{poly}(L)$.
Yet,
the explicit gradient of DiT denoising score matching loss \eqref{eqn:generic_dit_loss} is too complicated to characterize $\textsc{ADiTGC}$.
To combat this, we make the following observations.
\begin{itemize}[leftmargin=2.2em]
    \item [(O1)] \label{item:O1}
    Let $g_1(\cdot)\coloneqq W_AR^{-1}(\cdot): \R^{d\times L} \to \R^{d_0} $, 
    $g_2(\cdot)\coloneqq {\rm Att}(\cdot): \R^{d\times L}\to \R^{d\times L}$,
    and $g_3(\cdot)\coloneqq R(W_A^\top \cdot): \R^{D}\to \R^{d\times L}$ such that $g_3(x)=X$ for $x\in\R^{D}$ (with $D>d_0=dL$).

    \item [(O2)] \label{item:O2} \textbf{Vectorization of $f_\calT$.}
    For the ease of presentation, we use notation flexibly that $f_\calT$ to denote both a matrix in $\mathbb{R}^{d \times L}$ and a vector in $\R^{dL}$  in the following analysis.  
    This practice does not affect correctness. 
    The context in which $f_\calT$ is used should clarify whether it refers to a matrix or a vector.
    Explicit vectorization follows \cref{def:vectorization}.

    \item [(O3)] \label{item:O3} \textbf{Linearity of $g_1$.}
    By linearity of $W_AR^{-1}(\cdot)$, 
    we treat $g_1$ as a matrix in $\R^{d_0\times dL}$ acting on vector $f_\calT(\cdot)\in\R^{dL}$.

\end{itemize}
    Therefore, we have
    $\calL_0=\norm{g_1 \cdot \[g_2(g_3)-Y\]}_2^2$,
    such that its gradient involves $\dv{\calL_0}{W}=
    g_1 \dv{ g_2 }{W}$.
    From above,
    we only need to focus on proving the computation time and error control of term $\dv{ g_2 }{W}$ for gradient w.r.t $W$.
    Luckily, 
    with tools from fine-grained complexity theory \cite{alman2024fast,as23_tensor,as24_neurips,as24_rope} and tensor trick (see \cref{sec:comp_preliminary}), we prove the existence of almost-linear time algorithms for \cref{prob:ADiTGC} in the next theorem.
    Let $\vect(W)\coloneqq\underline{W}$ for any matrix $W$ following \cref{def:vectorization}.

\begin{theorem}[Existence of Almost-Linear Time Algorithms for \textsc{ADiTGC}]
\label{thm:main_comp_back}
Suppose all numerical values are in $\calO(\log L)$-bits encoding.
Let $\max(\|W_{OV}A_3\|_{\max},\norm{W_KA_1}_{\max},\norm{W_QA_2}_{\max})\le\Gamma$.
There exists a $L^{1+o(1)}$ time algorithm to solve $\textsc{ADiTGC}(L_p,L, d=
\calO(\log L), \Gamma=o(\sqrt{\log L}))$  (i.e., \cref{prob:ADiTGC}) with loss $\calL_0$ from \cref{def:generic_dit_loss} up to $1 / \poly(L)$ accuracy. 
In particular, this algorithm outputs  gradient matrices $\Tilde{G}^{(W)}\in\R^{d\times d}$ such that
$ \big\|\underline{\Tilde{G}}^{(W)}-\pdv{\calL}{\underline{W}}\big\|_{\max} \le 1/\poly (L)$.
\end{theorem}

\begin{proof}[Proof Sketch]
Our proof is built on the key observation that there exist low-rank structures within the DiT training gradients. 
Using the tensor trick \cite{diao2019optimal,diao2018sketching} and computational hardness results of attention \cite{hu2024computational,alman2024fast}, we approximate  DiT training gradients with a series of low-rank approximations and carefully match the multiplication dimensions so that the computation of $\dv{g_2}{\underline{W}}$ forms a chained low-rank approximation. 
We complete the proof by demonstrating that this approximation is bounded by a $1/\poly(L)$ error and requires only almost-linear time.
See \cref{proof:thm:main_comp_back} for a detailed proof.
\end{proof}
\begin{remark}
    We remark that \cref{thm:main_comp_back} is dominated by the relation between $L$ and $d$, hence by the subspace dimension\footnote{See \cref{assumption:1}.} $d_0 = d L$. 
    A smaller $d_0$ makes \cref{thm:main_comp_back} more likely to hold.
\end{remark}

\subsection{Computational Limits of Forward Inference}
\label{sec:comp_{i_0}nf}
Since the inference of score-matching diffusion models is a forward pass of the trained score estimator $s_W$,
the computational hardness of DiT ties to the transformer-based score network,
\begin{align}\label{eqn:inference_s}
    s_W(A_1,A_2,A_3)=W_AR^{-1}\big( \underbrace{W_{OV} A_3}_{d\times L}\underbrace{D^{-1}}_{L\times L}\exp\big(\underbrace{A_1^\top W_K^\top}_{L\times s} \underbrace{W_QA_2}_{s\times L}\big)\big),
\end{align}
following notation in \cref{def:generic_dit_loss}.
For inference, we study the following approximation problem.
Notably, by \cref{remark:generic}, \eqref{eqn:inference_s} subsumes both conditional and unconditional DiT inferences.
\begin{problem}[Approximate DiT Inference  $\textsc{ADiTI}(d,L,\Gamma,\delta_F)$]
\label{prob:ADiTI}
    Let $\delta_F > 0$ and $B>0$. 
    Given  $A_1,A_2,A_3\in\R^{d\times L}$, and $W_{OV},W_K,W_Q\in\R^{d\times d}$ with guarantees that $\norm{W_{OV}A_3}_\infty\le B$, $\norm{W_KA_1}_\infty\le B$ and $\norm{W_QA_2}_\infty\le B$,
    we aim to study an approximation problem $\textsc{ADiTI}(d,L,B,\delta_F)$, that approximates $s_W(A_1,A_2,A_3)$ with a vector $\Tilde{z}\in\R^{d_0}$ (with $d_0=d\cdot L$) such that
        $\norm{\Tilde{z} - W_AR^{-1}\(W_{OV}A_3D^{-1}\exp(A_1^\top W_K^\top W_QA_2)\)}_{\max} \leq \delta_F$.
   Here, $\norm{A}_{\max}\coloneqq\max_{i,j} \abs{A_{ij}}$ for any matrix $A$. 
\end{problem}
By \hyperref[item:O2]{(O2)} and \hyperref[item:O3]{(O3)}, we make an observation that \cref{prob:ADiTI} is just a special case of \cite{alman2024fast}.
Hence, we characterize the all possible efficient algorithms for $\textsc{ADiTI}$ with next proposition.
\begin{proposition}[Norm-Based Efficiency Phase Transition]
\label{thm:forward_efficiency_threshold}
Let $\norm{W_QA_2}_\infty\le B$, $\norm{W_KA_1}_\infty\le B$ and $\norm{W_{OV}A_3}_\infty\le B$ with $B=\calO(\sqrt{\log L})$.
Assuming SETH (\cref{hyp:seth}), for every $q>0$, there are constants $C,C_a,C_b>0$ such that: there is no $O(n^{2-q})$-time (sub-quadratic) algorithm for the problem $\textsc{ADiTI}(L,d = C \log L,B= C_b \sqrt{\log L},\delta_F = L^{-C_a})$.
\end{proposition}
\begin{remark}
\cref{thm:forward_efficiency_threshold} suggests an efficiency threshold for the upper bound of $\norm{W_KA_1}_\infty$, $\norm{W_QA_2}_\infty$, $\norm{W_{OV}A_3}_\infty$.
Only below this threshold are efficient algorithms for \cref{prob:ADiTI} possible.
\end{remark}
Moreover,
there exist almost-linear  DiT inference algorithms following \cite{alman2024fast}.
\begin{proposition}[Almost-Linear Time DiT Inference]\label{prop:nearly_linear_inf}
Assuming SETH, the DiT inference problem $\textsc{ADiTI}(L,d=\calO(\log  L),B=o(\sqrt{\log  L}),\delta_F=1/\poly ( L))$
can be solved in  $L^{1+o(1)}$ time.
\end{proposition}
\begin{remark}
    \cref{prop:nearly_linear_inf} is a special case of \cref{thm:forward_efficiency_threshold} under the efficiency threshold.
\end{remark}
\begin{remark}
\cref{prop:nearly_linear_inf,thm:forward_efficiency_threshold} are dominated by the relation between $L$ and $d$, hence by the subspace dimension $d_0 = d L$. 
    A smaller $d_0$ makes \cref{prop:nearly_linear_inf,thm:forward_efficiency_threshold} more likely to hold.
\end{remark}

\section{Discussion and Concluding Remarks}
\label{sec:conclusion}
We explore the fundamental limits of latent DiTs with 3 key contributions. 
First, we prove that transformers are universal approximators for the score functions in DiTs (\cref{theorem:1}), with approximation capacity and model size dependent only on the latent dimension, suggesting DiTs can handle high-dimensional data challenges. 
Second, we show that Transformer-based score estimators converge to the true score function (\cref{corollary:score_est}), ensuring the generated data distribution closely approximates the original (\cref{corollary:dist_est}). 
Third, we provide provably efficient criteria (\cref{thm:forward_efficiency_threshold}) and prove the existence of almost-linear time algorithms for forward inference (\cref{prop:nearly_linear_inf}) and  backward computation (\cref{thm:main_comp_back}). 
Our computational results hold for both unconditional and conditional generation of DiTs (\cref{remark:generic}).
These results highlight the potential of latent DiTs to achieve both computational efficiency and robust performance in practical scenarios.

\paragraph{Practical Guidance from Computational Results.}
\cref{sec:comp} analyzes the computational feasibility and identifies all possible efficient DiT algorithms/methods for both forward inference and backward training. These results provide practical guidance for designing efficient methods:

\begin{itemize}
    \item The latent dimension should be sufficiently small: $d=\calO(\log L)$ (\cref{thm:main_comp_back},  \cref{thm:forward_efficiency_threshold,prop:nearly_linear_inf}).
    \item Normalization of $K$, $Q$, and $V$ in DiT attention heads enhances performance and efficiency:
    \begin{itemize}
        \item For efficient inference: $\max{\left\{\| W_KA_1\|,\|W_QA_2\|,\|W_{OV}A_3\|\right\}} \leq B$ with $B=o(\sqrt{\log L})$ (\cref{prop:nearly_linear_inf}) and $A_1,A_2,A_3$ being the input data associated with $K,Q,V$.
        
        \item For efficient training: $\max{\left\{\|W_KA_1\|,\|W_QA_2\|,\|W_{OV}A_3\|\right\}} \leq \Gamma$ with $\Gamma=o(\sqrt{\log L})$ (\cref{thm:main_comp_back}).
    \end{itemize}
\end{itemize}
We remark that these conditions are necessary but not sufficient; sufficient conditions depend on the specific design of the methods used.
This is due to the best- or worst-case nature of  hardness results.

\paragraph{Limitations and Future Direction.}
As discussed in \cref{rk:neg_result}, the double exponential factor in our explicit sample complexity bound (\cref{corollary:score_est}) suggests a possible gap in our understanding of transformer universality and its interplay with DiT architecture.
This motivates us to rethink transformer universality and explore new proof techniques for DiTs, which we leave for future work.
Besides, due to its formal nature, this work does not provide immediate practical implementations. 
However, we expect that our findings provide valuable insights for future diffusion generative models.

\paragraph{Post-Acceptance Note [October, 29, 2024].}
During preparation of the camera-ready version, we learn of a follow-up work \cite{anonymous2024on} that alleviates the double exponential factor and achieves minimax optimal statistical rates for DiTs under Hölder smoothness data assumptions.

\clearpage
\section*{Broader Impact}
This theoretical work aims to shed light on the foundations of diffusion generative models and is not anticipated to have negative social impacts.

\section*{Acknowledgments}
JH would like to thank to Minshuo Chen, Sophia Pi, Yi-Chen Lee, Yu-Chao Huang, Yibo Wen, Damien Jian, Jialong Li, Zijia Li, Tim Tsz-Kit Lau, Chenwei Xu, Dino Feng and Andrew Chen for enlightening discussions on related topics, and the Red Maple Family for support.
The authors would like to thank the anonymous reviewers and program chairs for constructive comments.

JH is partially supported by the Walter P. Murphy Fellowship.
HL is partially supported by NIH R01LM1372201, AbbVie and Dolby.
The content is solely the responsibility of the authors and does not necessarily represent the official
views of the funding agencies.

\bibliographystyle{plainnat}
\bibliography{refs}

\newpage  %
\normalsize
\titlespacing*{\section}{0pt}{*1}{*1}
\titlespacing*{\subsection}{0pt}{*1.25}{*1.25}
\titlespacing*{\subsubsection}{0pt}{*1.5}{*1.5}
\setlist[itemize]{leftmargin=1em}
\setlist[enumerate]{leftmargin=1.4em}

\setlength{\abovedisplayskip}{10pt}
\setlength{\abovedisplayshortskip}{10pt}
\setlength{\belowdisplayskip}{10pt}
\setlength{\belowdisplayshortskip}{10pt}

\part*{Appendix}
\appendix	
{
\setlength{\parskip}{0.3em}
\startcontents[sections]
\printcontents[sections]{ }{1}{}
}
\label{sec:append}
\clearpage
\section{More Discussions on Low-Dimensional Linear Latent Space}

Our analysis is based on the low-dimensional linear latent space assumption (\cref{assumption:1}).
Here we further discuss this in light of our theoretical results

Our results are more general and extend beyond \cref{assumption:1}. 
In addition to the case where $d_0 < D$, our theoretical results apply to two other settings: $d_0 = D$ and $d_0 > D$.
Especially, for $d_0=D$, our  results still hold by
setting $B$ as the identity matrix $I_{D}$.
Namely, our  results hold after removing the linear subspace assumption.

\begin{itemize}
    \item Statistically, for score approximation, score estimation, and distribution estimation, the upper bounds depend on the dimension of the latent variable $d_0$, other than $d$.
    A smaller $d_0$ allows for a reduced model size to achieve a specified approximation error compared to a larger one (\cref{theorem:1}).
    Additionally, with a smaller $d_0$, both score and distribution estimation errors are reduced relative to scenarios with larger ones (\cref{corollary:score_est} and \cref{corollary:dist_est}).
    \item Computationally, smaller $d_0$ benefits the provably efficient criteria (\cref{thm:forward_efficiency_threshold}, almost-linear time algorithms for forward inference (\cref{prop:nearly_linear_inf}) and  backward computation (\cref{thm:main_comp_back}).
\end{itemize}

\clearpage
\section{Notation Table}
We summarize our notations in the following table for easy reference.

\begin{table}[h]
    \caption{Mathematical Notations and Symbols}
    \centering
    \resizebox{ \textwidth}{!}{ 
    \begin{tabular}{ll}
    \toprule
        Symbol & Description \\
    \midrule
        $\norm{z}_2$ & Euclidean norm, where $z$ is a vector\\
        $\norm{z}_{\infty}$ & Infinite norm, where $z$ is a vector\\
        $\norm{Z}_2$ & 2-norm, where $Z$ is a matrix\\
        $\norm{Z}_{\rm op}$ & Operator norm, where $Z$ is a matrix\\
        $\norm{Z}_F$ & Frobenius norm, where $Z$ is a matrix\\
        $\norm{Z}_{p,q}$ & $p,q$-norm, where $Z$ is a matrix\\
        $\norm{f(x)}_{L^2}$ & $L^2$-norm, where $f$ is a function\\
        $\norm{f(x)}_{L^2(P)}$ & $L^2(P)$-norm, where $f$ is a function and $P$ is a distribution\\
        $\norm{f(\cdot)}_{Lip}$ & Lipschitz-norm, where $f$ is a function\\
        $f_{\sharp} P$ & Pushforward measure, where $f$ is a function and $P$ is a distribution\\
    \midrule
        $n$ & Sample size\\
        $x$ & Data point in original data space, $x\in\mathbb{R}^{D}$\\
        $h$ & Latent variable in low-dimensional subspace, $h\in\mathbb{R}^{d_0}$\\
        $p_h$ & The destiny function of $h$\\
        $B$ & The matrix with orthonormal columns to transform $h$ to $x$, where $B \in \RR^{D\times d_0}$\\
        $\Bar{x}$ & Perturbed data variable at $t>0$ \\
        $\Bar{h}$ & $\Bar{h}=B^\top  \Bar{x} $\\
    \midrule
        $T$ & Stopping time in the forward process of Diffusion model\\
        $T_0$ & Stopping time in the backward process of Diffusion model\\
        $\mu$ & Discretized step size in backward process\\
        $p_t(\cdot)$ & The density function of $x$ for at time $t$\\
        $p_t^{h}(\cdot)$ & The density function of $\Bar{h}$ at time $t$\\
        $\psi$ & (Conditional) Gaussian density function\\
    \midrule
        $d$ & Input dimension of each token in the transformer network of DiT\\
        $L$ & Token length in the transformer network of DiT\\
        $X$ & Sequence input of transformer network in DiT, where $X \in \mathbb{R}^{d \times L}$\\
        $E$ & Position encoding, where $E \in \mathbb{R}^{d \times L}$\\
        $R(\cdot)$ & Reshape layer in DiT, $R(\cdot): \mathbb{R}^{d_0} \to \mathbb{R}^{d \times L}$\\
        $W_B$ & The orthonormal matrix to approximate $B$, where $W_B \in \RR^{D\times d_0}$\\
    \bottomrule
    \end{tabular}
    }
    \label{tab:nomenclature}
\end{table}

\clearpage
\section{Related Works}
\label{sec:related_works}

\paragraph{Organization.}
In the following, we first discuss recent developments in DiTs.
Then, we discuss the main technique of our statistical results: the universality (universal approximation) of transformer.
Next, we discuss recent theoretical developments in diffusion generative models.
Lastly, we discuss other aspects of transformer in foundation models beyond diffusion models.

\paragraph{Diffusion Transformers.}
Diffusion \cite{ho2020denoising} and score-based generative models \cite{song2019generative} have been particularly successful as generative models of images, video and biomedical data \cite{nichol2021glide,ramesh2022hierarchical,liu2024sora,zhou2024decompopt,zhou2024antigen,wang2024diffusion,wang2024protein}.
Recently, transformer-based diffusion models have garnered significant attention in research. 
The U-ViT model \cite{bao2022all} incorporates transformer blocks into a U-net architecture, treating all inputs as tokens.
In contrast, DiT \cite{peebles2023scalable} utilizes a straightforward, non-hierarchical transformer structure. 
Empirically, diffusion transformers (DiTs) \cite{peebles2023scalable} have emerged as a significant advancement (e.g., SoRA \cite{SoRA_2024,liu2024sora} from OpenAI), effectively combining the strengths of transformer architectures and diffusion-based approaches. 
Models like MDT \cite{gao2023masked} and MaskDiT \cite{zheng2023fast} improve the training efficiency of DiT by applying a masking strategy.

\paragraph{Universality and Memory Capacity of Transformers.}
The universality of transformers refers to their ability to serve as universal approximators. 
This means that transformers theoretically models any sequence-to-sequence function to a desired degree of accuracy.
\citet{yun2019transformers} establish that transformers can universally approximate sequence-to-sequence functions
by stacking numerous layers of feed-forward functions and self-attention functions.
In a different approach, \citet{jiang2023approximation} affirm the universality of transformers by utilizing the Kolmogorov-Albert representation Theorem. 
Most recently, \citet{kajitsuka2023transformers} show that transformers with one self-attention layer is a universal approximator.

The memory capacity of a transformer is a practical measure to test the theoretical results of the transformer's universality, by ensuring the model can handle necessary context and dependencies.
By memory capacity, we refer to the minimal set of parameters such that the model (i.e., transformer) approximates all input-output pairs in the training dataset with a bounded error.
Several works address the memory capacity of transformers. 
\citet{kim2022provable} show that transformers with $\Tilde{O}(d+L+\sqrt{NL})$ parameters are sufficient to memorize $N$ length-$L$ and dimension-$d$ sequence-to-sequence data points by constructing a contextual mapping with $\calO(L)$ attention layers.
\citet{mahdavi2023memorization} show that a multi-head-attention with $h$ heads is able to memorize $\calO(hL)$ examples under a linear independence data assumption.
\citet{kajitsuka2023transformers} show that a single layer transformer with $\calO(NLd+d^2)$ parameters is able to memorize $N$ length-$L$ and dimension-$d$ sequence-to-sequence data points by utilizing the connection between the softmax function and Boltzmann operator.
\citet{anonymous2024fundamental,wang2024universality} extend the results of \cite{kajitsuka2023transformers,yun2019transformers} to prompt tuning and 
discuss the memorization of the data sequences.
Another line of research establishes a different kind of memory capacity for transformers by connecting transformer attention with dense associative memory models (modern Hopfield models) \cite{hu2024outlier,hu2024computational,hu2024optimal,hu2023SparseHopfield,wu2024uniform,wu2023stanhop,ramsauer2020hopfield}. 
Notably, they define memory capacity as the smallest number of (length-$L$ and dimension-$d$) data points the model (transformer attention) is able to store and derive exponential-in-$d$ high-probability capacity lower bounds.
In particular, \citet{hu2024optimal} report a tight exponential scaling of capacity with feature dimension from the perspective of spherical codes.

Our work is motivated by and builds on \cite{yun2019transformers} to bridge the transformer's function approximation ability with data distribution estimation. 
While we do not address the memorization of DiTs (or diffusion models in general), recent studies on dense associative models suggest viewing pretrained diffusion generative models as associative memory models \cite{achilli2024losing,ambrogioni2023search,hoover2023memory}. 
We plan to explore this aspect in future work.

\paragraph{Theories of Diffusion Models.}
In addition to empirical success, there has been several theoretical analysis about diffusion models \cite{chen2024overview, tang2024score}.
\citet{chen2023score} studies score approximation, estimation, and distribution recovery of U-Net based diffusion models.
\citet{benton2023linear} provide convergence bounds linear in data dimensions, assuming accurate score function approximation.
\citet{zhu2024sample,wibisono2024optimal} provide statistical sample complexity bounds for score-matching under the similar assumptions. 
\citet{oko2023diffusion} analyze the distribution estimation under the assumption that the initial density is supported on $[-1,1]^{D}$ and smooth in the boundary.

Among these works, our work is built on and closest to \cite{chen2023score}, as both assume the data has a low-dimensional structure.
However, our work differs in three key aspects. 
First, beyond the simple ReLU networks considered in \cite{chen2023score}, we provide the first score approximation analysis for DiTs with a transformer-based score estimator. 
Second,  our work is the first to provide the statistical rates of DiTs (score and distribution estimation) based on transformer universality \cite{yun2019transformers} and norm-based converging number bound \cite{Benjamin2020transformer_covering_number}, supporting the practical success of DiTs \cite{esser2024scaling, ma2024sit}.
Lastly, our work provides the first comprehensive analysis of the computational limits and all possible efficient DiT algorithms/methods for both forward inference and backward training. 
This offers timely insights into the empirical computational inefficiency of DiTs \cite{liu2024sora} and guidance for future DiT architectures.

\paragraph{Transformers in Foundation Models: Transformer-Based Pretrained Models.}
Transformer-based pretrained models utilize attention mechanisms to process sequential data, enabling the learning of contextual relationships for tasks like natural language understanding and generation. 
These models encompass three types: encoder-based, decoder-based, and diffusion transformers. 
Encoder-based transformers, such as DNABERT \cite{zhou2024dnabert,zhou2023dnabert,ji2021dnabert}, employ bidirectional attention to extract feature representations 
DNABERT shows great potential to capture complex patterns of genome sequences and improve tasks such as gene prediction.
Decoder-based transformers generate output sequences from encoded information using unidirectional attention, such as  ChatGPT \cite{radford2019language,floridi2020gpt,brown2020language} for natural language.
The diffusion transformers generate a sequence toward a target distribution, such as SoRA \cite{liu2024sora} and Videofusion \cite{luo2023videofusion} for video generation and DecompDiff \cite{guan2024decompdiff} for drug design.
In our paper, we present an early exploration of the statistical and computational limits of diffusion transformer models.

\clearpage
\section{Supplementary Theoretical Background}
In this section, we provide some further background.
We show the details about the forward and backward process in Diffusion Models in \cref{appendix:subsec_diff}.
Besides, we give the details of the proof about the score decomposition in \cref{pf:subspace_score}.

\subsection{Diffusion Models}
\label{appendix:subsec_diff}

\paragraph{Forward Process.}
Diffusion models gradually add noise to the original data in the forward process. 
We describe the forward process as the following SDE
\begin{align}
    \dd x_t=-\frac{1}{2}w(t)x_t\dd t+\sqrt{w(t)}\dd W_t,
    ~
    x_t \in \RR^{D},
    \label{eq:forward_SDE}
\end{align}
where $x_0\sim P_0$, $(W_t)_{t\geq0}$ is a standard Brownian motion, and $w(t)>0$ is a nondecreasing weighting function. 
Let $P_t$ and $p_t$ denote the marginal distribution and destiny of $x_t$.
The conditional distribution $P(x_t|x_0)$ follows $N(\beta(t)x_0, \sigma(t)I_D)$, where $\beta(t)=\exp(-\int_0^t w(s)\dd  s/2)$ and $\sigma(t)=1-\beta^2(t)$. 
In practice, \eqref{eq:forward_SDE} terminates at a large enough $T$ such that $P_T$ is close to $N(0,I_D)$.

\paragraph{Backward Process.}
We obtain the backward process $y_t: =x_{T-t}$ by reversing \eqref{eq:forward_SDE}.
The backward process satisfies
\begin{align}
    \label{eq:backward_SDE}
    \dd y_t = \left[\frac{1}{2}w(T-t)y_t + w(T-t) \nabla \log p_{T-t}(y_t)\right] \dd t + \sqrt{w(T-t)} \dd \Bar{W}_t,
\end{align}
where the score function $\nabla \log p_t(\cdot)$ is the gradient of log probability density function of $x_t$, and $\Bar{W}_t$ is a reversed Brownian motion.
However, $\nabla \log p_t(\cdot)$ and $P_T$ are both unknown in \eqref{eq:backward_SDE}. 
To resolve this, we use a score estimator $s_W(\cdot,t)$ to replace $\nabla\log p_t(\cdot)$, where $s_W(\cdot,t)$ is usually a neural network with parameters $W$.
Secondly, we replace $P_T$ by the standard Gaussian distribution. 
Consequently, we obtain the following SDE
\begin{align}
    \dd \tilde{y}_t = \left[\frac{1}{2}w(T-t) \tilde{y}_t + w(T-t) s_W(\tilde{y}_t, T-t)\right] \dd t + \sqrt{w(T-t)} \dd \Bar{W}_t, 
    ~ \tilde{y}_0 \sim N(0, I_D).
    \label{eq:backward_practice}
\end{align}
In practice, we use discrete schemes of \eqref{eq:backward_practice} to generate data, following \cite{song2019generative}.
We use $\mu>0$ to denote the discretization step size.
For $t \in[k \mu,(k+1) \mu]$, we have
\begin{align}
    \dd \tilde{y}_t^{\mu} = \left[\frac{1}{2}w(T-t) \tilde{y}_{k\mu}^{\mu} + w(T-t) s_W(\tilde{y}_{k\mu}^{\mu}, T-k\mu)\right] \dd t + \sqrt{w(T-t)} \dd \Bar{W}_t.
    \label{eq:dis_backward_practice}
\end{align}

\subsection{Proof of \texorpdfstring{\cref{lemma:subspace_score}}{}}
Here we restate the proof of \cite[Lemma~1]{chen2023score} for 
 completeness.

\begin{proof}
\label{pf:subspace_score}
Recall $x=Bh$ by \cref{assumption:1} with  $x\in\R^D$, $B\in\R^{D\times d_0}$ and $h\in\R^{d_0}$.

By the forward process \eqref{eq:forward_SDE}, we have
\begin{align}
\label{eq:x_real_distribution}
p_t( \Bar{x} ) = \int \psi_t( \Bar{x}  \mid Bh) p_h(h) \dd h,
\end{align}
where 
\begin{align}\label{eqn:gaussian_kernel}
    \psi_t( \Bar{x} \mid Bh) = [2\pi h(t)]^{-D/2} \exp\left(-\frac{\norm{\beta(t)Bh - \Bar{x} }_2^2}{2\sigma(t)} \right),
\end{align}
is the Gaussian transition kernel.

Then we write the score function as
\begin{align}
\label{eq:nabla_pt}
\nabla \log p_t( \Bar{x} ) &= \frac{\nabla p_t( \Bar{x} )}{p_t( \Bar{x} )} 
\\
&= \frac{\nabla \int \psi_t( \Bar{x}  \mid Bh) p_h(h) \dd h}{\int \psi_t( \Bar{x}  \mid Bh) p_h(h) \dd  h} \annot{By pluging in $p_t( \Bar{x} )$} 
\\
&= \frac{\int \nabla \psi_t( \Bar{x}  \mid Bh) p_h(h) \dd  h}{\int \psi_t( \Bar{x}  \mid Bh) p_h(h) \dd  h}, \annot{By interchanging $\int$ with $\nabla$}
\end{align}
where the last equality holds since $\psi_t( \Bar{x}  \mid Bh)$ is continuously differentiable in $ \Bar{x} $. 

Plugging \eqref{eqn:gaussian_kernel} into \eqref{eq:nabla_pt}, we have
\begin{align*}
& ~ \nabla \log p_t( \Bar{x} ) \\
= &~ \frac{[2\pi h(t)]^{-D/2}}{\int \psi_t( \Bar{x}  \mid Bh) p_h(h) \dd  h} \int \frac{1}{\sigma(t)} \left(\beta(t)Bh -  \Bar{x} \right) \exp\left(-\frac{\norm{\beta(t)B h -  \Bar{x} }_2^2}{2\sigma(t)} \right) p_h(h) \dd  h. 
\end{align*}

We then decompose above score function by projecting of $\Bar{x}$ into ${\rm Span}(B)$, i.e., 
replacing $- \Bar{x} $ with $-BB^\top  \Bar{x}  - (I_D - BB^\top) \Bar{x} $:
\begin{align*}
& ~ \nabla \log p_t( \Bar{x} ) \\
= & ~ 
\frac{[2\pi h(t)]^{-D/2}}{\int \psi_t( \Bar{x}  \mid Bh) p_h(h) \dd  h} \\
& ~
\cdot \int \frac{1}{\sigma(t)} \Bigg[\left(\beta(t)Bh - BB^\top  \Bar{x} \right) - \left(I_D - BB^\top\right) \Bar{x} \Bigg]  \exp\left(-\frac{\norm{\beta(t)B h -  \Bar{x} }_2^2}{2\sigma(t)} \right) p_h(h) \dd  h . 
\end{align*}

Absorbing the factor of $[2\pi h(t)]^{-D/2}$ into the Gaussian kernel $\psi_t( \Bar{x}  \mid Bh)$, we have
\begin{align*}
&~ \nabla \log p_t( \Bar{x} ) \\
= &~ \frac{[2\pi h(t)]^{-D/2}}{\int \psi_t( \Bar{x}  \mid Bh) p_h(h) \dd  h} \int \frac{1}{\sigma(t)} \left(\beta(t)Bh - BB^\top  \Bar{x} \right) \exp\left(-\frac{\norm{\beta(t)B h -  \Bar{x} }_2^2}{2\sigma(t)} \right) p_h(h) \dd  h \\
& ~ 
- \frac{1}{ \int \psi_t( \Bar{x}  | Bh ) p_h(h) \dd  h} \left( \frac{1}{\sigma(t)} \left(I_D - BB^\top\right) \Bar{x}  \right)  \int \psi_t( \Bar{x}  \mid Bh) p_h(h) \dd  h  \\
= &~ \underbrace{\frac{1}{\int \psi_t( \Bar{x} \mid Bh) p_h(h) \dd  h} \int \frac{1}{\sigma(t)} \left(\beta(t)Bh - BB^\top \Bar{x} \right) \psi_t( \Bar{x} \mid Bh) p_h(h) \dd  h}_{\coloneqq s_{+}} 
\underbrace{-  \frac{1}{\sigma(t)}\left(I_D - BB^\top\right)  \Bar{x} }_{\coloneqq s_{-}}.
\end{align*}

To further simplify $s_{+}$, we decompose $\psi_t( \Bar{x} \mid Bh)$ as
\begin{align}
& ~ \psi_t( \Bar{x} \mid Bh) 
\nonumber\\
= &~ [2\pi h(t)]^{-D/2} \exp\left(-\frac{1}{2\sigma(t)}\norm{\beta(t)Bh -  \Bar{x} }_2^2 \right) 
\nonumber\\
= &~ [2\pi h(t)]^{-D/2} \exp\left(-\frac{1}{2\sigma(t)}\norm{\beta(t)Bh - BB^\top  \Bar{x}  - \left(I_D - BB^\top \right) \Bar{x} }_2^2 \right) 
\nonumber\\
= &~ [2\pi h(t)]^{-D/2} \exp\Bigg(-\frac{1}{2\sigma(t)}\Big(\norm{\beta(t)B h - BB^\top  \Bar{x} }_2^2 + \norm{\left(I_D - BB^\top \right) \Bar{x} }_2^2 
\nonumber\\
& \quad\quad\quad\quad\quad\quad\quad\quad\quad\quad\quad\quad\quad  
- 2(B(\beta(t) h - B^\top \Bar{x}))^\top (I_D - BB^\top) \Bar{x} \Big) \Bigg) \nonumber\\
= & ~ [2\pi h(t)]^{-D/2} \exp\left(-\frac{1}{2\sigma(t)}\left(\norm{\beta(t)B h - BB^\top  \Bar{x} }_2^2 + \norm{\left(I_D - BB^\top \right) \Bar{x} }_2^2\right) \right) 
\annot{$B(\beta(t) h - B^\top \Bar{x})$ is in ${\rm Span} (B)$ while $(I_D - BB^\top)\Bar{x}$ is orthogonal to ${\rm Span} (B)$}
\\
= & ~ 
\underbrace{[2\pi h(t)]^{-d_0/2} \exp\left(-\frac{\norm{\beta(t) h - B^\top  \Bar{x} }_2^2}{2\sigma(t)}\right)
}_{\coloneqq \psi_t\left(B^\top  \Bar{x}  \mid h\right)}
\cdot 
\underbrace{[2\pi h(t)]^{-(D-d_0)/2}  \exp\left(-\frac{\norm{\left(I_D - BB^\top \right) \Bar{x} }_2^2}{2\sigma(t)}\right)}_{\coloneqq\psi_t\left((I_D - BB^\top) \Bar{x} \right)},
\annot{since $B$ has orthonormal columns}
\label{eqn:s_+}
\end{align}
where both $\psi_t\left(B^\top  \Bar{x}  \mid h\right) $ and $\psi_t\left((I_D - BB^\top) \Bar{x} \right)$ are Gaussian.

Plugging $\psi_t( \Bar{x}  \mid Bh) = \psi_t\left(B^\top \Bar{x}  \mid h\right) \psi_t\left((I_D - BB^\top) \Bar{x} \right)$ into $s_{+}$, we obtain
\begin{align*}
s_{+}( \Bar{x} , t) &= C \int \frac{1}{\sigma(t)} \left(\beta(t)Bh - BB^\top  \Bar{x} \right) \psi_t(B^\top  \Bar{x} \mid h) \psi_t((I_D - BB^\top) \Bar{x}) p_h(h) \dd  h \\
&= C \psi_t((I_D - BB^\top) \Bar{x}) \int \frac{1}{\sigma(t)} \left(\beta(t)Bh - BB^\top  \Bar{x} \right) \psi_t(B^\top  \Bar{x}  \mid h) p_h(h) \dd  h 
\\
&= \frac{1}{\int \psi_t(B^\top  \Bar{x}  \mid h) p_{h}(h) \dd  h} \int \frac{1}{\sigma(t)} \left(\beta(t)Bh - BB^\top  \Bar{x} \right) \psi_t(B^\top \Bar{x} \mid h) p_h(h) \dd  h,
\end{align*}
where $C \coloneqq [\psi_t((I_D - BB^\top) \Bar{x}) \int \psi_t(B^\top  \Bar{x}  \mid h) p_{h}(h) \dd  h]^{-1}$.

Notably, $s_{+}$  depends only on the projected data $B^\top  \Bar{x} $. 
Therefore, we are able to replace $s_{+}( \Bar{x} , t)$ with $s_{+}(B^\top \Bar{x} , t)$. 
The benefit is that the dimension $d_0$ of the first input in $s_{+}(B^\top \Bar{x}, t)$ is much smaller. 

Lastly, by denoting $\Bar{h} = B^\top  \Bar{x} $ such that $\nabla_{\Bar{h}} \psi_t(\Bar{h} \mid h) =  (\beta(t) h - \Bar{h} ) \psi_t(\Bar{h}  \mid h)/\sigma(t)$, we arrive at
\begin{align*}
s_{+}(B^\top  \Bar{x} , t) & =  B \int\frac{\nabla_{\Bar{h}}\psi_t(\Bar{h} \mid h) p_h(h)}{\int \psi_t(\Bar{h} \mid h) p_{h}(h) \dd  h} \dd  h \\
& = B \nabla \log p_t^{h}(B^\top  x ).
\annot{$p_t^{h}(\Bar{h}) \coloneqq  \int  \psi_t(\Bar{h}|h)p_h(h) \dd h$}
\end{align*}
This completes the proof.
\end{proof}

\subsection{Preliminaries: Strong Exponential Time Hypothesis (SETH) and Tensor Trick}
\label{sec:comp_preliminary}

Here we present the ideas we built upon for \cref{sec:comp}.

\paragraph{Strong Exponential Time Hypothesis (SETH).}
\citet{ip01} introduce the Strong Exponential Time Hypothesis (SETH) as a stronger form of the $\mathtt{P} \neq \mathtt{NP}$ conjecture.
It suggests that our current best $\mathtt{SAT}$ algorithms are optimal and is a popular conjecture for proving fine-grained lower bounds for a wide variety of algorithmic problems \cite{cygan2016problems,williams2018some}.
\begin{hypothesis}[SETH]
\label{hyp:seth}
For every $\epsilon > 0$, there is a positive integer $k \geq 3$ such that $k$-$\mathtt{SAT}$ on formulas with $n$ variables cannot be solved in $\calO(2^{(1-\epsilon )n})$ time, even by a randomized algorithm.
\end{hypothesis}

\paragraph{Tensor Trick for Computing Gradients.}
The tensor trick \cite{diao2019optimal,diao2018sketching} is an instrument to compute complicated gradients in a clean and tractable fashion.
We start with some definitions.
\begin{definition}[Vectorization]
\label{def:vectorization}
    For any matrix $X\in\R^{L\times d}$, we define $\underline{X}\coloneqq \vect{(X)}\in \R^{Ld}$ such that $X_{i,j}=\underline{X}_{(i-1)d+j}$ for all $i\in[L]$ and $j\in[d]$.
\end{definition}
\begin{definition}[Matrixization]
\label{def:matrixization}
    For any vector $\underline{X}\in\R^{Ld}$, 
    we define $\mathrm{mat}(\underline{X})=X$ such that
    {$X_{i,j}=\mathrm{mat}(\underline{X})\coloneqq \underline{X}_{(i-1)d+j}$} for all $i\in[L]$ and $j\in[d]$, namely $\mathrm{mat}(\cdot)=\vect^{-1}(\cdot)$. 
\end{definition}

\begin{definition}[Kronecker Product]
\label{def:kronecher_prod}
    Let $A\in\R^{L_a\times d_a}$ and $B\in\R^{L_b\times d_b}$.
We define the Kronecker product of $A$ and $B$ as $A\otimes B\in \R^{L_aL_b\times d_a d_b}$ such that 
$(A\otimes B)_{(i_a-1)L_b+i_b, (j_a-1)d_b+j_b}$,
is equal to $A_{i_a,j_a}B_{i_b,j_b}$ with 
$i_a\in[L_a],j_a\in[d_a],i_b\in[L_b],j_b\in[d_b]$.
\end{definition}

\begin{definition}[Sub-Block of a Tensor]
\label{def:subblock}
    For any $A\in\R^{L_a\times d_a}$ and $B\in\R^{L_b\times d_b}$, let $\A\coloneqq A\otimes B \in \R^{L_aL_b\times d_ad_b}$.
    For any $\underline{j}\in[L_a]$, we define $\A_{\underline{j}}\in\R^{L_b\times d_ad_b}$ be the $\underline{j}$-th $L_b\times d_ad_b$ sub-block of $\A$.
\end{definition}
\begin{lemma}[Tensor Trick \cite{diao2019optimal,diao2018sketching}]
\label{lemma:tensor_trick}
    For any $A\in\R^{L_a\times d_a}$, $B\in\R^{L_b\times d_b}$ and $X\in\R^{d_a\times d_b}$,
    it holds $\vect\(A^\top XB\)=(A^\top \otimes B^\top) \underline{X}\in \R^{L_aL_b}$.
\end{lemma}
To showcase the tensor trick, let's consider a (single data point) attention following \cite{gao2023fast,gao2023context}.
Setting $D\coloneqq \diag\(\exp(X^\sT W_K^\sT  W_Q X)\one_L\)$ and $W\coloneqq W_KW_Q^\sT\in\R^{d\times d}$, we have
\bea 
\calL_{0}\coloneqq 
\big\| 
\underbrace{W_V}_{d\times d}\underbrace{X}_{\in\R^{d\times L}} 
\underbrace{D^{-1}}_{\in \R^{L\times L}} 
\underbrace{\exp{X^\sT W X}
}_{\in\R^{L\times  L}}
 - \underbrace{Y}_{\in\R^{d\times L}} 
\big\|_{2}^2.
\eea 
\begin{proposition}[Definition~4.7 of \cite{gao2023fast}]\label{prop:tensor_loss}
By \cref{def:kronecher_prod} and \cref{def:subblock},
we identify $D_{\underline{j},\underline{j}}\coloneqq\Braket{\exp(\A_{\underline{j}}\underline{W}),\one_L}\in\R$ for all $\underline{j}\in[L]$, with $\A\coloneqq X\otimes X\in\R^{L^2\times d^2}$ and $\underline{W}\in\R^{d^2}$.
Therefore, for each $\underline{j}\in[L]$ and $\underline{i}\in [d]$, it holds 
$\calL_0= \sum_{\underline{j}=1}^L\sum_{\underline{i}=1}^d 
\half \(\Braket{D^{-1}_{\underline{j},\underline{j}}\exp(\A_{\underline{j}}\underline{W}), XW_V[\cdot, \underline{i}]}-Y_{\underline{j},\underline{i}}\)^2$.
\end{proposition}
The elegance of \cref{prop:tensor_loss} emerges when we vectorize the weights into vectors $\underline{W},\underline{W}_V$, making the gradient computations (e.g., $\nicefrac{\dd \calL_0}{\underline{W}}$ and $\nicefrac{\dd \calL_0}{\underline{W}_V}$) more tractable by avoiding complex matrix or tensor derivatives. 
This approach systematically simplifies the handling of chain-rule terms in the gradient computation of losses like $\calL_0$.

\paragraph{Fine-Grained Complexity for Transformer.}
Many recent works also utilize similar techniques from fine-grained complexity to analyze transformer architectures.
\citet{alman2024fast, as24_neurips,lssz24_tat,as23_tensor} explore the computational feasibility of inference and training for standard softmax and tensor attention.
\citet{liang2024multi} extend the single-layer training results from \cite{as24_neurips} to deep transformer models.
\cite{lls+24_conv}  extend \cite{as24_neurips} to provide a fast attention gradient approximation based on Fourier transform.
\cite{lls+24_prune} extend \cite{as24_neurips} to  sparse attention matrix.
\citet{anonymous2024fundamental} study the computational limits of inference and training in prompt-tuning for pretrained transformers.
\citet{hu2024computationalLo} study the computational limits of LoRA \cite{hu2021lora} in transformers, identifying norm-bound conditions for efficient LoRA training and proving the existence of nearly linear-time LoRA algorithms.

Our work is closest to \cite{as24_neurips, alman2024fast}. Our forward inference computational results build on \cite{alman2024fast}. Our backward training computational results are related to \cite{as24_neurips} but include additional analysis on reshaping and latent embedding.

\clearpage
\section{More Background and Auxiliary Lemmas: Universal Approximation of Transformers via Piecewise Approximation}
\label{apendix_sec:transformer_approximation}
Here, we review the universal approximation of transformers following \cite{yun2019transformers}. 

Our goal is to reproduce the  results of \cite{yun2019transformers} and use or modify them as auxiliary lemmas for proofs of \cref{sec:method}  (i.e., \cref{sec:pf_thm1}.)

We start with their central result and prove it in the rest of the section.

\begin{lemma}[Universal Approximation of Transformers, Theorem 3 of \cite{yun2019transformers}]
\label{transformer_approximation}
Let $\epsilon>0$. 
For any given compact-supported continuous function $f:\R^{d\times L}\to \R^{d\times L}$, there exists a transformer network $f_{\cT} \in \mathcal{T}_p^{2,1,4}$, such that 
\begin{align*}
\left(\int \norm{f_{\cT}(X) -f(X)}_F^2 \dd  X\right)^{1/2} \leq\epsilon.
\end{align*}
    
\end{lemma}

\paragraph{Proof Overview.}
We use the following proof strategy:
\begin{itemize}
    \item {\bf Step 1.} We show that the piecewise-constant function is able to approximate compact-supported continuous function in \cref{appendix_sec:subsec_appro_compact_piece}.
    \item  {\bf Step 2.}
    We define modified self-attention and feed-forward layers to construct the modified transformer.
    We show that the modified transformer is able to approximate piecewise-constant function in \cref{appendix_sec:modified_approximation}.
    
    \item  {\bf Step 3.}
    We show that the modified transformer is able to approximate the normal transformer in \cref{appendix_sec:modified_approximation_normal}.
\end{itemize}    

We provide details of  {\bf Step 1.} in \cref{appendix_sec:subsec_appro_compact_piece},   {\bf Step 2.} in \cref{appendix_sec:modified_approximation}, and   {\bf Step 3.} in \cref{appendix_sec:modified_approximation_normal}.
Then we summarize our results in \cref{subsec:summary}.

\subsection{Piecewise-Constant Function Approximates Compact-Supported Continuous Function}
\label{appendix_sec:subsec_appro_compact_piece}
In this subsection, we show that the piecewise-constant function is able to approximate compact-supported continuous function.

We start with the definition of the compact-supported continuous functions of interest.
\begin{assumption}
Without loss of generality, we assume that the target function in discussion is supported on $[0,1]^{d\times L}$.    
We denote the set of $[0,1]^{d\times L}$-supported continuous functions as $\mathcal{F}$. 
\end{assumption}

We introduce the notion of grid and cube for the compact support $[0,1]^{d\times L}$.
\begin{definition}[Grid and Cube with Width $\delta$]
\label{def:grid_cube}
    Given a grid width $\delta$, let $\calG_{\delta} \coloneqq \{0, \delta, \dots, 1-\delta \}^{d\times L}$ denote the set of grids within $[0,1]^{d\times L}$.
    For a grid point $G=(G_{j\in[d],k\in[L]}) \in \calG_{\delta}$, we denote its associated cube as
    \begin{align*}
        \calS_G := \otimes_{j=1}^{d}\otimes_{k=1}^{L}[G_{j,k}, G_{j,k}+\delta) \subset [0,1]^{d\times L}.
    \end{align*}
    Each cube $\calS_G$ represents a hyper rectangular in the multi-dimensional space $[0,1]^{d\times L}$, constructed to discretize the space into smaller subspaces.
\end{definition}

We introduce the notion of piecewise-constant fucntion class w.r.t. the $[0,1]^{d\times L}$-supported continuous function class $\calF$.

\begin{definition}[Piecewise-Constant Function Class]
\label{def:pieceweise_const_func_class}    
    Let $f_{\delta}$ denote the piesewise constant function of grid width $\delta$, and $\one\{\cdot\}$ denote the indicator function.
    For each $G \in \calG_{\delta}$, and any matrix $A_G \in \RR^{d\times L}$, we define the piecewise-constant function class as 
    \begin{align}
    \label{def:piecewise-constant function}
        \calF(\delta) \coloneqq
        \left \{f_{\delta} : X \rightarrow \sum\nolimits_{G \in \calG_{\delta}} A_G \cdot \one\{X \in \calS_G\}, A_G \in \RR^{d\times L} \right \}.
    \end{align}
\end{definition}
 
We recall that for a given sequence-to-sequence function $f$, 
\begin{align*}
    \norm{f}_{L^2}:=\bigg(\int \norm{f(X)}_F^2\dd  X\bigg)^{1/2}.
\end{align*}
We approximate the compact-supported function with a piecewise-constant function in the next lemma.

\begin{lemma}
\label{lemma:part1_piecewise}
(Lemma 8 of \cite{yun2019transformers}) For any given $f \in \mathcal{F}$ and $\epsilon/3>0$, we can find a $\delta^\star > 0$, such that there exists a $f_{\delta^\star} \in \mathcal{F}(\delta^\star)$ satisfying $\norm{f-f_{\delta^\star}}_{L^2}\leq \epsilon/3$.
\end{lemma}
\begin{proof}
    See \cref{subsubsec:pf_piecewise} for a detailed proof.
\end{proof}

\subsection{Modified Transformer Approximates Piecewise-constant Function}
\label{appendix_sec:modified_approximation}
In this subsection, we define modified self-attention and feed-forward layers to construct the modified transformers.
We use the modified transformers to approximate the piecewise-constant function.

\begin{definition}[Modified Transformer Networks]
\label{def:modified_trans}
    The modified transformer network $\Bar{\mathcal{T}}_p^{r,m,l}$ includes two modifications to the standard transformer network $\mathcal{T}_p^{r,m,l}$:
    \begin{itemize}
    \item Modified attention layer: Replace $\Softmax$ operator with $\Hardmax$ operator $\sigma_{H}(\cdot)$. 
    \item Modified feed-forward layer: Replace ${\rm ReLU(\cdot)}$ with an activation function $\zeta\in\Psi$.
    Here, $\Psi$ denotes the set of all piecewise linear functions with at most three pieces and at least one constant. 
    \end{itemize}
\end{definition}

We approximate $\mathcal{F}(\delta)$ with this modified transformer networks $\Bar{\cT}_p^{r,m,l}$.
\begin{lemma}[Modified from Proposition 4 of \cite{yun2019transformers}]
\label{lemma:part2_approx_piecewise_modified}
    For each $f_{\delta}\in \mathcal{F}(\delta)$, there exists a $f_{\cT,c}\in\Bar{\mathcal{T}}_p^{2,1,1}$ such that $\norm{f_{\delta}-f_{\cT,c}}_{L^2}=\mathcal{O}(\delta^{d/2})$.
\end{lemma}
\begin{proof}[Proof Sketch]
    Given $\delta$, and for any grid $G \in \calG_{\delta}$, we have a grid set  $\calG_{\delta}$ and the cube $\calS_{G}$.
    
    Our proof follows two steps:
    \begin{itemize}
        \item \textbf{Quantization.} 
        For all $X \in \RR^{d \times L}$, we quantize it to a finite set:
        \begin{itemize}
            \item If $X \in \calS_{G} \subset [0, 1]^{d \times L}$, we quantize it to the element $G \in \calG_{\delta}$.
            \item If $X \notin [0, 1]^{d \times L}$, we quantize it to an element out of $\calG_{\delta}$.
        \end{itemize}
        \item \textbf{Mapping.}
        For any $G \in \calG_{\delta}$, we map it to the desired output $A_G$.
    \end{itemize}

    For \textbf{Quantization},
    we achieve this by a series of modified feed-forward layers.
    We show this in \cref{subsubsec:quantize}.
    
    For \textbf{Mapping}, we follow two steps:
    \begin{itemize}
        \item For any $G \neq G^\prime \in \calG_{\delta}$, we use a ``contextual mapping'' $q_c(\cdot)$ (defined as \cref{def:contextmapping}). The mapping maps all the elements in $q_c(G)$ and $q_c(G^\prime)$ to different values.
        Then, we use a series of modified self-attention layers to achieve ``contextual mapping''.
        We show this in \cref{subsubsec:context}.
        
        \begin{definition}[Contextual Mapping]
        \label{def:contextmapping}
            Consider a finite set $\calG_{\delta}\in \RR^{d\times L}$. 
            A map $q_c:\calG_{\delta}\rightarrow \RR^{1\times L}$ defines a contextual mapping if the map satisfies the following:
            \begin{itemize}
                \item For any $G\in \calG_{\delta}$, the entries in $q_c(G)$ are all distinct. 
                \item For any $G \neq G^\prime \in \calG_{\delta}$, all entries of $q_c(G)$ and $q_c(G^\prime)$ are distinct. 
            \end{itemize}
        \end{definition}
        \item For any $G \in \calG_{\delta}$, we use a series of modified feed-forward layers to map $q_c(G)$ to $A_G$.
        We show this in \cref{subsubsec:map_to_desired_output}.
    \end{itemize}
\end{proof}
\begin{remark}
    Our proof differs from \cite{yun2019transformers} in one aspect: Although \cite[Proposition 4]{yun2019transformers} outlines a proof for transformer networks without positional encoding and sketches the proof for networks with it, we provide a detailed proof for the latter to support our proof.
\end{remark}

\subsubsection{Quantization by Modified Feed-forward Layers}
\label{subsubsec:quantize}

We use a series of modified feed-forward layers in $\Bar{\mathcal{T}}_p^{r,m,l}$ to quantize an input $X\in \RR^{d\times L}$ to an element $G$ of the following grid: 
\begin{align*}
    \{-J,0,\delta,\dots,1-\delta\}^{d\times L},
\end{align*}
where $J>L>0$ is a large number to be determined later. 
We achieve this via two steps.
\begin{itemize}
    \item {\bf Step 1: Map the element out of $[0,1)$ to $-J$. }
    
    We use {$e_i$} to represent the standard unit vector where the $i$-th element is $1$.
    For the $i$-th row of $X$, we define the following feed-forward layer to achieve our aim.
    \begin{definition}[Feed-forward Layer 1]
    \label{def:ff_1}
    The vector $e_i$ acts as the weight parameters, and $\zeta_1(\cdot)$ acts as the activation function in the feed-forward layer
        \begin{align}
        \label{eq:FF_1}
            X \rightarrow X + e_i \zeta_1( e_i^\top X ),
            ~~
            \zeta_1(t) = 
            \begin{cases}
            -t-J, & \text{for } t < 0 \text{ or } t \geq 1,\\
            0, & \text{otherwise}.
        \end{cases}
    \end{align}
    \end{definition}
    We take $i=1$ as an example to give the specific calculation.
    Let $X=(x_{i,j})_{d\times L}$, then we have
    \begin{align*}
        ~{\rm FF}(X)&=X+
        \begin{pmatrix}
          1 
          \\
          0 
          \\
          \vdots 
          \\
          0
        \end{pmatrix}
        \begin{pmatrix}
          \zeta_1(x_{1,1}) &
          \zeta_1(x_{1,2}) &
          \cdots &
          \zeta_1(x_{1,L})
        \end{pmatrix}
        \\
        & =X+
        \begin{pmatrix}
          \zeta_1(x_{1,1}) & \zeta_1(x_{1,2}) & \cdots & \zeta_1(x_{1,L}) 
          \\
          0 & 0 &\cdots&0
          \\
          \vdots & \vdots &\vdots&\vdots
          \\
          0 & 0 &\cdots&0
        \end{pmatrix}.
    \end{align*}
    In the first row of $X$, the above layer transforms the element that is out of $[0,1)$ to $-J$. 
    
    We stack the above layers together for $i=1,2,\dots,d$.
    If the element of $X$ is out of $[0,1)$, the series of layers maps it to $J$.

    \item {\bf Step 2: Map the element in $[0,1)$ to $\{0, \delta, 2\delta, \dots, 1-\delta\}$. }
    
    For the $i$-th row of $X$, we take 
    $k=0,1,\dots,1/\delta-1$ respectively.
    We define the following layer.
    \begin{definition}[Feed-forward Layer 2] 
    \label{def:ff_2}
    The vector $e_i$ acts as the weight parameters and $\zeta_2(\cdot)$ acts as the activation function in the feed-forward layer
        \begin{align}
        \label{eq:FF_2}
        X \rightarrow X + e_i \zeta_2( e_i^\top X - k\delta \one_n^\top),
        ~~
        \zeta_2(t) = 
        \begin{cases}
        0, & t < 0 \text{ or } t \geq \delta,\\
        -t, & 0 \leq t < \delta.
        \end{cases}
    \end{align}
    \end{definition}
    We take $i=1$ and $k=1$ as an example.
    We give the following specific calculation
    \begin{align*}
        {\rm FF}(X)
        &=X+
        \begin{pmatrix}
          1 
          \\
          0 
          \\
          \vdots
          \\
          0
        \end{pmatrix}
        \begin{pmatrix}
            \zeta_2(x_{1,1}-\delta)
            & \zeta_2(x_{1,2}-\delta) & \cdots & \zeta_2(x_{1,L}-\delta)
        \end{pmatrix}
        \\
        &=X+
        \begin{pmatrix}
            \zeta_2(x_{1,1}-\delta) & \zeta_2(x_{1,2}-\delta) & \cdots & \zeta_2(x_{1,L}-\delta) 
            \\
            0 & 0 & \cdots & 0
            \\
            \vdots & \vdots & \vdots & \vdots
            \\
            0 & 0 &\cdots&0           
        \end{pmatrix}.
    \end{align*}
    In the first row of $X$, the above layer transforms the element in $[\delta, 2\delta]$ to $\delta$. 
    
    We stack the above layers together for $i=1,2,\dots,d$ and $k=0,1,\dots,1/\delta-1$.
    If the element of $X$ is in $[k\delta,(k+1)\delta]$, the series layers maps it to $k\delta$.
\end{itemize}

Combining the above two parts, we achieve our goal with $d/\delta+d$ feed-forward layers.
We denote the $d/\delta+d$ series layers as $f_{\cT,c1}$.

\subsubsection{ Contextual Mapping by Modified Self-attention Layers}
\label{subsubsec:context}
In our attention layers, we use the following positional encoding $E \in \RR^{d \times L}$
\begin{align}
\label{eq:pos_encoding}
    E = 
    \begin{pmatrix}
    0 & 1 & 2 & \cdots & L-1
    \\
    0 & 1 & 2 & \cdots & L-1
    \\
    \vdots & \vdots & \vdots & & \vdots
    \\
    0 & 1 & 2 & \cdots & L-1
    \end{pmatrix}.
\end{align}
According to \cref{subsubsec:quantize}, the output of $f_{\cT,c1}$ is in the grid $\{-J,0,\delta,\dots,1-\delta\}^{d\times L}$.
For any $X$ in this grid, the first column of $X+E$ is in 
\begin{align*}
    \{-J,0,\delta,\dots,1-\delta\}^d,
\end{align*}
and the second column is in
\begin{align*}
    \{-J+1,1,1+\delta,\dots,2-\delta\}^d.
\end{align*}
The results are similar in the other columns.

For $i=0, 1, \dots, L-1$, we use the following notation:
\begin{align*}
    [i:\delta:i+1-\delta]_{J} \coloneqq \{i-J, i, i+\delta, \dots, i+1-\delta\}.
\end{align*}
Then, we define the grid $\calG_{\delta}^+$ as the following.
\begin{definition}[Grid $\calG_{\delta}^+$]
\label{def:grid}
    We add $E$ to all the grid points in $\calG_{\delta}$ to generate the modified grid $\calG_{\delta}^+$, defined as follows:
    \begin{align*} 
    \calG_{\delta}^+ \coloneqq [0:\delta:1-\delta]_{J}^d 
    \times [1:\delta:2-\delta]_{J}^d 
    \times \cdots 
    \times [L-1:\delta:L-\delta]_{J}^d.
    \end{align*}
\end{definition}

Next, we show that the modified attention layer computes contextual mapping (\cref{def:contextmapping}) for $\calG_{\delta}^+$.
For $i=1, 2, \dots, L-1$, we use the following notation:
\begin{align*}
    [i:\delta:i+1-\delta] \coloneqq \{i, i+\delta, i+2\delta, \dots, i+1-\delta\}.
\end{align*}

\begin{lemma}[Modified from Lemma 6 of \cite{yun2019transformers}]
\label{lemma:contextmap}
We consider the following subset of $\calG_{\delta}^+$:
\begin{align*}
\tilde {\calG}_{\delta}
:=\underbrace{[0:\delta:1-\delta]^d\times[1:\delta:2-\delta]^d\times\cdots\times[L-1:\delta:L-\delta]^d}_{L}.
\end{align*}
Assume that $L\geq 2$ and $\delta^{-1}\geq2$. 
Then, there exist a function $f_{\cT,c2}: \RR^{d\times L} \to \RR^{d\times L}$ composed of $\delta^{-d}+1$ modified attention layers (\cref{def:modified_trans}), a vector $u \in \RR^d$, and two constants $t_l, t_r \in \RR$ ($0 < t_l < t_r$), such that $q_c(G) \coloneqq u^\top f_{\cT,c2}(G), G \in \calG_{\delta}^+$ satisfies the following properties:
\begin{enumerate}
    \item \label{cond:1} For any $G \in \tilde{\calG}_{\delta}$, all the entries of $q_c(G)$ are distinct.
    
    \item \label{cond:2} For any different $G, G^\prime \!\in\! \tilde{\calG}_{\delta}$, all the entries of $q_c(G)$, $q_c(G^\prime)$ are distinct.
    
    \item \label{cond:3} For any $G \in \tilde{\calG}_{\delta}$, all the entries of $q_c(G)$ are in $[t_l, t_r]$.
    
    \item \label{cond:4} For any $G \in \calG^+_{\delta} \setminus \tilde{\calG}_{\delta}$, all the entries of $q_c(G)$ are outside $[t_l, t_r]$.
\end{enumerate}
\end{lemma}
\begin{proof}
    See \cref{subsubsec:pf_conextmap} for a detailed proof.
\end{proof}
\begin{remark}
    Our proof differs from \cite{yun2019transformers} in one aspect: The original \cite[Lemma~6]{yun2019transformers} does not include positional encoding \eqref{eq:pos_encoding}. 
    Although \citet{yun2019transformers} sketches the proof for networks with \eqref{eq:pos_encoding} in the attention layer input, we detail the proof.
\end{remark}

\subsubsection{Map to the Desired Output by Modified Feed-forward Layers}
\label{subsubsec:map_to_desired_output}
Next, we show that a series of feed-forward layers map the output of modified attention layers $f_{\cT, c2}$ to the desired output of function $f_{\delta^\star}$.

\begin{lemma}[Lemma 7 of \cite{yun2019transformers}]
\label{lemma:memorize}
There exists a function $f_{\cT,c3}: \RR^{d \times L} \to \RR^{d \times L}$ composed of $\calO(L (1/\delta)^{dL}/L!)$ modified feed-forward layers, such that
\begin{align*}
    f_{\cT,c3} \circ f_{\cT,c2}(G) =
    \begin{cases}
        A_{G} & \text{ if } G \in \tilde {\calG}_{\delta},
        \\
        \vzero_{d \times L} & \text{ if } G \in \calG^+_{\delta} \setminus \tilde{\calG}_{\delta}.
    \end{cases}
\end{align*}
\end{lemma}
\begin{proof}
    See \cref{subsec:pf_memorize} for a detailed proof.
\end{proof}

In conclusion, we have the following lemma for the required number of layers in the modified transformer.
\begin{lemma}[Total Number of Layers]
\label{coro:network_depth_m}
    From the proof of \cref{lemma:part2_approx_piecewise_modified}, if we want to achieve a approximation error $\calO(\delta^{d/2})$ by the modified transformer, we need $\calO(\delta^{-1})$ modified feed-forward layers in $f_{\cT, c1}$, $\calO(\delta^{-d})$ modified self-attention layers in $f_{\cT, c2}$, and $\calO(\delta^{-dL})$ modified feed-forward layers in $f_{\cT, c3}$.
\end{lemma}
\begin{proof}
    By the proof of \cref{lemma:part2_approx_piecewise_modified}, we complete the proof.
\end{proof}

\subsection{Standard Transformers Approximate Modified Transformers}
\label{appendix_sec:modified_approximation_normal}
In this subsection, we show that standard neural network layers are able to approximate the modified self-attention layers and the modified feed-forward layers (\cref{def:modified_trans}). 
We have the following \cref{lemma:part3_original_transformer_approximation}.
\begin{lemma}[Lemma 9 of \cite{yun2019transformers}]
\label{lemma:part3_original_transformer_approximation}
For each $f_{\cT,c} \in \Bar {\cT}_p^{2,1,1}$ and any $\epsilon>0$, there exists $f_{\cT} \in \mathcal{T}_p^{2,1,4}$ such that $\norm{f_{\cT} -f_{\cT,c}}_{L^2} \leq \epsilon/3$.
\end{lemma}
\begin{proof}
See \cref{subsubsec:pf_original_transformer_approximation} for a detailed proof.
\end{proof}

\subsection{All Together: Standard Transformers Approximate Compact-supported Continuous Functions}
\label{subsec:summary}
We summarize the results of \cref{lemma:part1_piecewise,lemma:part2_approx_piecewise_modified,lemma:part3_original_transformer_approximation}.
Then we prove \cref{transformer_approximation}.

Furthermore, to achieve the $\epsilon$ approximation error in \cref{transformer_approximation}, we take $\delta = \calO(\epsilon^{2/d})$ in \cref{lemma:part2_approx_piecewise_modified}.

\clearpage
\subsection{Supplementary Proofs}
We first present two preliminary concepts: selective shift operation and bijective column ID mapping in \cref{subsubsec:pre}.

Then we show 
\begin{itemize}
    \item Proof of \cref{lemma:part1_piecewise} in \cref{subsubsec:pf_piecewise}

    \item Proof of \cref{lemma:contextmap} in \cref{subsubsec:pf_conextmap}

    \item Proof of \cref{lemma:memorize} in \cref{subsec:pf_memorize}

    \item Proof of \cref{lemma:part3_original_transformer_approximation} in \cref{subsubsec:pf_original_transformer_approximation}
\end{itemize}     

\subsubsection{Preliminaries}
\label{subsubsec:pre}

Here, we give the definition of two preliminary concepts: selective shift operation and bijective column ID mapping.
 
\paragraph{Selective Shift Operation.}
\label{item1:selective_shift_operation}
This operation refers to shifting certain entries of the input selectively.

To achieve this, we consider the following function $\xi(\cdot; \cdot): \RR^{d \times L} \rightarrow \RR^{d \times L}$
\begin{align}
\label{eq:xi1}
    \xi(X; b_Q) = e_1 u^\top X \sigma_H \left[ (u^\top X)^\top (u^\top X - b_Q \one_n^\top) \right],
\end{align}
where $X \in \RR^{d \times L}$, $e_1=(1,0,0,\cdots,0)^\top\in\RR^d$, and $b_Q \in \RR$.
$u\in \RR^d$ is a vector to be determined.

To see the output, we consider the $j$-th column of $u^\top X \sigma_H \left[ (u^\top X)^\top (u^\top X - b_Q \one_n^\top) \right]$:
\begin{itemize}
    \item If $u^\top X_{:,j}>b_Q$, it calculates $\argmax$ of $u^\top X$;
    \item If $u^\top X_{:,j}<b_Q$, it calculates $\argmin$ of $u^\top X$.
\end{itemize}
All rows of $\xi(X; b_Q)$ except the first row are zero. 
We consider the $j$-th entry of the first row in $\xi(X; b_Q)$, which is denoted as $\xi(X; b_Q)_{1, j}$.
Then for all $j \in [L]$, we have
\begin{align*}
\xi(X; b_Q)_{1, j} = 
u^\top X \sigma_H \[ (u^\top X)^\top (u^\top X_{:,j} - b_Q) \] =
    \begin{cases}
        \max_k u^\top X_{:,k} & \text{ if } u^\top X_{:,j} > b_Q, 
        \\
        \min_k u^\top X_{:,k} & \text{ if } u^\top X_{:,j} < b_Q.
    \end{cases}
\end{align*}

From this observation, we define a function parametrized by $b_Q$ and $b^\prime_Q$ (with $b_Q < b^\prime_Q$)
\begin{align}
\label{eq:xi2}
    \xi(X; b_Q, b^\prime_Q) := \xi(X; b_Q) - \xi(X; b^\prime_Q).
\end{align}
Then we have
\begin{align*}
    \xi(X; b_Q, b^\prime_Q)_{1, j} =
    \begin{cases}
        \max_k u^\top X_{:,k} - \min_k u^\top X_{:,k}, & ~ \text{if} ~ b_Q < u^\top X_{:,j} < b^\prime_Q, 
        \\
        0, & ~ \text{others}.
    \end{cases}
\end{align*}
We define an attention layer of the form $X \rightarrow X + \xi(X;b_Q,b'_Q)$.
For any column $X_{:,j}$, if $b_Q < u^\top X_{:,j} < b^\prime_Q$, its first coordinate $X_{1,j}$ is shifted up by $\max_k u^\top X_{:,k} - \min_k u^\top X_{:,k}$, while all the other coordinates stay untouched. 
We call this the selective shift operation because we can choose $b_Q$ and $b'_Q$ to shift certain entries of the input selectively.

\paragraph{Bijective Column ID Mapping.}
\label{item2:bijective_column_id_mapping}
We consider the input $G \in \calG^+_{\delta}$ (\cref{def:grid}).
We use 
\begin{align}
    J = L+3L\delta^{-dL},
    ~ \text{and} ~ 
    u = (1, \delta^{-1}, \delta^{-2}, \dots, \delta^{-d+1}).
    \label{eq:u_in_attn}
\end{align}

For any $j \in [L]$, we have the following two conclusions:
\begin{itemize}
\item \label{item1_bijection}
If $G_{i,j} \geq 0 $ for all $i \in [d]$, i.e., $G_{:,j} \in [j-1:\delta:j-\delta]^d$, then we have
\begin{align}
\label{eq:map_domain}
u^\top G_{:,j}\in \[\delta_j:\delta:\delta_j+\delta^{-d+1}-\delta\], ~ \text{where} ~ \delta_j = (j-1)\cdot \(\frac{\delta-\delta^{-d+1}}{\delta-1}\).
\end{align}
The mapping $G_{:,j} \rightarrow u^\top G_{:,j}$ maps the elements in $[j-1:\delta:j-\delta]^d$ to 
$\[\delta_j:\delta:\delta_j+\delta^{-d+1}-\delta\]$. 
This is a bijection.

\item \label{item2_bijection}
If there exists $i \in [d]$ such that $G_{i,j} = -J+j $, then 
\begin{align}
\label{eq:uG}
    u^\top G_{:,j} \leq -3L\delta^{-dL} + (j-1)\cdot \(\frac{\delta^{-d+1}-\delta}{1-\delta} \) + \delta^{-d+1} < 0.
\end{align}
\end{itemize}
 
We say that $u^\top G_{:,j}$ gives the ``column ID'' for each possible value of $G_{:,j} \in [j-1:\delta:j-\delta]^d$.

\begin{remark}[Illustration of Bijection Properity]
    For the bijection property, we give the following illustration.
    Let $G_{:j}=(g_{1j},g_{2j},\cdots,g_{dj})^\top$ and $\Bar{G}_{:j}=(\Bar g_{1j},\Bar g_{2j},\cdots,\Bar g_{dj})^\top$. 
    If $u^\top G_{:j}=u^\top \Bar{G}_{:j}$ and $G_{:j} \neq \Bar{G}_{:j}$, we deduce
    \begin{align}
    \label{eq:bijection}
        (g_{1j}-\Bar{g}_{1j})+\delta^{-1}(g_{2j}-\Bar{g}_{2j})+\cdots +\delta^{-d+1}(g_{dj}-\Bar{g}_{dj})=0.
    \end{align}
    Because $G_{:j} \neq \Bar{G}_{:j}$, then there exists a $k ~ (k<d)$, such that $g_{kj} \neq \Bar{g}_{kj}$ and $g_{ij} = \Bar{g}_{ij} (i > k)$. 
    We have 
    \begin{align*}
        \abs{\delta^{-k+1}(g_{kj}-\Bar{g}_{kj})}
        \geq \delta^{-k+2}.
    \end{align*}
    However, 
    \begin{align*}
        &~ \abs{(g_{1j}-\Bar{g}_{1j})+\cdots +\delta^{-k+2}(g_{k-1,j}-\Bar{g}_{k-1,j})}
        \\
        \leq &~  \abs{g_{1j}-\Bar{g}_{1j}} + \cdots + \abs{\delta^{-k+2}(g_{k-1,j}-\Bar{g}_{k-1,j})}
        \\
        \leq &~ (1-\delta) + \cdots + \delta^{-k+2}(1-\delta)
        \\
        < &~ \delta^{-k+2}.
    \end{align*}
    This contradicts with \eqref{eq:bijection}.
    Thus we prove the property of bijection. 
\end{remark}

\clearpage

\subsubsection{Proof of \texorpdfstring{\cref{lemma:part1_piecewise}}{}}
\label{subsubsec:pf_piecewise}

\begin{proof}[Proof of \cref{lemma:part1_piecewise}]
We restate the proof from \cite{yun2019transformers} for completeness.

By the nature of the compact-supported continuous function, $f$ is uniformly continuous. 

Because $\norm{\cdot}_{\infty}$ is equivalent to $\norm{\cdot}_F$ when the number of entries are finite, we have the following by the definition of uniform continuity.

For any $\epsilon/3>0$, there exists a $\delta^\star > 0$, such that for any $X, Y \in \RR^{d\times L}$, and $\norm{X-Y}_{\infty} < \delta^\star$, we have $\norm{f(X)-f(Y)}_F < \epsilon/3$.

Then we perform the following steps following \cref{def:grid_cube,def:pieceweise_const_func_class}:
\begin{itemize}

    \item We create a grid $\calG_{\delta^\star}$ by choosing grid width $\delta^\star$.
    We also create cube $\calS_G$ with respect to $G \in \calG_{\delta^\star}$.
    
    \item For any grid point $G \in \calG_{\delta^\star}$, we define $C_G \in \calS_G$ as the center point of the cube $\calS_G$.
    
    \item We define a piecewise-constant function $f_{\delta^\star}(X) = \sum\nolimits_{L \in \calG_{\delta^\star}} f(C_G) \one\{X \in \calS_G\}$.

\end{itemize}

For any $X \in \calS_G$, we have $\norm{X - C_G}_{\infty} < \delta^\star$.
According to the uniform continuity, we drive 
\begin{align*}
    \norm{f(X) - f_{\delta^\star}(X)}_F = \norm{f(X) - f(C_G)}_F< \epsilon / 3.
\end{align*}
This implies that $\norm{f- f_{\delta^\star}}_{L^2} < \epsilon/3$ and completes the proof.
\end{proof}

\subsubsection{Proof of \texorpdfstring{\cref{lemma:contextmap}}{}}
\label{subsubsec:pf_conextmap}

We give the proof of \cref{lemma:contextmap} by constructing the network to satisfy the requirements.

\begin{proof}[Proof of \cref{lemma:contextmap}]
Recall the selective shift operation in \cref{item1:selective_shift_operation}.
The overall idea of the construction includes two steps:
\begin{itemize}
    \item \textbf{Step 1:} For each $j \in [L]$, we stack $\delta^{-d}$ attention layers.
    For $g \in [\delta_j:\delta:\delta_j+\delta^{-d+1} - \delta] ~ \eqref{eq:map_domain}$ in the increasing order, we use the attention layer as 
    \begin{align}
    \label{eq:attn1}
        \delta^{-d} \xi(\cdot;g-\delta/2,g+\delta/2).
    \end{align}
    The total number of layers is $L\delta^{-d}$.
    These layers cast $G\in \wt \calG_{\delta}$ to $L$ different entries required by \hyperref[cond:1]{Property 1} of \cref{lemma:contextmap}.  
    \item \textbf{Step 2: }We add an extra single-head attention layer with the following attention part 
    \begin{align}
    \label{eq:attn2}
        L\delta^{-(L+1)d-1} \xi(\cdot;0).
    \end{align}
    This layer achieves a global shifting and casts different $G\in \wt \calG_{\delta}$ to unique elements required by the \hyperref[cond:2]{Property 2} of \cref{lemma:contextmap}. 
\end{itemize}
The two operations map $\tilde{\calG}_{\delta}$ and $\calG_{\delta}^+ \setminus \tilde{\calG}_{\delta}$ to different sets, as required by \hyperref[cond:3]{Property 3} and \hyperref[cond:4]{Property 4} of \cref{lemma:contextmap}. 
The bounds $t_l$ and $t_r$ are calculated then.

Then, we give a detailed proof by showing the impact of the two steps and verifying the four properties of \cref{lemma:contextmap}.
We achieve this by making a category division of $\calG^+_{\delta}$:
\begin{itemize}
    \item \textbf{Category 1:}
    $G \in \tilde {\calG}_{\delta}$, all entries in the point $G$ are between $0$ and $L-\delta$.
    \item \textbf{Category 2:}
    $G \in \calG^+_{\delta} \setminus \wt \calG_{\delta}$, the point $G$ has at least one entry that equals to $-J$.
\end{itemize}
Let $u = (1, \delta^{-1}, \delta^{-2}, \ldots, \delta^{-d+1})$. 
Recall that $\delta_j = (j-1)(\delta-\delta^{-d+1})/(\delta-1)$ for any $j \in [L]$ in \eqref{eq:map_domain}. 

\paragraph{Category 1.}
We denote $g_j \coloneqq u^\top G_{:,j}$, then we have $g_1 < g_2 < \cdots < g_L$.
The first $\delta^{-d}$ layers sweep the set $[\delta_j:\delta:\delta_j+\delta^{-d+1}-\delta], j\in [L]$ and apply selective shift operation on each element in the set. This means that selective shift operation will be applied to $g_1$ first, then $g_2$, followed by $g_3$, and so on.

\begin{itemize}
\item
{\bf The First Shift Operation.}
In the first selective shift operation with $g$ going through $[\delta_1:\delta:\delta_1+\delta^{-d+1}-\delta]$, the $(1,1)$-th entry of $G$ (i.e., $G_{1,1}$) is shifted by the operation, while the other entries are left untouched. The updated value $ \tilde{G}_{1,1}$ is
\begin{align*}
    \tilde{G}_{1,1} 
    = G_{1,1} + \delta^{-d} \[\max_k \(u^\top G_{:,k}\) - \min_k \(u^\top G_{:,k}\)\] 
    = G_{1,1} + \delta^{-d} (g_L - g_1).
\end{align*}
Therefore, the output of the layer after the operation is 
\begin{align*}
    \begin{pmatrix} 
    \tilde{G}_{:,1} & G_{:,2} & \cdots & G_{:, L} 
    \end{pmatrix}.
\end{align*}
Let $\tilde{g}_1  \coloneqq u^T \tilde{G}_{:,1} $. 
We have
\begin{align*}
    \tilde{g}_1 
    = & ~  \tilde{G}_{1,1} + \sum_{i=2}^d \delta^{-i+1} G_{i,1}
    \\
    = & ~ G_{1,1} + \delta^{-d} (g_L - g_1) + \sum_{i=2}^d \delta^{-i+1} G_{i,1}
    \\
    = & ~ g_1 + \delta^{-d} (g_L - g_1).
\end{align*}
Then we deduce $g_L < \tilde{g}_1$, because
\begin{align*}
    \tilde{g}_1  = & ~  g_1 + \delta^{-d} (g_L - g_1) 
    \\
    \geq & ~ 
    0+\delta^{-d}\[(L-1)\cdot \frac{\delta-\delta^{-d+1}}{\delta-1}-\delta^{-d+1}+\delta\]
    \annot{By \eqref{eq:map_domain}}
    \\
    = & ~ 
    \delta^{-d}\[(L-1)\frac{\delta}{1-\delta}+\delta+(L-1)\frac{\delta^{-d+1}}{1-\delta}-\delta^{-d+1}\] 
    \\
    \geq & ~ 
    \delta^{-d} \cdot \((L-1)\frac{\delta}{1-\delta}+\delta\)
    \\
    = & ~ 
    (L-1)\frac{\delta^{-d+1}}{1-\delta} + \delta^{-d + 1} 
    \\
    > & ~  g_L.
    \annot{By $\delta <1$ and \eqref{eq:map_domain}}
\end{align*}
Thus, after updating, we have
\begin{equation*}
    \max u^\top \begin{pmatrix} \tilde{G}_{:,1} & G_{:,2} & \cdots & G_{:,L} \end{pmatrix}
    = \max \{ \tilde{g}_1, g_2, \dots, g_L \} = \tilde{g}_1,
\end{equation*}
and the new minimum is $g_2$.

\item
{\bf The Second Shift Operation.}
In the second selective shift operation with $g$ going through $[\delta_2:\delta:\delta_2+\delta^{-d+1}-\delta]$, the $(1,2)$-th entry of $G$ (i.e., $G_{1,2}$) is shifted by the operation, while the other entries are left untouched. The updated value $ \tilde{G}_{1,2}$ is
\begin{align*}
    \tilde{G}_{1,2}
    = & ~  G_{1,2} + \delta^{-d} (\tilde{g}_1 - g_2) 
    \\
    = & ~  G_{1,2} + \delta^{-d} (g_1 - g_2) + \delta^{-2d} (g_L - g_1).
\end{align*}
Therefore, the output of the layer after the operation is 
\begin{align*}
    \begin{pmatrix} 
    \tilde{G}_{:,1} & \tilde{G}_{:,2} & \cdots & G_{:, L} 
    \end{pmatrix}.
\end{align*}
We have
\begin{align*}
    \tilde{g}_2 
    &\coloneqq u^\top \tilde{G}_{:,2} 
    \\
    & = g_2 + \delta^{-d} (g_1 - g_2) + \delta^{-2d} (g_L - g_1).
\end{align*}
Then we deduce $\tilde{g}_1 < \tilde{g}_2$, because
\begin{align*}
    & ~ g_1 + \delta^{-d} (g_L - g_1) < g_2 + \delta^{-d} (g_1 - g_2) + \delta^{-2d} (g_L - g_1) 
    \\
    \iff & ~
    (\delta^{-d}-1) (g_2 - g_1) < \delta^{-d}(\delta^{-d}-1) (g_L - g_1).
    \annot{By $\delta^{-d} > 1$ and $g_L > g_2$}
\end{align*}
Thus, after updating, we have
\begin{equation*}
    \max u^\top \begin{pmatrix} \tilde{G}_{:,1} & \tilde{G}_{:,2} & \cdots & G_{:,L} \end{pmatrix}
    = \max \{ \tilde{g}_1, \tilde{g}_2, \dots, g_L \} = \tilde{g}_2,
\end{equation*}
and the new minimum is $g_3$.

\item
{\bf Repeating the Process.}
By repeating this process, we show that the $j$-th shift operation shifts $G_{1,j}$ by $\delta^{-d} (\tilde{g}_{j-1} - g_j)$. 
Then we have
\begin{align*}
    \tilde{g}_j & \coloneqq 
    u^\top \tilde{G}_{:,j} 
    \\
    & =
    g_j + \sum_{k=1}^{j-1} \delta^{-kd}(g_{j-k}-g_{j-k+1}) + \delta^{-jd} (g_L-g_1).
\end{align*}
We deduce $\tilde{g}_{j-1} < \tilde{g}_j$ holds for all $2 \leq j \leq L$, because
\begin{align*}
     & ~ \tilde{g}_{j-1} < \tilde{g}_j 
     \\
     \iff~ & ~
     g_{j-1} + \sum_{k=2}^{j-1} \delta^{-kd+d}(g_{j-k}-g_{j-k+1}) + \delta^{-(j-1)d} (g_L-g_1) 
     \\
     & ~ < g_j + \sum_{k=1}^{j-1} \delta^{-kd}(g_{j-k}-g_{j-k+1}) + \delta^{-jd} (g_L-g_1)
     \\
     \iff~ & ~
     \sum_{k=1}^{j-1} \delta^{-kd+d} (\delta^{-d}-1)(g_{j-k+1}-g_{j-k}) <
     \delta^{-(j-1)d}(\delta^{-d}-1)(g_L-g_1),
\end{align*}
where the last inequality holds because
\begin{align*}
    &~ \sum_{k=1}^{j-1} \delta^{-kd+d} (g_{j-k+1}-g_{j-k})
    \\
    < &~ \delta^{-(j-1)d} \sum_{k=1}^{j-1}(g_{j-k+1} - g_{j-k})
    \\
    < &~ \delta^{-(j-1)d}(g_L-g_1).
\end{align*}
Therefore, after the $j$-th selective shift operation, $\tilde{g}_j$ is the new maximum among $\{\tilde{g}_1, \dots, \tilde{g}_j, g_{j+1}, \dots, g_L\}$ and $g_{j+1}$ is the new minimum.

\item
{\bf After $L$ Shift Operations.}
After the whole $L$ shift operations, the input $G$ is mapped to a new point $\tilde{G}$, where $u^\top \tilde{G} = \begin{pmatrix} \tilde{g}_1 & \tilde{g}_2 & \dots & \tilde{g}_L \end{pmatrix}$ and $\tilde{g}_1 < \tilde{g}_2 < \dots < \tilde{g}_L$.
For the lower and upper bound of $\tilde{g}_L$, we have the following lemma.
\begin{lemma}[Lemma 10 of \cite{yun2019transformers}]
\label{sublemma:1}
$\tilde{g}_L = u^\top \tilde{G}_{:,L}$ satisfies the following bounds:
\begin{align*}
    \delta^{-(L-1)d+1}(\delta^{-d} - 1) 
    \leq 
    \tilde{g}_L 
    \leq 
    L\delta^{-(L+1)d}.
\end{align*}
Also, the mapping from $\begin{pmatrix} g_1 & g_2 & \cdots & g_L \end{pmatrix}$ to $\tilde{g}_L$ is one-to-one mapping.
\end{lemma}

\item 
{\bf Global Shifting by the Last Layer.}
We note that after the above $L$ shift operations, there is another attention layer with attention part $L\delta^{-(L+1)d-1} \xi(\cdot;0)$.
Since $0<\tilde{g}_1 < \cdots < \tilde{g}_L$, it adds the following to each entry in the first row of $\tilde{G}$:
\begin{align*}
    L\delta^{-(L+1)d-1} \max_k u^\top \tilde{G}_{:,k} = L\delta^{-(L+1)d-1} \tilde{g}_L.
\end{align*}
The output of this layer is defined to be the function $f_{\cT,c2}(G)$.
\end{itemize}

In summary, for any $G \in \wt {\calG}_{\delta}$, $i \in [d]$, and $j \in [L]$, we have
\begin{align*}
    f_{\cT,c2}(G)_{i,j} & = 
    \begin{cases}
        G_{1,j} + \delta_j^+ & \text { if } i = 1,
        \\
        G_{i,j} & \text { if } 2 \leq i \leq d,
    \end{cases} 
\end{align*}
where $\delta_j^+ = \sum_{k=1}^{j-1} \delta^{-kd}(g_{j-k}-g_{j-k+1}) + \delta^{-jd} (g_L-g_1) + L\delta^{-(L+1)d-1} \tilde{g}_L$.

For any $G \in \tilde{\calG}_{\delta}$ and $j \in [L]$,
\begin{equation*}
    u^\top f_{\cT,c2}(G)_{:,j} = \tilde{g}_{j} + L\delta^{-(L+1)d-1} \tilde{g}_L.
\end{equation*}

Next, we check the \hyperref[cond:1]{Property 1}, \hyperref[cond:2]{Property 2} and \hyperref[cond:3]{Property 3} of \cref{lemma:contextmap}.

\begin{itemize}
\item
{\bf Checking \hyperref[cond:1]{Property 1} of \cref{lemma:contextmap}.}
Given any $G \in \wt {\calG}_{\delta}$, we already prove that 
\begin{align*}
\tilde{g}_1 < \tilde{g}_2 < \dots < \tilde{g}_L,
\end{align*}
All of them are distinct.

\item
{\bf Checking \hyperref[cond:2]{Property 2} of \cref{lemma:contextmap}.}
Note that the upper bound on $\tilde{g}_L$ from \cref{sublemma:1} also holds for other $\tilde{g}_j$ ($j \in [L-1]$).
For all $j \in [L]$, we have
\begin{align*}
L\delta^{-(L+1)d-1} \tilde{g}_L \leq  u^\top f_{\cT,c2}(G)_{:,j} < L\delta^{-(L+1)d-1} \tilde{g}_L + L\delta^{-(L+1)d}.
\end{align*}

By \cref{sublemma:1}, two different $G, G^\prime \in \wt {\calG}_{\delta}$ are mapped to different $\tilde{g}_L$ and $\tilde{g}^{\prime}_L$, and they differ at least by $\delta$. 
This means that the following two intervals are guaranteed to be disjoint:
\begin{align*}
    &~ [L\delta^{-(L+1)d-1} \tilde{g}_L, L\delta^{-(L+1)d-1} \tilde{g}_L + L\delta^{-(L+1)d}),
    \\
    &~ [L\delta^{-(L+1)d-1} \tilde{g}^{\prime}_L, L\delta^{-(L+1)d-1} \tilde{g}^{\prime}_L + L\delta^{-(L+1)d}).
\end{align*} 
Thus, the entries of $u^\top f_{\cT,c2}(G)$ and $u^\top f_{\cT,c2}(G^\prime)$ are all distinct. 

Now, we finish showing that the mapping $f_{\cT,c2}(\cdot)$ uses $(1/\delta)^d+1$ attention layers to implement a contextual mapping on $\wt {\calG}_{\delta}$.

\item 
{\bf Checking \hyperref[cond:3]{Property 3} of  \cref{lemma:contextmap}.}
Given Lemma~\ref{sublemma:1} and $u^\top f_{\cT,c2}(G)_{:,j} \in [L\delta^{-(L+1)d-1} \tilde{g}_L, L\delta^{-(L+1)d-1} \tilde{g}_L + L\delta^{-(L+1)d})$, for any $G \in \tilde{\calG}_{\delta}$, we have
\begin{align*}
    &~ u^\top f_{\cT,c2}(G)_{:,j} \geq 
    L\delta^{-2(L+1)d}(\delta^{-d} - 1), 
    \\
    &~ u^\top f_{\cT,c2}(G)_{:,j} < L^2\delta^{-2(L+1)d-1}+ L\delta^{-(L+1)d}.
\end{align*}
This proves that all $u^\top f_{\cT,c2}(L)_{:,j}$ are between $t_l$ and $t_r$, where
\begin{align*}
    &~ t_l = L\delta^{-2(L+1)d}(\delta^{-d} - 1),
    \\
    &~ t_r = L^2\delta^{-2(L+1)d-1}+ L\delta^{-(L+1)d}.
\end{align*}

\end{itemize}

\paragraph{Category 2.}
Now we check the \hyperref[cond:4]{Property 4} of \cref{lemma:contextmap}.
For the input points $G \in \calG^+_{\delta} \setminus \tilde{\calG}_{\delta}$, note that the point $G$ has at least one entry that equals to $-J+k,k\in[L-1]$.
Let $g_j \coloneqq u^\top G_{:,j}$.
Recall that whenever a column $G_{:,j}$ has an entry that equals to $-J+k,k\in[L-1]$, we have $g_j  < 0$.
Without loss of generality, assume that $g_1 <0$.

Because the selective shift operation is applied to each element of $[0:\delta:\delta_L+\delta^{-d+1}-\delta]$ and is not applied to negative values, thus we have $\min_k u^\top G_{:,k} = g_1 < 0$.
$g_1$ never gets shifted upwards and remains the minimum for the whole time.

\begin{itemize}
\item 
{\bf All $g_j$'s are Negative.}
When all $g_j$'s are negative, selective shift operation never shifts the input $G$.
Thus $\tilde{G} = G$. 
Recall that $u^\top \tilde{G}_{:,j} < 0$ for all $j \in [L]$. 
The last layer with attention part $L\delta^{-(L+1)d-1} \xi(\cdot;0)$ adds $L\delta^{-(L+1)d-1} \min_k u^\top \tilde{G}_{:,k} < 0$ to each entry in the first row of $\tilde{G}$.
This makes $\tilde{G}$ remain negative. Therefore, $f_{\cT,c2}(G)$ satisfies $u^\top f_{\cT,c2}(G)_{:,j} < 0 < t_l$ for all $j \in [L]$.

\item
{\bf Not All $g_j$'s are Negative.}
Now consider the case where at least one $g_j$ is positive. 
Suppose that there are $k$ positive elements and they satisfy $g_{i_1}<g_{i_2}<\cdots<g_{i_k}$. 
Thus selective shift operation does not affect $g_i$, where $i \in [L] \setminus \{i_1, \dots, i_k\}$.
It shifts $g_{i_1}$ by
\begin{align*}
    &~ \delta^{-d}(\max_k u^\top G_{:,k} - \min_k u^\top G_{:,k})
    \\
   \geq &~ \delta^{-d} (2L\delta^{-dL}-(L-1)\frac{\delta^{-d+1}-\delta}{1-\delta}-\delta^{-d+1}+(i_k-1)\frac{\delta^{-d+1}-\delta}{1-\delta})
   \annot{By \eqref{eq:uG}}
    \\
    = &~ \delta^{-d}(3L\delta^{-dL}-\delta^{-d+1}-(L-i_k)\frac{\delta^{-d+1}-\delta}{1-\delta})
    \\
    \geq &~  \delta^{-d} \cdot 2L\delta^{-dL}
    \annot{By $\delta^{-1} \geq 2$}
    \\
    = &~ 2L\delta^{-(L+1)d}.
\end{align*}
The next shift operations shift $g_{i_2}, \dots, g_{i_k}$ by an even larger amount.
Therefore, at the end of the first $L(1/\delta)^d$ layers, we have 
$L\delta^{-(L+1)d} \leq \tilde{g}_{i_1} \leq \dots \leq \tilde{g}_{i_k}$,
and $\tilde{g}_j < 0$ for all $j \in [L]\setminus\{i_1,\dots,i_{k}\} $.

Then, we shift $G$ by the last layer.
The last layer with attention part $L\delta^{-(L+1)d-1} \xi(\cdot;0)$ acts differently for negative and positive $\tilde{g}_j$'s. (i). For negative $\tilde{g}_j$'s, it adds the following to $\tilde{g}_j,j \in [L]\setminus\{i_1,\dots,i_{k}\}$:
\begin{align*}
L\delta^{-(L+1)d-1} \min_k u^\top \tilde{G}_{:,k} = L\delta^{-(L+1)d-1} g_1 < 0.
\end{align*}
This term pushes them further to the negative side.
(ii). For positive $\tilde{g}_{i}$'s, it adds 
\begin{align*}
    L\delta^{-(L+1)d-1} \max_k u^\top \tilde{G}_{k} = L\delta^{-(L+1)d-1} \tilde{g}_{i_k} \geq 2L^2\delta^{-2(L+1)d-1}.
\end{align*}
Thus they are all greater than or equal to $2L^2\delta^{-2(L+1)d+1}$.
Note that 
\begin{align*}
    2L^2\delta^{-2(L+1)d-1} > t_r, ~\text{where}~ t_r=L^2\delta^{-2(L+1)d-1}+ L\delta^{-(L+1)d}.
\end{align*}
Then the final output $f_{\cT,c2}(G)$ satisfies $u^\top f_{\cT,c2}(G)_{:,j} \notin [t_l, t_r]$, for all $j \in [L]$.
This completes the verification of \hyperref[cond:4]{Property 4} of \cref{lemma:contextmap}.
\end{itemize}

In conclusion, we need $\mathcal{O}(L\delta^{-d})$ layers of modified self-attention layer to obtain our approximation.
This completes the proof.
\end{proof}

\clearpage
\subsubsection{Proof of \texorpdfstring{\cref{lemma:memorize}}{}}
\label{subsec:pf_memorize}
\begin{proof}[Proof of \cref{lemma:memorize}]
We restate the proof from \cite{yun2019transformers} for completeness.

Note that $|\calG^+_{\delta}| = (1/\delta+1)^{dL}<\infty$, so the output of $f_{\cT,c2}(\calG^+_{\delta})$ has finite number of distinct real values. 
Let $M$ be the upper bound of all these possible values. 
By the construction of $f_{\cT,c2}$, $M>0$.

\paragraph{Construct the Layers: $f_{\cT,c3}(f_{\cT,c2}(G)) = \vzero_{d \times L}$ if $G \in \calG^+_{\delta} \setminus \tilde\calG_{\delta}$.}
According to \cref{lemma:contextmap}, for all $j \in [L]$, we have $u^\top f_{\cT,c2}(G)_{:,j} \in [t_l, t_r]$ if $G \in \tilde\calG_{\delta}$, and $u^\top f_{\cT,c2}(G)_{:,j} \notin [t_l, t_r]$ if $G \in \calG^+_{\delta} \setminus \tilde \calG_{\delta}$.
Due to this property, we add the following feed-forward layer.
\begin{definition}[Feed-forward Layer 3] 
\label{def:ff_3}
    The vectors $u$ and $\one_L$ act as the weight parameters, and $\zeta_3(\cdot)$ acts as the activation function in the feed-forward layer.
        \begin{align}
        \label{eq:FF_3}
        X \rightarrow X - (M+1) \one_L \zeta_3 (u^\top X),
        ~~
        \zeta_3(t) =
        \begin{cases}
            0 & \text{ if } t \in [t_l, t_r]
            \\
            1 & \text{ if } t \notin [t_l, t_r].
        \end{cases}
    \end{align}
\end{definition}
\begin{itemize}
    \item 
    {\bf Case for $G \in \calG^+_{\delta} \setminus \tilde \calG_{\delta}$.}
    We have $\zeta_3(u^\top f_{\cT,c2}(G)) = \one_L^\top$. 
    Thus, all the entries of the input are shifted by $-M-1$ and become strictly negative.
    \item 
    {\bf Case for $G \in \tilde \calG_{\delta}$.}
    We have $\zeta_3(u^\top f_{\cT,c2}(G)) = \vzero_L^\top$, so the output stays the same as the $f_{\cT,c2}(G)$.
\end{itemize}
With the input $f_{\cT,c2}(G)$, if $G \in \tilde \calG_{\delta}$, then $\zeta_3(u^\top f_{\cT,c2}(G)) = \vzero_L^\top$.
Thus, the output stays the same as the input.
If $G \in \calG^+_{\delta} \setminus \tilde \calG_{\delta}$, then $\zeta_3(u^\top f_{\cT,c2}(G)) = \one_L^\top$.
Thus, all the entries of the input are shifted by $-M-1$ and become strictly negative.

Next, we map those negative entries to zero.
For $i=1,2,\cdots,d$, we add the following layer:
\begin{definition}[Feed-forward Layer 4] 
\label{def:ff_4}
    The vectors $u$ and $e_i$ act as the weight parameters and $\zeta_4(\cdot)$ acts as the activation function in the feed-forward layer.
    \begin{align}
        \label{eq:FF_4}
        X \rightarrow X + e_i \zeta_4 ((e_i)^\top X),
        ~~
        \zeta_4(t) =
        \begin{cases}
            -t & \text{ if } t < 0
            \\
            0 & \text{ if } t \geq 0.
        \end{cases}
    \end{align}
\end{definition}
After these $d$ layers, the output for $G \in \calG^+_{\delta} \setminus \tilde \calG_{\delta}$ is a zero matrix, while the output for $G \in \tilde \calG_{\delta}$ remains $f_{\cT,c2}(G)$.

\paragraph{Construct the Layers: $f_{\cT,c3}(f_{\cT,c2}(G)) = A_{G}$ if $G \in \tilde\calG_{\delta}$.}

Each different $G$ is mapped to $L$ unique numbers $u^\top f_{\cT,c2}(G)$, which are at least $\delta$ apart from each other. 
We map each unique number to the corresponding output column as follows.
We choose one $\Bar{G} \in \tilde \calG_{\delta}$.
For each $u^\top f_{\cT,c2}(\Bar{G})_{:,j}$, $j \in [L]$, we add the following feed-forward layer.
\begin{definition}[Feed-forward Layer 5] 
\label{def:ff_5}
    The vectors $u$ and $e_i$ act as the weight parameters, and $\zeta_4(\cdot)$ acts as the activation function in the feed-forward layer.
    \begin{align}
        \label{eq:FF_5}
        X \rightarrow & X + \((A_{\Bar{G}})_{:,j} - f_{\cT,c2}({\Bar{G}})_{:,j}\) \zeta_5 (u^\top X - u^\top f_{\cT,c2}(\Bar{G})_{:,j} \one_L^\top),
        \\
        & \zeta_5(t) = 
        \begin{cases}
        1 & -\delta/2 \leq t < \delta/2,
        \\
        0 & ~\text{others}.
        \end{cases}
    \end{align}
\end{definition}

\begin{itemize}
\item 
{\bf Case for $G \in \calG^+_{\delta} \setminus \tilde \calG_{\delta}$. }
Recall that the input $X$ of this layer is $f_{\cT,c2}({G})$.
If $X$ is a zero matrix, which is the case for $G \in \calG^+_{\delta} \setminus \tilde \calG_{\delta}$, we have $u^\top X = \vzero_L^\top$. 
Then $u^\top X - u^\top f_{\cT,c2}({\Bar{G}})_{:,j} \one_L^\top<-t_l\one_L$. 
Since $t_l > \delta/2 $, the output remains the same as $X$.

\item 
{\bf Case for $G \in \tilde \calG_{\delta}$. }
Let the input $X$ be $f_{\cT,c2}(G)$, where $G \in \tilde \calG_{\delta}$ is not equal to $\Bar{G}$.
According to the \hyperref[cond:2]{Property 2} of \cref{lemma:contextmap} and given a $j \in [L]$, $u^\top f_{\cT,c2}(G)_{:,k}, ( k\in [L])$ differs from $u^\top f_{\cT,c2}({\Bar{G}})_{:,j}$ by at least $\delta$. 
Then we have
\begin{equation*}
    \zeta_5(u^\top f_{\cT,c2}(G) - u^\top f_{\cT,c2}({\Bar{G}})_{:,j} \one_L^\top) = \vzero_L^\top.
\end{equation*}
Thus the input is left untouched.

If $G=\Bar{G}$, then 
\begin{equation*}
    \zeta_5(u^\top f_{\cT,c2}(G) - u^\top f_{\cT,c2}({\Bar{G}})_{:,j} \one_L^\top) = (e_j)^\top.
\end{equation*}
Thus we shift the $j$-th column of $f_{\cT,c2}(G)$ to 
\begin{align*}
    f_{\cT,c2}(G)_{:,j} + ((A_{\Bar{G}})_{:,j} - f_{\cT,c2}({\Bar{G}})_{:,j}) 
    =
    f_{\cT,c2}(G)_{:,j} + ((A_G)_{:,j} - f_{\cT,c2}(G)_{:,j})
    = (A_G)_{:,j}.
\end{align*}
\end{itemize}

In other word, this layer maps the column $f_{\cT,c2}(G)_{:,j}$ to $(A_{G})_{:,j}$, without affecting any other columns.

For each $G \in \tilde \calG_{\delta}$, we defer that we need one layer for each unique value of $u^\top f_{\cT,c2}(G)_{:,j}$. 
Note that there are $\calO(\delta^{-dL})$ such numbers, so we use $\calO(\delta^{-dL})$ layers to finish our construction.

This completes the proof.
\end{proof}

\clearpage
\subsubsection{Proof of \texorpdfstring{\cref{lemma:part3_original_transformer_approximation}}{}}
\label{subsubsec:pf_original_transformer_approximation}

\begin{proof}[Proof of \cref{lemma:part3_original_transformer_approximation}] We restate the proof from \cite{yun2019transformers} for completeness.

The proof follows two steps:
(i) Approximate the modified self-attention layers.
(ii) Approximate the modified feed-forward layers. 

\begin{itemize}
    \item \textbf{Step 1: Approximate the Modified Self-Attention Layers.}
    
    We achieve this by approximating the $\Softmax$ operator $\sigma_S$ with the $\Hardmax$ operator $\sigma_H$. Given a matrix $X \in \RR^{d \times L}$, we have
    \begin{equation*}
        \sigma_S(\lambda X) \rightarrow \sigma_H(X), 
        \quad \text{as} \quad 
        \lambda \rightarrow \infty.
    \end{equation*}
    The operator is the only difference between the normal and the modified self-attention layers.
    We approximate the modified self-attention layer in $\Bar{\cT}_p^{r,m,l}$ by the normal self-attention layer with the same number of heads $r$ and head size $m$.

    \item \textbf{Step2: Approximate the Modified Feed-Forward Layers.}
     
    We achieve this by approximating the activation function in $\Psi$ with four ${\rm ReLU}$ functions. 
    From \cref{def:modified_trans}, we recall that $\Psi$ denotes three-piecewise functions with at least a constant piece. 
    We consider the following $\zeta\in\Psi$:
    \begin{equation*}
        \zeta(x) = 
        \begin{cases}
            b_1 & \text { if } x < c_1,\\
            a_2 x + b_2 & \text { if } c_1 \leq x < c_2,\\
            a_3 x + b_3 & \text { if } c_2 \leq x,
        \end{cases}
    \end{equation*}
    where $a_2, a_3, b_1, b_2, b_3, c_1, c_2 \in \RR$, and $c_1 < c_2$.
    
    We approximate $\zeta(x)$ by $\tilde{\zeta}(x)$ composed of four ${\rm ReLU}$ functions:
    \begin{align*}
    \tilde{\zeta}(x) = 
    &b_1 + \frac{a_2 c_1 + b_2 - b_1}{\epsilon} \rm{ReLU}(x-c_1+\epsilon)
    + \(a_2 - \frac{a_2 c_1 + b_2 - b_1}{\epsilon} \) \rm{ReLU}(x-c_1) 
    \\
    &+ \( \frac{a_3 c_2 + b_3 - a_2(c_2 - \epsilon) - b_2}{\epsilon} - a_2 \) \rm{ReLU}(x - c_2 + \epsilon)
    \\
    &+ \left ( a_3 - \frac{a_3 c_2 + b_3 - a_2(c_2 - \epsilon) - b_2}{\epsilon} \right ) \rm{ReLU}(x - c_2)
    \\
    =&
    \begin{cases}
        b_1 & \text { if } x < c_1-\epsilon,
        \\
        (a_2 c_1 + b_2 - b_1) (x - c_1)/\epsilon + a_2c_1 + b_2 & \text { if } c_1-\epsilon \leq x < c_1,
        \\
        a_2 x + b_2 & \text { if } c_1 \leq x < c_2-\epsilon,
        \\
        (a_3 c_2 + b_3 - a_2(c_2 - \epsilon) - b_2)(x - c_2)/\epsilon  + a_3 c_2 + b_3 & \text { if } c_2-\epsilon \leq x < c_2,
        \\
        a_3 x + b_3 & \text { if } c_2 \leq x.
    \end{cases}
    \end{align*}
    As $\epsilon \rightarrow 0$, we approximate $\zeta(x)$ by $\tilde{\zeta}(x)$. 
    The activation function is the only difference between the normal and modified feed-forward layers.
    We approximate the modified feed-forward layer in $\Bar{\cT}_p^{r,m,l}$ by the normal one.
    
    Thus, for any $f_{\cT, c} \in \Bar{\cT}_p^{2,1,1}$, there exists a function $f_{\cT} \in \cT_p^{2,1,4}$ to approximate $f_{\cT, c}$.    
\end{itemize}
This completes the proof.
\end{proof}

\clearpage
\section{Proofs of \texorpdfstring{\cref{sec:method}}{}}
\label{sec:pf_thm1}
Our proofs are motivated by the approximation and estimation theory of U-Net-based diffusion models in \cite{chen2023score}.
We use transformer networks' universal approximation theory in \cref{apendix_sec:transformer_approximation} and the covering number to proceed with our proof.
Specifically, we derive the approximation error bound in \cref{subsec:pf_thm_1} and the corresponding sample complexity bound in \cref{subsec:pf_thm_2}.
Then we show that the data distribution generated from the estimated score function converges toward a proximate area of the original one in \cref{subsec:pf_thm_3}.

\subsection{Proof of \texorpdfstring{\cref{theorem:1}}{}}
\label{subsec:pf_thm_1}
Here we present some auxiliary theoretical results in \cref{subsubsec:aux_thm1} to prepare for our main proof of \cref{theorem:1}.
Then we derive the approximation error bound of DiTs (i.e., the proof of \cref{theorem:1}) in \cref{sec:proof_of_them1}.

\subsubsection{Auxiliary Lemmas for \texorpdfstring{\cref{theorem:1}}{}.}
\label{subsubsec:aux_thm1}

We restate some auxiliary lemmas and their proofs from \cite{chen2023score} for later convenience.

\begin{lemma}[Lemma 16 of \cite{chen2023score}]
\label{Tool_lemma_1_t1}
    Consider a probability density function $p_h(h)=\exp(-C\norm{h}_2^2 /2)$ for $h\in \RR^{d_0}$ and constant $C>0$. Let $r_h>0$ be a fixed radius. Then it holds
    \begin{align*}
        & \int_{\norm{h}_2>r_h} p_h(h) \dd  h \leq \frac{2d_0 \pi^{d_0/2}}{C\Gamma(d_0/2+1)}r_h^{d_0-2}\exp(-Cr_h^2/2),\\
        & \int_{\norm{h}_2>r_h}\norm{h}_2^2 p_h(h) \dd  h \leq \frac{2d_0 \pi^{d_0/2}}{C\Gamma(d_0/2+1)}r_h^{d_0}\exp(-Cr_h^2/2).
    \end{align*}
\end{lemma}

\begin{lemma}[Lemma 2 of \cite{chen2023score}]
\label{lemma:g_out_of_range}
    Suppose Assumption \ref{assumption:2} holds and $g$ is defined as:
    \begin{align*}
        q(\Bar{h},t)=\int \frac{h \psi_t(\Bar{h} | h) p_h(h)}{\int \psi_t(\Bar{h}| h) p_h(h) \dd  h} \dd h, \quad
        \Bar{h}=B^\top \Bar{x}.
    \end{align*}
    Given $\epsilon > 0$, with $r_h = c \left(\sqrt{d_0\log (d_0/T_0) + \log (1/\epsilon)} \right)$ for an absolute constant $c$, it holds
    \begin{align*}
    \norm{q(\Bar{h}, t)\one\{\norm{\Bar{h}}_2 \geq r_h\}}_{L^2(P_t)} \leq \epsilon, ~ \text{for} ~ t \in [T_0, T].
    \end{align*}
\end{lemma}

\begin{lemma}[Theorem 1 of \cite{chen2023score}]
\label{lemma:coarse_upper_bound}
    We denote 
    \begin{align*}
\tau(r_h) = \sup_{t \in [T_0, T]} \sup_{\Bar{h} \in [0, r_h]^d} \norm{\frac{\partial}{\partial t} q(\Bar{h}, t)}_2.
\end{align*}
With $q(\Bar{h},t)=\int h\psi_t(\Bar{h}|h)p_h(h)/(\int \psi_t(\Bar{h}|h)p_h(h)\dd  h) \dd  h$ and $p_h$ satisfies \cref{assumption:2}, we have a coarse upper bound for $\tau (r_h)$: 
\begin{align*}
    \tau(r_h) = \calO\left(\frac{1+\beta^2(t)}{\beta(t)} \left(L_{s_+} + \frac{1}{\sigma(t)}\right) \sqrt{d_0} r_h\right) = \calO\left(e^{T/2}L_{s_+} r_h \sqrt{d_0}\right).
\end{align*}
\end{lemma}

\begin{lemma}[Lemma 10 of \cite{chen2020score}]
\label{lemma:approximation_property_for_lipschitz}
    For any given $\epsilon>0$, and $L$-Lipschitz function $g$ defined on $[0,1]^{d_0}$, there exists a continuous function $\Bar{f}$ constructed by trapezoid function, such that
    \begin{align*}
        \norm{g-\Bar{f}}_\infty \leq \epsilon.
    \end{align*}
    Moreover, the Lipschitz continuity of $\bar{f}$ is bounded:
\begin{align*}
\left\lvert \bar{f}(x) - \bar{f}(y) \right\rvert \leq 10d_0L \norm{x - y}_2 \quad \text{for any} \quad x, y \in [0, 1]^{d_0}.
\end{align*}
\end{lemma}

\subsubsection{Main Proof of \texorpdfstring{\cref{theorem:1}}{}}
\label{sec:proof_of_them1}
\begin{proof}[Proof of \cref{theorem:1}]

With $\nabla \log p_t^{h} \left( \Bar{h}\right)=B^\top s_{+} (\Bar{h},t) $, we have the following in \eqref{eq:score_docom_rearange}
\begin{align}
    q(\Bar{h},t)= \sigma(t)\nabla \log p_t^{h} \left( \Bar{h}\right)+B^{\top} \Bar{x} = \sigma(t)B^\top (s_{+} (\Bar{h},t)+ \Bar{x}).
\end{align}

We proceed as follows:
\begin{itemize}
    \item {\bf Step 1.}
    Approximate $q(\Bar{h}, t)$ with a compact-supported continuous function $\bar{f}(\Bar{h}, t)$.
    
    \item {\bf Step 2.}
    Approximate $\bar{f}(\Bar{h}, t)$ with a transformer network.
\end{itemize}

\textbf{Step 1. Approximate $q(\Bar{h}, t)$ with a Compact-supported Continuous Function $\bar{f}(\Bar{h}, t)$. }
We partition $\RR^{d_0}$ into a compact subset $H_1:=\{\Bar{h}|\norm{\Bar{h}}_2 \leq r_h \}$ and its complement $H_2$, where $r_h$ is to be determined later. 
We approximate $q(\Bar{h},t)$ on the two subsets respectively and then prove $\Bar{f}$'s continuity. 
Such a step achieves an estimation error of $\sqrt{d_0}\epsilon$ between $q(\Bar{h},t)$ and $\Bar{f}(\Bar{h},t)$. 
We show the main proof here. 

\begin{itemize}
\item 
\textbf{Approximation on $H_2 \times [T_0,T]$.} For any $\epsilon >0$, we take $r_h=c(\sqrt{d_0\log (d_0/T_0)-\log \epsilon})$. 
From \cref{lemma:g_out_of_range}, we have
\begin{align*}
   \norm{q(\Bar{h}, t)\one\{\norm{\Bar{h}}_2 \geq r_h\}}_{L^2(P_t)} \leq \epsilon \quad \text{for}\quad t \in [T_0, T].
\end{align*}
So we set $\Bar{f}(\Bar{h},t)=0$ on $H_2 \times [T_0,T]$.

\item 
\textbf{Approximation on  $H_1 \times [T_0,T]$.}
On $H_1 \times [T_0,T]$, we approximate $q(\Bar{h},t)$ by approximating each coordinate $q_k(\Bar{h},t)$ respectively, where $q(\Bar{h},t)=[q_1(\Bar{h},t),q_2(\Bar{h},t),\cdots ,q_{d_0}(\Bar{h},t)]$. 
We rescale the input by $y' =(\Bar{h} + r_h \one)/2r_h$ and $t' = t/T$.
Then the transformed input space is $[0, 1]^{d_0} \times [T_0/T, 1]$. 
We implement such a transformation by a single feed-forward layer. 

By \cref{assumption:3}, on-support score $s_{+}(\Bar{h}, t)$ is $L_{s_{+}}$-Lipschitz in $\Bar{h}$. 
This implies $q(\Bar{h}, t)$ is $(1+L_{s_{+}})$-Lipschitz in $\Bar{h}$. 
When taking the transformed inputs, $g(y', t') = q(2r_hy'-r_h \one, T t')$ becomes $2r_h(1+L_{s_{+}})$-Lipschitz in $y'$.
Similarly, each coordinate $g_k(y',t)$ is also $2r_h(1+L_{s_{+}})$-Lipschitz in $y'$. 
Here we take $L_h = 1 + L_{s_{+}}$.

Besides, $g(y', t')$ is $T\tau(r_h)$-Lipsichitz with respect to $t$, where 
\begin{align*}
\tau(r_h) = \sup_{t \in [T_0, T]} \sup_{\Bar{h} \in [0, r_h]^d} \norm{\frac{\partial}{\partial t} q(\Bar{h}, t)}_2.
\end{align*}
We have a coarse upper bound for $\tau(r_h)$ in \cref{lemma:coarse_upper_bound}. 
We restate it here for convenience
\begin{align*}
    \tau(r_h) = \calO\left(\frac{1+\beta^2(t)}{\beta(t)} \left(L_{s_{+}} + \frac{1}{\sigma(t)}\right) \sqrt{d_0} r_h\right) = \calO\left(e^{T/2}L_{s_{+}} r_h \sqrt{d_0}\right).
\end{align*}

In conclusion, each $g_k(y',t)$ is Lipsichitz continuous. 
So we can apply \cref{lemma:approximation_property_for_lipschitz} to determine $\Bar{f}_k(y',t)$ for approximating each coordinate. 
We concatenate $\bar{f}_i$'s together and construct $\bar{f} = [\bar{f}_1, \dots, \bar{f}_{d_0}]^\top$.
According to the construction in \cref{lemma:approximation_property_for_lipschitz} and for any given $\epsilon$, we achieve
\begin{align*}
\sup_{y', t' \in [0, 1]^d \times [T_0/T, 1]} \norm{\bar{f}(y', t') - g(y', t')}_\infty \leq \epsilon,
\end{align*}

Considering the input rescaling (i.e., $\Bar{h} \to y'$ and $t \to t'$), we obtain:
\begin{itemize}
    \item The constructed function is Lipschitz continuous in $\Bar{h}$. 
    For any $\Bar{h}_1, \Bar{h}_2 \in H_1$ and $t \in [T_0, T]$, it holds
    \begin{align}
    \label{eq:lip_h}
    \norm{\bar{f}(\Bar{h}_1, t) - \bar{f}(\Bar{h}_2, t)}_{\infty} \leq 10d_0 L_h \norm{\Bar{h}_1 - \Bar{h}_2}_2.
    \end{align}
    \item The function is also Lipschitz in $t$.
    For any $t_1, t_2 \in [T_0, T]$ and $\norm{\Bar{h}}_2 \leq r_h$, it holds
\begin{align*}
\norm{\bar{f}(\Bar{h}, t_1) - \bar{f}(\Bar{h}, t_2)}_{\infty} \leq 10 \tau(r_h) \norm{t_1 - t_2}_2.
\end{align*}
\end{itemize}

Due to the fact that the construction of $\bar f (\Bar{h},t)$ is based on trapezoid function, we have $\bar f (\Bar{h},t)=0$ for $\norm{\Bar{h}}_2=r_h$ and any $t\in [T_0,T]$. 
Thus, the two parts of $\bar f (\Bar{h},t)$ can be joined together. 
To be more specific, the above Lipschitz continuity in $\Bar{h}$ extends to the whole $\RR^{d_0}$. 

\item 
\textbf{Approximation Error Analysis under $L^2$ Norm. }The $L^2$ approximation error of $\bar{f}$ can be decomposed into two terms:
\begin{align*}
& ~ \norm{q(\Bar{h}, t) - \Bar{f}(\Bar{h}, t)}_{L^2(P_t^{h})} 
\\
= & ~ 
\norm{(q(\Bar{h}, t) - \bar{f}(\Bar{h}, t)) \mathds{1}\{\norm{\Bar{h}}_2 < r_h\}}_{L^2(P_t^{h})} + \norm{q(\Bar{h}, t) \mathds{1}\{\norm{\Bar{h}}_2 > r_h\}}_{L^2(P_t^{h})}.
\end{align*}
The second term in the RHS above has already been bounded with the selection of $r_h$:
\begin{align*}
    \norm{g(\Bar{h}, t) \mathds{1}\{\norm{\Bar{h}}_2 > r_h\}}_{L^2(P_t^{h})} \leq \epsilon.
\end{align*}
The first term is bounded by: 
\begin{align*}
     & ~ \norm{(q(\Bar{h}, t) - \bar{f}(\Bar{h}, t)) \mathds{1}\{\norm{\Bar{h}}_2 
    <  r_h\}}_{L^2(P_t^{h})} \\
    \leq & ~
    \sqrt{d_0}\sup_{y', t' \in [0, 1]^d \times [T_0/T, 1]} \norm{\bar{f}(y', t') - g(y', t')}_\infty \\
    \leq  & ~ \sqrt{d_0}\epsilon.
\end{align*}
Then we obtain
\begin{align*}
    \norm{q(\Bar{h}, t) - \bar{f}(\Bar{h}, t)}_{L^2(P_t^{h})}\leq (\sqrt{d_0}+1)\epsilon.
\end{align*}
If we substitute $\epsilon$ with $\epsilon/2$, we obtain that the approximation error of $\bar f (\Bar{h},t)$ is $\sqrt{d_0}\epsilon$. 

\end{itemize}

\textbf{Step 2. Approximate  $\Bar{f}(\Bar{h},t)$ by a Transformer. }
This step is based on the universal approximation of transformers for the compact-supported continuous function in \cref{transformer_approximation}. DiT uses time point $t$ to calculate the scale and shift value in the transformer backbone \cite{peebles2023scalable}.
It also transforms an input picture into a sequential version.
We ignore time point $t$ in the notation of the transformer network in DiT.
Recall the reshape layer $R(\cdot)$ in \cref{def:reshape_layer}, we consider using $f(\cdot): ={R^{-1}\circ f_{\calT} \circ R}(\cdot)$ to approximate $\Bar{f}_t(\cdot): =\Bar{f}(\cdot,t)$, where $f_{\calT} \in \calT_p^{2,1,4}$.

\begin{itemize}
\item 
\textbf{Overall Approximation Error. }
With \cref{transformer_approximation}, we approximate $\Bar{f}_t(\cdot)$ with $\hat{f}(\cdot): = {R^{-1}\circ \hat{f}_{\calT} \circ R}(\cdot)$.
We denote 
\begin{align*}
    H=R(\Bar{h}).
\end{align*}
We have 
\begin{align}
\label{eq:app_error}
    \norm{\Bar{f}_t(\Bar{h})-\hat{f}(\Bar{h})}_{L^2(P_t^{h})}
    &=\left(\int_{P_t^{h}} \norm{\Bar{f}_t(\Bar{h})-\hat{f}(\Bar{h})}_2^2 \dd  h\right)^{1/2} \nonumber 
    \\
    &=\left(\int_{P_t^{h}} \norm{R\circ \Bar{f}_t \circ R^{-1}(H) -R \circ \hat{f}\circ R^{-1}(H)}_F^2 \dd  h\right)^{1/2} \nonumber
    \\
    &=\left(\int_{P_t^{h}} \norm{R\circ \Bar{f}_t \circ R^{-1}(H) -\hat{f}_{\cT}(H)}_F^2 \dd  h\right)^{1/2} \nonumber
    \\
    &\leq \epsilon.
\end{align}
Along with Step 1, we obtain
\begin{align*}
     \norm{q(\Bar{h},t)-\hat{f}(\Bar{h})}_{L^2(P_t^{h})}\leq \norm{q(\Bar{h},t)-\Bar{f}(\Bar{h},t)}_{L^2(P_t^{h})}+\norm{\Bar{f}(\Bar{h},t) - \hat{f}(\Bar{h})}_{L^2(P_t^{h})}\leq (1+\sqrt{d_0})\epsilon.
\end{align*}
The constructed approximator to $\nabla \log p_t(x)$ is $s_{\hat{W}}=(B\hat{f}(B^\top x,t)-x)/\sigma(t)$, and the approximation error is 
\begin{align*}
    \norm{\nabla \log p_t(\cdot)-s_{\hat{W}}(\cdot,t)}_{L^2(P_t)}\leq \frac{1+\sqrt{d_0}}{\sigma(t)}\epsilon \quad \text{for any} \quad t\in[T_0,T].
\end{align*}

\item 
\textbf{Settling-down of Hyperparameters. }
We settle down the hyperparameters to configure our network here. 
We refer to \cref{appendix_sec:modified_approximation} for some of the following calculations. 

\begin{enumerate}
    \item \textbf{Model Architecture Depth $K$.} 

From \cref{coro:network_depth_m}, we have $K=\calO(\left(1/\delta\right)^{d L})$. 
To achieve $\epsilon$-error approximation, we set $\delta=\calO\left(\epsilon^{2/d}\right)$ according to \cref{lemma:part2_approx_piecewise_modified}. 
Thus we obtain
\begin{align}
\label{eq:K_est}
K=\calO\left( \epsilon^{-2L}\right) .
\end{align}

\item 
\textbf{Lipchitz Upperbound for Transformer: $L_{\cT}$.}

We denote $\Bar{f}_{t,R}(\cdot) = R\circ \Bar{f}_t \circ R^{-1}(\cdot)$.
We get the Lipshitz upper bound for $\hat{f}_{\cT} \in \mathcal{T}_{p}^{2.1 .4}$ in the following way

\begin{align*}
\left\|\hat{f}_{\cT}\left(H_{1}\right)-\hat{f}_{\cT}\left(H_{2}\right)\right\|_{F} & \leq \left\|\hat{f}_{\cT}\left(H_1\right)-\Bar{f}_{t,R}\left(H_1\right)\right\|_{F} + \left\|\Bar{f}_{t,R}\left(H_1\right)-\Bar{f}_{t,R}\left(H_2\right)\right\|_{F} 
\\ 
& \hspace{0.5cm} +
\left\|\Bar{f}_{t,R}\left(H_2\right)-\hat{f}_{\cT}\left(H_2\right)\right\|_{F} 
\\
& \leq 2\epsilon+\left\|\Bar{f}_{t,R}\left(H_1\right)-\Bar{f}_{t,R}\left(H_2\right)\right\|_{F} \annot{By \eqref{eq:app_error}} 
\\
& \leq 2\epsilon+10 d_{0}L_{s_{+}}\norm{H_1 - H_2}_{F}. \annot{By \eqref{eq:lip_h}}
\end{align*}
Then we get 
\begin{align}
\label{eq:L_tau_est}
    L_{\cT}=\calO\left(d_{0}L_{s_{+}}\right).
\end{align}

\item 
\textbf{Model Output Bound for $\calS_{\cT_{p}^{2,1,4}}$.}

For the output of the constructed transformer $\hat{f}_{\cT}(\cdot)$, according to \cref{lemma:memorize}, we have $\hat{f}_{\cT}(O)=O$, where $O=\vzero_{d\times L}$. 
Thus, with the Lipschitz upperbound $\calO(d_0 L_{s_{+}})$, we have $\|\hat{f}_{\cT}(H)\|_F = \calO(d_0 L_{s_{+}}r_h)$, where $\norm{H}_F \leq r_h$.
With $r_h=c(\sqrt{d_0\log (d_0/T_0) + \log (1/ \epsilon ) })$, we obtain
\begin{align}
\label{eq:C_tau_est}
    C_{\cT}=\calO\left(d_{0} L_{s_{+}} \cdot \sqrt{d_0\log (d_0/T_0)+\log (1/\epsilon)}\right).
\end{align}

\item 
\textbf{Model Parameters Bound: $C_{OV}^{2,\infty}, C_{OV}, C_{KQ}^{2,\infty}, C_{KQ}, C_E$.}

By definition, we have:
\begin{align*}
~\norm{(W_{OV}^i)^\top}_{2,\infty}\leq C_{OV}^{2,\infty}, ~\norm{(W_{OV}^i)^\top}_2 \leq C_{OV}, ~\norm{W_{KQ}^i}_{2,\infty}\leq C_{KQ}^{2,\infty}, ~\norm{W_{KQ}^i}_{2}\leq C_{KQ},
\end{align*}
where $i=1,2$. 
For simplicity, we omit $i$ hereafter, which does not affect our discussion. 

Recall that $\norm{Z}_{2,\infty}$ denotes the $2,\infty$-norm, where the $2$-norm is over columns and $\infty$-norm is over rows.
By the construction of modified attention layers \eqref{eq:attn1} and \eqref{eq:attn2} in \cref{subsubsec:pf_conextmap}, we consider $W_{OV}$ to have the largest norm, i.e.,
\begin{align*}
& W_{OV}= L\delta^{-(L+1)d-1} \cdot \left(\begin{array}{cccc}
1 & \delta^{-1} & \cdots & \delta^{-d+1} \\
0 & 0 & \cdots & 0 \\
\vdots & \vdots & \cdots & \vdots \\
0 & 0 & \cdots & 0
\end{array}\right).
\end{align*}
We give the following upper bounds
\begin{align}
\label{eq:W_ov_est_inf}
    \norm{W_{OV}^\top}_{2,\infty} 
    = &~ Ld \delta^{-(L+2)d} = \calO\left(\delta^{-Ld}\right),
\\
\label{eq:W_ov_est_2}
    \norm{W_{OV}^\top}_2 
    = &~ \sup_{\norm{x}_2=1} \norm{W_{OV}^\top x}_2 
    = L\delta^{-(L+1)d-1} \cdot \sqrt{\sum_{i=0}^{d-1}\delta^{-2i}} 
    = \calO\left(\delta^{-Ld}\right).
\end{align}

By \eqref{eq:attn1} and \eqref{eq:attn2} in \cref{subsubsec:pf_conextmap}, and the self-attention layers in \cref{subsubsec:pf_original_transformer_approximation}, we consider $W_{KQ}$ to have the largest norm, i.e.,
\begin{align*}
& W_{KQ}:=\left(\begin{array}{c}
1 \\
\delta^{-1} \\
\vdots \\
\delta^{-d+1}
\end{array}\right)\left(1, \delta^{-1}, \cdots, \delta^{-d+1}\right)=\left(\begin{array}{cccc}
1 & \delta^{-1} & \cdots & \delta^{-d+1} \\
\delta^{-1} & \delta^{-2} & \cdots & \delta^{-d} \\
\vdots & \vdots & \cdots & \vdots \\
\delta^{-d+1} & \delta^{-d} & \cdots & \delta^{-2d+2}
\end{array}\right).
\end{align*}
Then we have
\begin{align}
\label{eq:W_kq_est_inf}
 \left\|W_{KQ}\right\|_{2,\infty}
 =&~ \sqrt{\sum_{i=0}^{d-1}\delta^{-2i-2d+2}} 
 = \calO(\delta^{-2d}),
\\
 \norm{W_{KQ}}_2 
 =&~ \sup_{\norm{x}_2=1} \norm{W_{KQ}x}_2 = \delta^{-2d+2} = \calO(\delta^{-2d}).
 \label{eq:W_kq_est_2}
\end{align}

We substitute $\delta$ with $\calO\left(\epsilon^{2/d}\right)$ (according to \cref{subsec:summary}) and get:
\begin{align*}
& ~ C_{OV}^{2,\infty}=  (1/\epsilon)^{\calO(1)}, \\
& ~ C_{OV}=  (1/\epsilon)^{\calO(1)}, \\
& ~ C_{KQ}^{2,\infty}=  (1/\epsilon)^{\calO(1)}, \\
& ~ C_{KQ}=  (1/\epsilon)^{\calO(1)}.
\end{align*}

From the construction of positional encoder \eqref{eq:pos_encoding} in \cref{appendix_sec:modified_approximation}, we have
$$
\begin{aligned}
& E=\left(\begin{array}{cccc}
0 & 1 &\cdots & L-1 \\
0 & 1 & \cdots & L-1 \\
\vdots & \vdots & \vdots & \vdots \\
\vdots & \vdots & \vdots& \vdots \\
0 & 1 & \cdots & L-1
\end{array}\right).
\end{aligned}
$$
We deduce
\begin{align*}
\norm{E^\top}_{2,\infty}=\sqrt{L}(L-1) = \calO(L^{3/2}).
\end{align*}
Thus we have
\begin{align}
\label{eq:C_e_est}
    C_E = \calO(L^{3/2}).
\end{align}

\item 
\textbf{Parameters Bound in Feed Forward Layers: $C_{F}^{2,\infty}, C_{F}$.}

Recall the construction of modified feed-forward layers in the proof of \cref{lemma:contextmap}, which includes \cref{def:ff_1,def:ff_2,def:ff_3,def:ff_4,def:ff_5}.
With the approximation by normal feed-forward layers in \cref{subsubsec:pf_original_transformer_approximation}, we consider the weight parameters with the largest norm in the feed-forward layers, i.e.,
$$
\begin{aligned}
& W_{1}:=\left(\begin{array}{c}
1 \\
1 \\
1 \\
1
\end{array}\right)\left(1, \delta^{-1}, \cdots, \delta^{-d+1}\right)=\left(\begin{array}{cccc}
1 & \delta^{-1} & \cdots & \delta^{-d+1} \\
1 & \delta^{-1} & \cdots & \delta^{-d+1} \\
1 & \delta^{-1} & \cdots & \delta^{-d+1} \\
1 & \delta^{-1} & \cdots & \delta^{-d+1} \\
\end{array}\right)
\in \mathbb{R}^{4 \times d}.
\end{aligned}\\
$$
Then we have
\begin{align}
\label{eq:C_f_inf_est}
    C_{F}^{2,\infty} &= \calO\left(\sqrt{\sum_{i=0}^{d-1}\delta^{-2i}}\right) 
    = \calO\left(\delta^{-d}\right)
    \\
    &=(1/\epsilon)^{\calO(1)}. \nonumber
    \annot{By setting $\delta=\calO(\epsilon^{2/d})$ according to \cref{subsec:summary}}
\end{align}
and
\begin{align}
\label{eq:C_f_est}
    C_{F} &= \sup_{\norm{x}_2=1} \norm{W_{1}x}_2 
    = \calO\left(\delta^{-d}\right) 
    \\
    &=(1/\epsilon)^{\calO(1)}. \nonumber
    \annot{By setting $\delta=\calO(\epsilon^{2/d})$ according to \cref{subsec:summary}}
\end{align}
\end{enumerate}

\end{itemize}
This completes the proof.
\end{proof}

\clearpage
\subsection{Proof of \texorpdfstring{\cref{corollary:score_est}}{}}
\label{subsec:pf_thm_2}

Here we present the auxiliary theoretical results about the covering number of transformer networks in \cref{subsubsec:cover_trans}. 
The results are based on \cite[Theorem A.17]{Benjamin2020transformer_covering_number}.
Then we derive the sample complexity bound of DiTs (i.e., the proof of \cref{corollary:score_est}) in \cref{subsec:pf_thm_2}.

\subsubsection{Auxiliary Lemmas for \texorpdfstring{\cref{corollary:score_est}}{}}
\label{subsubsec:cover_trans}
\begin{lemma}[Lemma 15 of \cite{chen2023score}]
\label{lemma_15_used}
    Let $\mathcal{G}$ be a bounded function class.
    Then there exists a constant $b$ such that the output of any $g\in \mathcal{G}:\RR^{d_0}\mapsto[0,b]$ is bounded by $b$. 
    Let $z_1,z_2,\cdots,z_n\in \RR^{d_0}$ be i.i.d. random variables. 
    For any $\delta\in(0,1),a\leq 1$, and $c >0$, we have
    \begin{align*}
        &P\left(\sup_{g\in\mathcal{G}}\frac{1}{n}\sum_{i=1}^n g(z_i)-(1+a)\EE\left[g(z)\right]>\frac{(1+3/a)B}{3n}\log\frac{\mathcal{N}(c,\mathcal{G},\norm{\cdot}_\infty)}{\delta}+(2+a)c\right)\leq \delta, \\
        &P\left(\sup_{g\in\mathcal{G}}\EE\left[g(z)\right]-\frac{1+a}{n}\sum_{i=1}^n g(z_i)>\frac{(1+6/a)B}{3n}\log\frac{\mathcal{N}(c,\mathcal{G},\norm{\cdot}_\infty)}{\delta}+(2+a)c\right)\leq \delta .
    \end{align*}
\end{lemma}

Now, we give the definition of the covering number as follows.
\begin{definition}[Covering Number]
    Given a function class $\calF$ and a data distribution $P$. 
    Sample n data points $\{X_i\}_{i=1}^{n}$ from $P$.
    For any $\epsilon>0$, the covering number $\calN(\epsilon,\calF,\{X_i\}_{i=1}^{n},\norm{\cdot})$ is the smallest size of a collection (a cover) $\calC \in \calF$, such that for any $f \in \calF$, there exists a $\hat{f} \in \calC$ satisfying 
    \begin{align*}
        \max_i \norm{f(X_i)-\hat{f}(X_i)} \leq \epsilon.
    \end{align*}
    Furthermore, we define the covering number with respect to the data distribution as
    \begin{align*}
        \calN(\epsilon,\calF,\norm{\cdot}) = \sup_{\{X_i\}_{i=1}^{n} \sim P}\calN(\epsilon,\calF,\{X_i\}_{i=1}^{n},\norm{\cdot}).
    \end{align*}
\end{definition}
Then we give the covering number of the transformer networks.
\begin{lemma}[Modified from Theorem A.17 of \cite{Benjamin2020transformer_covering_number}]
\label{lemma:covering_number}
Let $\mathcal{T}_p^{r,m,l}(K, C_{\cT}, C_{OV}^{2,\infty}, C_{OV}, C_{KQ}^{2,\infty}, C_{KQ}, C_{F}^{2,\infty}, C_{F}, C_E, L_{\cT})$ represent the class of functions of $K$-layer transformer blocks satisfying the norm bound for matrix and Lipsichitz property for feed-forward layers. Then for all data point $\norm{X}_{2,\infty}\leq C_X$, we have
\begin{align*}
    & \log\mathcal{N}(\epsilon_c,\mathcal{T}_p^{r,m,l}(K, C_{\cT}, C_{OV}^{2,\infty}, C_{OV}, C_{KQ}^{2,\infty}, C_{KQ}, C_{F}^{2,\infty}, C_{F}, C_E, L_{\cT}),\norm{\cdot}_2)\\
    \leq & \frac{\log (nL)}{\epsilon_c^2}\cdot \(\sum_{i=1}^{K}
    \alpha^{\frac{2}{3}}\(d^{\frac{2}{3}}\(C_F^{2,\infty}\)^{\frac{4}{3}} 
    +d^{\frac{2}{3}}\(2(C_F)^2 C_{OV} C_{KQ}^{2,\infty}\)^{\frac{2}{3}}
    +\tau m^{\frac{2}{3}}\((C_F)^2 C_{OV}^{2,\infty}\)^{\frac{2}{3}}\)
    \)^3,
\end{align*}
where $\alpha \coloneqq \prod_{j<i} (C_F)^2 C_{OV} (1+4C_{KQ})(C_X+C_E)$.
\end{lemma}
\begin{remark}
    We modify \cite[Theorem A.17]{Benjamin2020transformer_covering_number} in seven aspects:
    \begin{enumerate}
        \item We do not consider the last linear layer in the model, which converts each column vector of the transformer output to a scalar. 
        Therefore, we ignore the item related to the last linear layer in \cite[Theorem A.17]{Benjamin2020transformer_covering_number}.
        \item We do not consider the normalization layer in our model.
        Because the normalization layer $\prod_{\rm norm}(\cdot)$ in the original proof only ensures that  $\norm{\prod_{\rm norm}(X_1)-\prod_{\rm norm}(X_2)}_{2, \infty}\leq \norm{X_1 - X_2}_{2, \infty}$, ignoring this layer does not change the result.
        \item Our activation function is ${\rm ReLU}$.
        Thus, we replace the Lipschitz upperbound of activate function by 1.
        \item We consider the positional encoding \eqref{eq:pos_encoding}.
        Then we need to replace the upperbound $C_X$ for the inputs with the upperbound $C_X+C_E$.
        Besides, for multi-layer transformer, the original conclusion in \cite[Theorem A.17]{Benjamin2020transformer_covering_number} uses 1 as the upperbound for the $2, \infty$-norm of inputs.
        We incorporate the upperbound for the inputs into the result stated in \cref{lemma:covering_number}.
        \item We use \eqref{eq:FF} as the feed-forward layer, including two linear layers and a residual layer. 
        Thus, we replace the original upperbound for the norm of weight matrix with the upperbound for the norm of $I_d+W_2W_1$ in \cref{lemma:covering_number}.
        In the following, we use $\calO$ to estimate the log-covering number, thus we ignore the item for $I_d$ here for converience. 
        This is the same for the self-attention layer.
        \item We use multi-head attention, and incorporate the number of heads $\tau$ into our result, which is similar to \cite[Theorem A.12]{Benjamin2020transformer_covering_number}.
        \item In our work, we use the transformer $\cT_p^{2,1,4}$, i.e., $\tau=2, m=1$.
    \end{enumerate}
\end{remark}

\subsubsection{Proof of \texorpdfstring{\cref{corollary:score_est}}{}}
\begin{proof}[Proof of \cref{corollary:score_est}]
Our proof is built on \citep[Appendix B.2]{chen2023score}.
For one data sample, we define the empirical score matching loss objective \eqref{eq:denoising_score_matching} as follows
\begin{align*}
    \ell(x; s_{\hat{W}}) = \frac{1}{T-T_0} \int_{T_0}^T \EE_{x_t |x_0 = x} [\norm{\nabla_{x_t} \log \psi_t(x_t | x_0) - s_{\hat{W}}(x_t, t)}_2^2] \dd t.
\end{align*}
Then we define $ \mathcal{L}(s_{\hat{W}}) = \EE_{ x  \sim P_{0}} \left[\ell(x; s_{\hat{W}})\right]$.

Following \citep[Appendix B.2]{chen2023score}, for any $a\in (0,1)$, we have
\begin{align*}
& ~ \cL(s_{\hat{W}}) \\
\leq & ~
\underbrace{\cL^{\rm trunc}(s_{\hat{W}}) - (1+a)\hat{\cL}^{\rm trunc}(s_{\hat{W}})}_{(I)} + \underbrace{\cL(s_{\hat{W}}) - \cL^{\rm trunc}(s_{\hat{W}})}_{(II)} + (1+a) \underbrace{\inf_{s_{W} \in \calS_{\rm NN}} \hat{\cL}(s_{W})}_{(III)},
\end{align*}
where
\begin{align*}
\cL^{\rm trunc}(s_{\hat{W}}) \coloneqq \EE_{ x  \sim P_{0}} \left[\ell^{\rm trunc}( x ; s_{\hat{W}})\right] = \EE_{ x  \sim P_{0}}  \left[\ell( x ; s_{\hat{W}})\mathds{1}\{\norm{ x }_2 \leq r_x\} \right], ~ r_x>B.
\end{align*}

We denote 
\begin{align*}
    ~ \eta 
    & \coloneqq 4C_{\cT}(C_{\cT}+r_x) (r_x/D)^{D-2} \exp(- r_x^2/\sigma(t))/(T_0(T - T_0)),
    \\
    ~ r_x & \coloneqq \calO\left(\sqrt{d_0\log d_0 + \log C_{\cT} + \log (n/\Bar{\delta})}\right).
\end{align*}
Then we have 
\begin{align}
\label{eq:tau_value}
    \eta \leq \frac{1}{nT_0(T-T_0)}.
\end{align}

For any $\Bar{\delta} >0$, according to \cref{lemma_15_used}, the following holds for term $(I)$ with probability $1-\Bar{\delta}$,
\begin{align*}
(I) = \calO\left(\frac{(1+6/a)(C_{\cT}^2 + r_x^2)}{n T_0(T-T_0)} \log \frac{\calN\left(\frac{(T-T_0)(\iota-\eta)}{(C_{\cT} + r_x)\log (T/T_0)}, \calS_{\cT_{p}^{2,1,4}}, \norm{\cdot}_2 \right)}{\Bar{\delta}} + (2+a) \iota \right),
\end{align*}
where $c \leq 0$ is a constant, and $\iota >0$ will be determined later.

We set 
\begin{align*}
    \iota \coloneqq \frac{2}{n^b T_0 (T-T_0)},
\end{align*} 
where $0 < b \leq 1$ is a constant to be determined later.

\begin{remark}[Selection Criteria of $\tau$]
We have two criteria:
\begin{itemize}
    \item 
    Recall that the covering number used in our setting is $\calN\left(\frac{(T-T_0)(\iota-\eta)}{(C_{\cT} + r_x)\log (T/T_0)}, \calS_{\cT_{p}^{2,1,4}}, \norm{\cdot}_2 \right)$. 
    Thus, we must ensure $\iota \geq \eta$.
    According to \eqref{eq:tau_value}, we consider $\iota$ satisfying the condition $\iota \geq (nT_0(T-T_0))^{-1}$. 
    Therefore, we consider $0 < b \leq 1$.

    \item 
    For the exponent of $(T-T_0)$, although selecting a value smaller than 1 is possible, we find that the convergence rate with respect to $T$ is dominated by the $1/T$ term appearing later in the second term of \eqref{eqn:above_eqn}.
    Therefore, we continue to consider the exponent to be 1.
\end{itemize}

\end{remark}
Then we have
{
\small
\begin{align*}
(I) = \calO\left(\frac{(1+6/a) \left(C_{\cT}^2 + r_x^2\right)}{n T_0 (T-T_0)} \log \frac{\calN\left( (n^b (C_{\cT} + r_x)T_0 \log (T/T_0))^{-1}, \calS_{\cT_{p}^{2,1,4}}, \norm{\cdot}_2 \right)}{\Bar{\delta}} + \frac{4+2a}{n^b T_0(T-T_0)}\right),
\end{align*}
}
with probability $1-\Bar{\delta}$. 

Following the proof structure of term $(II)$ in \citep[Appendix B.2]{chen2023score}, we have
\begin{align*}
(II) = \calO\left(\frac{1}{T_0} C_{\calT}^2 r_x^2 \exp{-A_2 r_x^2/2} \right).
\end{align*}

For any $\epsilon>0$, let $s_{\bar{W}}$ be the transformer network approximator to the score function in \cref{theorem:1}.
For the term $(III)$, we have
\begin{align*}
    (III) \leq \underbrace{\hat{\calL}(s_{\bar{W}}) - (1+a)\calL^{\rm trunc}(s_{\bar{W}})}_{(III)_1} + (1+a) \underbrace{\calL^{\rm trunc}(s_{\bar{W}})}_{(III)_2}.
\end{align*}

For any $\Bar{\delta} >0$, according to \cref{lemma_15_used} and given that $s_{\bar{W}}$ is a fixed function, the following holds for term $(III)_1$ with probability $1-\Bar{\delta}$, 
\begin{align*}
    (III)_1 = \calO\left(\frac{(1+3/a) \left(C_{\cT}^2 + r_x^2\right)}{n T_0 (T-T_0)} \log \frac{1}{\Bar{\delta}}\right).
\end{align*}

Following the proof structure of term $(III)_2$ in \citep[Appendix B.2]{chen2023score}, we have
\begin{align*}
    (III)_2 = \calO \left(\frac{d \epsilon^2}{T_0(T-T_0)} \right) + C_3,
\end{align*}
where $C_3$ is a constant.

Putting $(I)$, $(II)$, and $(III)$ together and setting $a=\epsilon^2$, then we have
{
\small
\begin{align}\label{eq:bound_with_covering}
&~ \frac{1}{T-T_0} \int_{T_0}^T \norm{s_{\hat{W}}(\cdot, t) - \nabla \log p_t(\cdot)}_{L^2(P_t)}^2 \dd  t \nonumber 
\\
= &~ \calO\left(\frac{\left(C_{\cT}^2 + r_x^2\right)}{\epsilon^2 n T_0(T - T_0)}\log \frac{\calN\left((n^b (C_{\cT} + r_x)T_0 \log (T/T_0))^{-1}, \calS_{\cT_{p}^{2,1,4}}, \norm{\cdot}_2 \right)}{\Bar{\delta}} + \frac{n^{-b}+d_0 \epsilon^2}{T_0(T-T_0)} \right),
\end{align}
}
with probability $1 - 3\Bar{\delta}$.

\paragraph{Covering Number of $\calS_{\cT_{p}^{2,1,4}}$.} 
The next step is to calculate the covering number of $\calS_{\cT_{p}^{2,1,4}}$. 
$\calS_{\cT_{p}^{2,1,4}}$ consists of two components: 
(i) Matrix $W_B$ with orthonormal columns; 
(ii) Network function $f_{\cT}$.

Suppose we have $W_{B1}, W_{B2}$ and $f_{1}, f_{2}$, such that $\norm{W_{B1} - W_{B2}}_{F} \leq \delta_1$ and $\sup_{\norm{ x }_2 \leq 3r_x+ \sqrt{D\log D}, t \in [T_0, T]} \norm{f_{1}( x , t) - f_{2}( x , t)}_2 \leq \delta_2$, where $f_{1}=R^{-1} \circ f_{\cT 1} \circ R, f_{2}=R^{-1} \circ f_{\cT 2} \circ R$.
Then we have
\begin{align}
\label{eq:lip_score_cover}
& \quad \sup_{\norm{ \Bar{x} }_2 \leq 3r_x + \sqrt{D\log D}, t \in [T_0, T]} \norm{s_{W_{B1}, f_{\cT 1}}( \Bar{x} , t) - s_{W_{B2}, f_{\cT 2}}( \Bar{x} , t)}_2 \nonumber
\\
& = \frac{1}{\sigma(t)} \sup_{\norm{ \Bar{x} }_2 \leq 3r_x + \sqrt{D\log D}, t \in [T_0, T]} \norm{W_{B1} f_{1}(W_{B1}^\top  \Bar{x} , t) - W_{B2}f_{2}(W_{B2}^\top  \Bar{x} , t)}_2 \nonumber
\\
& \leq \frac{1}{\sigma(t)} \sup_{\norm{ \Bar{x} }_2 \leq 3r_x + \sqrt{D\log D}, t \in [T_0, T]} \Bigg(\norm{W_{B1} f_{1}(W_{B1}^\top  \Bar{x} , t) - W_{B1} f_{1}(W_{B2}^\top  \Bar{x} , t)}_2 \nonumber
\\
& \quad + \norm{W_{B1} f_{1}(W_{B2}^\top  \Bar{x} , t) - W_{B1} f_{2}(W_{B2}^\top  \Bar{x} , t)}_2 + \norm{W_{B1} f_{2}(W_{B2}^\top  \Bar{x} , t) - W_{B2} f_{2}(W_{B2}^\top  \Bar{x} , t)}_2 \Bigg) \nonumber
\\
& \leq \frac{1}{\sigma(t)} \left(L_{\cT} \delta_1 
\sqrt{d_0}(3r_x + \sqrt{D\log D}) + \delta_2 + \delta_1 K\right),
\end{align}
where $L_{\cT}$ upper bounds the Lipschitz constant of $f_{\cT}$. 

For the set $\{W_B \in \RR^{D \times d_0}: \norm{W_B}_{\rm 2} \leq 1\}$, its $\delta_1$-covering number is $\left(1 + 2 \sqrt{d_0}/\delta_1\right)^{Dd_0}$ \cite[Lemma 8]{chenminshuo2019}. 
The $\delta_2$-covering number of $f$ needs further discussion as there is a reshaping process in our network. 
The input is reshaped from $ \Bar{h} \in \RR^{d_0}$ to $H \in \RR^{d\times L}$, and
\begin{align*}
    \norm{ \Bar{h} }_2 \leq r_x \Longleftrightarrow \norm{H}_{F} \leq r_x.
\end{align*}
Thus we have
\begin{align*}
    &~\sup_{\norm{ \Bar{h} }_2 \leq 3r_x+ \sqrt{D\log D}, t \in [T_0, T]} \norm{f_{1}( \Bar{h} , t) - f_{2}( \Bar{h} , t)}_2 \leq \delta_2 \\
    \Longleftrightarrow &~
    \sup_{\norm{H}_F \leq 3r_x + \sqrt{D\log D}, t \in [T_0, T]} \norm{f_{\cT 1}(H) -f_{\cT 2}(H)}_2 \leq \delta_2.
\end{align*}
Then we follow the covering number of sequence-to-sequence transformer $\cT_{p}^{2,1,4}$ in \cref{lemma:covering_number}.
We get the following $\delta_2$-covering number
\begin{align*}
 \frac{\log (nL)}{\delta_2^2}\cdot \(\sum_{i=1}^{K}
    \alpha_i^{\frac{2}{3}}\(d^{\frac{2}{3}}\(C_F^{2,\infty}\)^{\frac{4}{3}} 
    +d^{\frac{2}{3}}\(2(C_F)^2 C_{OV} C_{KQ}^{2,\infty}\)^{\frac{2}{3}}
    +\tau m^{\frac{2}{3}}\((C_F)^2 C_{OV}^{2,\infty}\)^{\frac{2}{3}}\)
    \)^3,
\end{align*}
where 
\begin{align*}
    \alpha_i \coloneqq \prod_{j<i} (C_F)^2 C_{OV} (1+4C_{KQ})(C_X+C_E).
\end{align*}
According to the \eqref{eq:K_est}, \eqref{eq:L_tau_est}, \eqref{eq:W_ov_est_inf}, \eqref{eq:W_ov_est_2}, \eqref{eq:W_kq_est_inf}, \eqref{eq:W_kq_est_2}, \eqref{eq:C_f_inf_est}, \eqref{eq:C_f_est}, \eqref{eq:C_e_est} and \eqref{eq:C_tau_est} in \cref{sec:proof_of_them1}, we derive the following with $\delta=\calO(\epsilon^{2/d})$ (\cref{subsec:summary}) and $d=4$ (\cref{theorem:1}):
\begin{align}
\label{eq:upper_bound_ori}
    & ~ K=\calO\left( \epsilon^{-2L}\right), L_{\cT} = \calO\left(d_{0}L_{s_{+}}\right),
    ~ C_{OV}^{2, \infty} 
    = \calO(d\epsilon^{-4L}),
    ~ C_{OV}
    = \calO(\epsilon^{-4L}), \nonumber
    \\
    & ~ C_{KQ}^{2, \infty} = \calO(\epsilon^{-4}), 
    ~ C_{KQ}
    = \calO(\epsilon^{-4}),
    ~ C_{F}^{2,\infty}
    = \calO(\epsilon^{-4}), ~ C_{F} = \calO(\epsilon^{-2}),
    ~ C_E = \calO(L^{3/2}), 
    \\
    & ~ C_{\cT}=\calO\left(d_{0} L_{s_{+}} \cdot \sqrt{d_0\log (d_0/T_0)+\log (1/\epsilon)}\right), ~r_x = \calO\left(\sqrt{d_0\log d_0 + \log C_{\cT} + \log (n/\Bar{\delta})}\right). \nonumber
\end{align}
Each element of the input data is within $[0,1]$, as shown in \cref{apendix_sec:transformer_approximation}.

For any $\delta_3>0$, we get the log-covering number of $\cT_{p}^{2,1,4}$,
\begin{align*}
    ~ \log \calN\left(\delta_3, \cT_{p}^{2,1,4}, \norm{\cdot}_2 \right) 
    = & ~ \calO
    \(\frac{\epsilon^{-8K}\cdot L^{K} d^2 \log(nL)}{\delta_3}\) \\
    = & ~ \calO(1) 
     \cdot 
     \(\frac{ 2^{ 8 K \log(L/\epsilon) } d^2 \log(nL)}{\delta_3} \).
\end{align*}

According to \eqref{eq:bound_with_covering}, we adopt the following value for $\delta_3$ in our setting
\begin{align*}
    \delta_3 = \frac{1}{n^b (C_{\cT} + r_x)T_0\log (T/T_0)}.
\end{align*}

According to \citep[Appendix B.2]{chen2023score}, the log-covering number of $\calS_{\cT_{p}^{2,1,4}}$ is
\begin{align*}
& ~ \log \calN\left(\delta_3, \calS_{\cT_{p}^{2,1,4}}, \norm{\cdot}_2 \right) 
\\
= & ~ \calO \( 2Dd_0 \cdot \log \(1+\frac{6 C_{\cT} L_{\cT} \sqrt{d_0}(3r_x+\sqrt{D\log D})}{T_0 \delta_3}\) + \frac{ 2^{8 K \log(L/\epsilon)}  d^2 \log(nL)}{T_0^2 \delta_3^2}\) 
\annot{By \eqref{eq:lip_score_cover}}
\\
= & ~ \calO \(n^{2b} 2^{8 (1/\epsilon)^L \log(L/\epsilon)} D d^2 d_0^6 L_{s_{+}}^2 \cdot \log(nL)  \) 
\annot{By \eqref{eq:upper_bound_ori}}
\\
= & ~ \calO \(n^{2b} 2^{ (1/\epsilon)^{2L} } D d^2 d_0^6 L_{s_{+}}^2 \cdot \log(nL)  \) 
\annot{By $(1/\epsilon)^L \geq 8 \log(L/\epsilon)$}
\\
= & ~ \wt{\calO} \(n^{2b} 2^{ (1/\epsilon)^{2L} } D d^2 d_0^6 L_{s_{+}}^2  \) 
\annot{By ignoring the log factors}
\\
= & ~  \wt{\calO} \(n^{2b} 2^{ (1/\epsilon)^{2L} } D d^2 d_0^6 L_{s_{+}}^2  \).
\end{align*}

Substituting the log-covering number into \eqref{eq:bound_with_covering}, we have
\begin{align}
& ~ \frac{1}{T-T_0} \int_{T_0}^T \norm{s_{\hat{W}}(\cdot, t) - \nabla \log p_t(\cdot)}_{L^2(P_t)}^2 \dd  t \nonumber
\\
= & ~ \calO \Big(\frac{ C_{\cT}^2 + r_x^2 }{\epsilon^2 n T_0(T - T_0)} ( \log \calN (\delta_3, \calS_{\cT_{p}^{2,1,4}}, \norm{\cdot}_2 ) + \log(1/\Bar{\delta}) ) + \frac{1}{n^b T_0(T-T_0)} + \frac{d_0}{T_0(T-T_0)} \epsilon^2  \Big)  \nonumber
\\
= & ~ \calO \Big( \underbrace{ \frac{ C_{\cT}^2 + r_x^2 }{\epsilon^2 n T_0T} ( \log \calN (\delta_3, \calS_{\cT_{p}^{2,1,4}}, \norm{\cdot}_2 )  + \log(1/\Bar{\delta}) ) }_{ \mathrm{1st~term} } + \frac{1}{n^b T_0 T} + \underbrace{ \frac{d_0}{T_0 T} \epsilon^2 }_{ \mathrm{2nd~term} }  \Big)  .
\label{eqn:above_eqn}
\end{align}

Recall the following parameters:
\begin{itemize}
    \item $C_{\cT}^2=\calO(d_{0}^2 L_{s_{+}}^2  d_0\log (d_0/T_0)+\log (1/\epsilon) )$,
    \item $r_x^2 = \calO( d_0\log d_0 + \log C_{\cT} + \log (n/\Bar{\delta}) )$,
    \item $\Bar{\delta}$: probability error,
    \item $\epsilon$: approximation error,
    \item $n$: sample size,
    \item $T_0 < T/2$,
    \item $D, d, d_0 > 1$: feature dimension,
    \item $L > 1$: sequence length,
    \item $d_0 = L \cdot d$,
    \item $L_{s_+}$: Lipschitz coefficient.
\end{itemize}

Ignoring the $\log $ factors and $\poly(D,d,d_0, L_{S_+})$, the first term in \eqref{eqn:above_eqn} becomes
\begin{align*}
\frac{1}{n^{1-2b}} \cdot \frac{1}{T_0 T} \cdot 2^{(1/\epsilon)^{2L}}.
\end{align*}
The second term is simplified to
\begin{align*}
    \frac{1}{T_0 T} \epsilon^2 .
\end{align*}

Thus, the final bound is
\begin{align*}
    \wt{O} \Bigg( \frac{1}{n^{1-2b} } \cdot \frac{1}{T_0 T}\cdot 2^{(1/\epsilon)^{2L}} + \frac{1}{n^b T_0 T} + \frac{1}{ T_0 T} \epsilon^2 \Bigg).
\end{align*}

To balance the first and second terms with respect to $n$, we select $b=1/3$.
Therefore, we give the final bound as 
\begin{align*}
    \wt{O} \Bigg( \frac{1}{n^{1/3} } \cdot \frac{1}{T_0 T}\cdot 2^{(1/\epsilon)^{2L}} + \frac{1}{n^{1/3} T_0 T} + \frac{1}{ T_0 T} \epsilon^2 \Bigg).
\end{align*}

This completes the proof.
\end{proof}

\clearpage
\subsection{Proof of \texorpdfstring{\cref{corollary:dist_est}}{}}
\label{subsec:pf_thm_3}
Our proof is built on \citep[Appendix C]{chen2023score}. 
The main difference between our work and \cite{chen2023score} is our score estimation error in \cref{corollary:score_est}. 
Consequently, only the subspace error and the total variation distance differ from \cite[Theorem 3]{chen2023score}.

First, we introduce the ground truth backward SDE and the learned backward SDE of the latent variable.
Recall from \eqref{eq:backward_SDE}, $y_t$ denotes the backward process.
We denote the backward latent variable by $h_t^{\leftarrow} = B^\top y_t$.
Since we write the time index explicitly, we drop the $\bar{y},\bar{h}$ notation for $t>0$.

Following  \citep[Appendix C.2]{chen2023score}, we have the following ground truth backward process
\begin{align*}
    \dd  h_t^{\leftarrow} = \left[\frac{1}{2} h_t^{\leftarrow} + \nabla \log p_{T-t}^{h}(h_t^{\leftarrow})\right] \dd  t + \dd  (B^\top \Bar{W}_t),
\end{align*}
where $\Bar{W}_t$ denotes the reversed Wiener process (standard Brownian motion) at time $t$ (see \cref{sec:background} for more details).

We define $P_{T_0}^{h}$ as the \textit{ground truth} marginal distribution of $h_{T_0}^{\leftarrow}$.

For the learned process $\tilde{y}_t$, we consider $\tilde{h}_t^{\leftarrow} = W_B^\top \tilde{y}_t$.
For any orthogonal matrix $U \in \RR^{d_0 \times d_0}$, we define the $U$ transformed version of $\tilde{h}_t^{\leftarrow}$ as $\tilde{h}_t^{\leftarrow, U} = U^\top \tilde{h}_t^{\leftarrow}$. 
Then the backward SDE for $\tilde{h}_t^{\leftarrow, U}$ is 
\begin{align*}
    \dd  \tilde{h}_t^{\leftarrow, U} = \left[\tilde{h}_t^{\leftarrow, U} + \tilde{s}_{U,f}^{h}(\tilde{h}_t^{\leftarrow, U}, T-t) \right] \dd  t + \dd  (U^\top W_B^\top \Bar{W}_t),
\end{align*}
where 
\begin{align*}
    \widetilde{s}_{U, f}^{h}\left(\tilde{h}_t^{\leftarrow, U}, t\right) \coloneqq \frac{1}{\sigma(t)} [-\tilde{h}_t^{\leftarrow, U} + U^\top f(U\tilde{h}_t^{\leftarrow, U}, t)].
\end{align*}
We define $\hat{P}_{T_0}^{h}$ as the \textit{estimated }marginal distribution of $\tilde{h}_{T_0}^{\leftarrow, U}$ from above continuous SDE.

The discretized backward SDE of $\tilde{h}_{T_0}^{\leftarrow, U}$ is 
\begin{align*}
    \dd  \tilde{h}_t^{\Leftarrow, U} = \left[\tilde{h}_{k\mu}^{\Leftarrow, U} + \tilde{s}_{U,f}^{h}(\tilde{h}_{k\mu}^{\Leftarrow, U}, T-k\mu) \right] \dd  t + \dd  (U^\top W_B^\top \Bar{W}_t), t \in [k\mu, (k+1)\mu).
\end{align*}
We define $\hat{P}_{T_0}^{h, \rm{dis}}$ as the \textit{estimated} marginal distribution of $\tilde{h}_{T_0}^{\Leftarrow, U}$ from above discrete SDE.

Next, we present the auxiliary theoretical results in \cref{subsubsec:aux_thm3} to prepare our main proof of \cref{corollary:dist_est}. 
Then we give a detailed proof of \cref{corollary:dist_est} in \cref{subsubsec:proof_coro_2}.

\subsubsection{Auxiliary Lemmas}
\label{subsubsec:aux_thm3}
Here we include a few auxiliary lemmas from \cite{chen2023score} without proofs.
Recall the definition of Lipschitz norm: for a given function $f$, $\norm{f(\cdot)}_{Lip} = \sup_{x\neq y}(\norm{f(x)-f(y)}_2/\norm{x-y}_2)$.
\begin{lemma}[Lemma 3 of \cite{chen2023score}]
\label{lemma_3_used}
    Assume that the following holds
    \begin{align*}
        \EE_{h\sim P_h}\norm{\nabla \log p_h(h)}_2^2\leq C_{sh},\quad
        \lambda_{\rm min}\EE_{h\sim P_h}[hh^\top]\geq c_0,\quad
        \EE_{h\sim P_h}\norm{h}_2^2\leq C_{h},
    \end{align*}
    where $\lambda_{\rm min}$ denotes the smallest eigenvalue.
    We denote
    \begin{align}
    \label{eq: expect_h_x}
        \Bar{\EE}[\phi(\Bar{h},t)] = \int_{T_0}^{T}\frac{1}{\sigma^2(t)}\EE_{\Bar{x} \sim P_t}[\phi(B^\top \Bar{x}, t)]dt.
    \end{align}
    
    Let $T_0\leq \min \{2 \log(d_0/C_{sh}), 1, 2 \log(c_0),c_0 \}$ and $T\geq \max\{ 2 \log(C_{h}/d_0),1\}$. 
    Suppose we have
    \begin{align}
    \label{eq: recover_estimation_error}
        \Bar{\EE}\norm{W_B f(W_B ^\top \Bar{x} ,t)-B q(B^\top \Bar{x} ,t)}_2^2\leq \epsilon.
    \end{align}
    Then we have 
    \begin{align*}
        \norm{W_BW_B^\top-BB^\top}_{\rm F}^2=\calO(\epsilon T_0 /c_0),
    \end{align*}
    and there exists an orthogonal matrix $U\in \RR^{d_0\times d_0}$, such that:
    \begin{align*}
        &\quad\Bar{\EE}\norm{U^\top f(U\Bar{h},t)-q(\Bar{h},t)}_2^2\\
        &=\epsilon\cdot\calO\left(1+\frac{T_0}{c_0}\[(T-\log T_0)d_0\cdot \max_{t}\norm{f(\cdot,t)}_{\rm Lip}^2+C_{sh}\]+\frac{\max_t\norm{f(\cdot,t)}_{\rm Lip}^2\cdot C_{h} }{c_0}\right).
    \end{align*}
\end{lemma}

\begin{lemma}[Lemma 4 of \cite{chen2023score}]
\label{lemma_4_used}
Assume that $P_h$ is sub-Gaussian and that $f(\Bar{h}, t)$ and $\nabla \log p_t^{h}(\Bar{h})$ are Lipschitz continuous with respect to $\Bar{h}$ and $t$. 
For any orthogonal matrix $U \in \RR^{d_0 \times d_0}$, we define 
\begin{align*}
    \widetilde{s}_{U, f}^{h}\left(\Bar{h}, t\right) \coloneqq \frac{1}{\sigma(t)} [-\Bar{h} + U^\top f(U\Bar{h}, t)].
\end{align*}
Assume that we have the latent score matching error-bound
\begin{align*}
\int_{T_0}^T \mathbb{E}_{\Bar{h} \sim P_t^{h}}\left\|\widetilde{s}_{U, f}^{h}\left(\Bar{h}, t\right)-\nabla \log p_t^{h}\left(\Bar{h}\right)\right\|_2^2 \mathrm{~d} t \leq \epsilon_{\text {latent }}(T-T_0),
\end{align*}
where $\epsilon_{\text {latent }} > 0$.
Then we have the following latent distribution estimation error for the continuous backward SDE:
$$
\operatorname{TV}\left(P_{T_0}^{h}, \widehat{P}_{T_0}^{h}\right) \lesssim \sqrt{\epsilon_{\text {latent }}(T-T_0)}+\sqrt{\mathrm{KL}\left(P_h \| N\left(0, I_{d_0}\right)\right)} \cdot \exp (-T),
$$
where $\widehat{P}_{T_0}^{h}$ is the marginal distribution of the generated $h_{T_0}$ using the continuous backward SDE.

Furthermore, let $\widehat{P}_{T_0}^{h, \mathrm{dis}}$ denote the marginal distribution of the generated $h_{T_0}$ using the discretized backward SDE.
Then we have the following latent distribution estimation error for the discretized backward SDE
$$
\operatorname{TV}\left(P_{T_0}^{h}, \widehat{P}_{T_0}^{h, \mathrm{dis}}\right) \lesssim \sqrt{\epsilon_{\text {latent}}(T-T_0)}+\sqrt{\mathrm{KL}\left(P_h \| N\left(0, I_{d_0}\right)\right)} \cdot \exp (-T)+\sqrt{\epsilon_{\text {dis}}(T-T_0)},
$$
where
\begin{align*}
    \epsilon_{\rm d i s}=&\left(\frac{\max _{\Bar{h}}\left\|f(\Bar{h}, \cdot)\right\|_{\text {Lip }}}{\sigma\left(T_0\right)}+
    \frac{\max_{\Bar{h}, t}\left\|f(\Bar{h}, t)\right\|_2}{T_0^2}\right)^2 \eta^2 \\
    &+\left(\frac{\max _t\left\|f(\cdot, t)\right\|_{\text {Lip }}}{\sigma\left(T_0\right)}\right)^2 \eta^2 \max \left\{\mathbb{E}\left\|h_0\right\|^2, d_0\right\}+\eta d_0 ,
\end{align*}
and $\eta$ is the step size in the backward process.
\end{lemma}

\begin{lemma}[Lemma 6 of \cite{chen2023score}]
\label{lemma_6_used}
Consider the following discretized SDE with step size $\mu$ satisfying $T-T_0=K_T \mu$
for some $K_T\in\mathbb{N}_+$,
$$
\mathrm{d} y_t=\left[\frac{1}{2}-\frac{1}{\sigma(T-k \mu)}\right] {y}_{k \mu} \mathrm{d} t+\mathrm{d} {B}_t, \quad\text {for} \quad
t \in[k \mu,(k+1) \mu),
$$
where $y_0 \sim \mathrm{N}(0, I)$.
Then, for $T>1$ and $T_0+\mu \leq 1$, we have $y_{T-T_0} \sim \mathrm{N}\left(0, \sigma^2 I\right)$ with $\sigma^2 \leq e\left(T_0+\mu\right)$.
\end{lemma}

\begin{lemma}[Lemma 10 in \cite{chen2023score}]
\label{lemma_10_used}
    Assume that $\nabla \log p_h(h)$ is $L_{h}$-Lipschitz. 
    Then we have $\mathbb{E}_{h \sim P_h}\left\|\nabla \log p_h(h)\right\|_2^2 \leq d_0 L_{h}$.
\end{lemma}

\subsubsection{Main Proof of \texorpdfstring{\cref{corollary:dist_est}}{}}
\label{subsubsec:proof_coro_2}
\begin{proof}
    Recall the estimation error in \cref{corollary:score_est} is $\xi(n,\epsilon,L)/(TT_0)$, where
    \begin{align*}
        \xi(n,\epsilon,L):=\frac{1}{n^{1/3}}\cdot 2^{(1/\epsilon)^{2L}} + \frac{1}{n^{1/3}} + \epsilon^2.
    \end{align*}
    
    \begin{itemize}
    \item \textbf{Proof of  (i).} 
    By the definition of \eqref{eq: expect_h_x} and the estimation error in \cref{corollary:score_est}, the error bound in \eqref{eq: recover_estimation_error} equals to $\xi(n,\epsilon,L) (T-T_0)/(TT_0)$ in \cref{lemma_3_used}.
    By 
    \cref{lemma_10_used}, we set $C_{sh}= d_0 L_{h}$.
    Then, we have
    \begin{align*}
        \norm{W_B W_B^\top - BB^\top}_F^2=\calO\Bigg(\frac{\xi(n,\epsilon,L)}{c_0}\Bigg).
    \end{align*}

    By substituting the value of $\xi(n,\epsilon,L)$ and $T = \calO(\log n)$ into the bound above, we deduce
    \begin{align*}
        \norm{W_B W_B^\top - BB^\top}_F^2= \calO \(\frac{1}{c_0 n^{1/3} } 2^{(1/\epsilon)^{2L}} + \frac{1}{c_0 n^{1/3}} + \frac{\epsilon^2}{c_0} \).
    \end{align*}

    \item \textbf{Proof of (ii).}
    Recall that $\max_{t}\norm{f(\cdot,t)}_{\rm Lip} \leq L_{\calT}$.
    Furthermore, according to \cref{lemma_3_used} and \cref{lemma_10_used}, we have
    \begin{align*}
        \Bar{\EE}\norm{U^\top f(U\Bar{h},t)-q(\Bar{h},t)}_2^2=\calO(\epsilon_{\text{latent}}(T-T_0)),
    \end{align*}
    where
    \begin{align*}
        \epsilon_{\text{latent}}=\frac{\xi(n,\epsilon,L)}{TT_0} \cdot \calO\left(\frac{T_0}{c_0}\left[(T-\log T_0)d_0\cdot L_{\calT}^2+d_0L_{h}\right]+\frac{L_{\calT}^2\cdot C_{h}}{c_0}\right). 
    \end{align*}
    Following the proof structure in \citep[Appendix C.4]{chen2023score}, we get
    \begin{align*}
        \bar\EE\norm{U^\top f(U\Bar{h},t)-q(\Bar{h},t)}_2^2 
        & = \int_{T_0}^T \EE_{\Bar{h}\sim P_t^{h}}\norm{\frac{U^\top f(U\Bar{h},t)-\Bar{h}}{\sigma(t)}-\nabla\log p_t^{h}(\Bar{h})}_2^2 \dd  t \\
        & \leq \epsilon_{\text{latent}}(T-T_0). 
    \end{align*}
    Following the proof structure in \citep[Appendix C.4]{chen2023score} and setting $T=\calO(\log n)$, we obtain
    \begin{align*}
        {\sf TV} (P_{T_0}^{h}, \widehat{P}_{T_0}^{h, \mathrm{dis}}) & = \tilde\calO\left(\sqrt{\epsilon_{\text {latent }}(T-T_0)}\right)
        \\
        & = \tilde{\calO} \(\sqrt{\(\frac{1}{n^{1/3} } 2^{(1/\epsilon)^{2L}} + \frac{1}{n^{1/3}} + \epsilon^2\) \cdot \log n} \),
    \end{align*}
    where $\tilde{\calO}$ hides the factor about $D, d_0, d, L_{s_+}, \log n$, and $T-T_0$

    By definition, $\hat{P}_{T_0}^{h,{\rm dis}}=(UW_B)_{\sharp}^\top \hat{P}_{T_0}$, where $\hat{P}_{T_0}$ is the distribution generated by $s_{\hat{W}}$ using the discretized backward process.
    This completes the proof of 
    the total variation distance.

    \item \textbf{Proof of (iii).}
    We apply \cref{lemma_6_used} due to our score decomposition. 
    With the marginal distribution at time $T-T_0$ and observing $\mu\ll T_0$, we obtain the last property.
    \end{itemize}
    This completes the proof.
\end{proof}

\clearpage
\section{Proofs of \texorpdfstring{\cref{sec:comp}}{}}

Our proofs are motivated by the observation of low-rank gradient decomposition in transformer-like models \cite{as24_neurips,gu2024tensor}.
With our simplifications and observations made in \cref{sec:comp},
we utilize the fine-grained complexity results of transformer and attention \cite{hu2024computational,as23_tensor,as24_neurips} and tensor trick (\cref{lemma:tensor_trick} and \cite{diao2019optimal,diao2018sketching}) to proceed our proofs.
Specifically, we approximate  DiT training gradients with a series of  low-rank approximations in \cref{sec:DiT_grad_decomp,sec:proof_building_block_I,sec:proof_building_block_II},
and carefully match the multiplication dimensions so that the computation of $\dv{g_2}{\underline{W}}$ forms a chained low-rank approximation in \cref{proof:thm:main_comp_back}.

\subsection{Auxiliary Theoretical Results for \texorpdfstring{\cref{thm:main_comp_back}}{}} 

Here we present some auxiliary theoretical results to prepare our main proof of the Existence of almost-linear Time Algorithms for \textsc{ADITGC} \cref{thm:main_comp_back}.

\subsubsection{Low-Rank Decomposition of DiT Gradients}
\label{sec:DiT_grad_decomp}

We start by some definitions.
Recall that  $W \in \R^{d \times d}$ and  $\underline{W}\in\R^{d^2}$ denotes the vectorization of $W\in\R^{d\times d}$ following \cref{def:vectorization}.

\begin{definition}\label{def:u}
Let $A_1, A_2 \in \mathbb{R}^{d \times L}$ be two matrices. 
Suppose $\A = A_1^\top \otimes A_2^\top \in \mathbb{R}^{L^2 \times d^2}$. 
Define $\A_{j_0} \in \mathbb{R}^{L \times d^2}$ as an $L \times d^2$ sub-block of $\A$. 
There are $L$ such sub-blocks in total.
For each $j_0 \in [L]$, define the function $u(\underline{W})_{j_0}: \mathbb{R}^{d^2} \to \mathbb{R}^L$ by
    $u(\underline{W})_{j_0} :=  \exp( \A_{j_0} \underline{W} ) \in\R^{L }$.
\end{definition}

\begin{definition}\label{def:alpha}
Let $A_1, A_2 \in \mathbb{R}^{d \times L}$ be two matrices. 
Suppose $\A = A_1^\top \otimes A_2^\top \in \mathbb{R}^{L^2 \times d^2}$. 
Define $\A_{j_0} \in \mathbb{R}^{L \times d^2}$ as an $L \times d^2$ sub-block of $\A$. 
There are $L$ such sub-blocks in total.
For every index $j_0 \in [L]$, consider the function $\alpha(\underline{W})_{j_0}: \mathbb{R}^{d^2} \to \mathbb{R}$ defined by
  $\alpha(\underline{W})_{j_0}:= \langle \underbrace{ \exp( \A_{j_0} \underline{W} ) }_{L \times 1} , \underbrace{ \one_L }_{L \times 1} \rangle$.
\end{definition}

\begin{definition}\label{def:f}
Suppose that $\alpha(\underline{W})_{j_0} \in \mathbb{R}$ and $u(\underline{W})_{j_0} \in \mathbb{R}^L$ are defined as in \cref{def:alpha,def:u}, respectively.
For a fixed $j_0 \in [L]$, consider the function $f(\underline{W})_{j_0} : \mathbb{R}^{d^2} \rightarrow \mathbb{R}^L$ defined by
\begin{align*}
    f(\underline{W})_{j_0} := \underbrace{ \alpha(\underline{W})_{j_0}^{-1} }_{ \mathrm{scalar} } \underbrace{ u(\underline{W})_{j_0} }_{ L \times 1 } .
\end{align*}
Define $f(\underline{W}) \in \mathbb{R}^{L \times L}$ as the matrix where the $j_0$-th row is $(f(\underline{W})_{j_0})^\top$.
\end{definition}

\begin{definition}\label{def:h}
For every $i_0 \in [d]$, define the function $h(\underline{W}_{OV})_{i_0} : \mathbb{R}^{d^2} \rightarrow \mathbb{R}^L$ by
\begin{align*}
    h(\underline{W}_{OV})_{i_0}:= \underbrace{ A_3^\top }_{L \times d} \underbrace{ (W_{OV}^\top)_{*,i_0} }_{d \times 1}.
\end{align*}
Here, $W_{OV} \in \mathbb{R}^{d \times d}$ denotes the matrix representation of $\underline{W}_{OV} \in \mathbb{R}^{d^2}$, and $(W_{OV})^\top_{*,i_0}$ represents the $i_0$-th column of $W_{OV}^\top$. Define $h(\underline{W}_{OV}) \in \mathbb{R}^{L \times d}$ as the matrix where the $i_0$-th column is $h(\underline{W}_{OV})_{i_0}$.
\end{definition}

\begin{definition}\label{def:c}
For each $j_0 \in [L]$, we denote $f(\underline{W})_{j_0} \in \mathbb{R}^L$ as the normalized vector defined by \cref{def:f}. For each $i_0 \in [d]$, $h(\underline{W}_{OV})_{i_0}$ is defined as per \cref{def:h}.
For every pair $(j_0, i_0) \in [L] \times [d]$, define the function $c(\underline{W})_{j_0,i_0}: \mathbb{R}^{d^2} \times \mathbb{R}^{d^2} \rightarrow \mathbb{R}$ by
\begin{align*}
    c(\underline{W})_{j_0,i_0} := \langle f(\underline{W})_{j_0}, h(\underline{W}_{OV})_{i_0} \rangle - Y^\top_{j_0,i_0},
\end{align*}
where $(W_{OV})_{j_0,i_0}$ is the element at the $(j_0, i_0)$ position of the matrix $W_{OV} \in \mathbb{R}^{L \times d}$.
$c(\cdot)$ has matrix form
\begin{align*}
\underbrace{ c(\underline{W}) }_{L \times d} = \underbrace{ f(\underline{W}) }_{L \times L} \underbrace{ h(\underline{W}_{OV}) }_{L \times d} - \underbrace{Y^\top}_{L \times d}.
\end{align*}
\end{definition}

With the tensor trick (\cref{sec:comp_preliminary}),
    we compute the gradient $\dv{g_2}{\underline{W}}$ of the DiT loss as follows:
    \begin{align}\label{eqn:simplified_dg2_dw}
         \dv{g_2}{\underline{W}}=
         \dv{}{\underline{W}}
        \[\half \sum_{j_0=1}^{L}\sum_{i_0=1}^d c_{j_0,i_0}^2(\underline{W})\].
    \end{align}
\eqref{eqn:simplified_dg2_dw} presents a neat decomposition of $\dv{g_2}{\underline{W}}$.
Each term is easy enough to handle.
Thus, we arrive at the following lemma.
Let $Z[i,\cdot]$ and $Z[\cdot,j]$ be the $i$-th row and $j$-th column of matrix $Z$.
\begin{lemma}[Low-Rank Decomposition of DiT Gradient]
\label{lemma:low_rank_grad}
    Let matrix $A_1,A_2,A_3,W,W_{OV},Y$ and loss function $\mathcal{L}$ follow \cref{def:generic_dit_loss}, and $\A\coloneqq A_1^\top\otimes A_2^\top$.
    It holds
    \bea
    \label{eqn:grad_low_rank}
    \dv{g_2}{\underline{W}}
    =\sum_{j_0=1}^L \sum_{i_0=1}^d c(\underline{W})_{j_0, i_0} \A_{j_0}^{\top}\underbrace{\Big(\overbrace{\diag\left(f(\underline{W})_j\right)}^{(II)}-\overbrace{f(\underline{W})_{j_0} f(\underline{W})_{j_0}^{\top}}^{(III)}\Big) 
    }_{(I)}h(\underline{W}_{OV})_{i_0}.
    \eea
\end{lemma}
\begin{proof}

Let $Z[i,\cdot]$ and $Z[\cdot,j]$ be the $i$-th row and $j$-th column of matrix $Z$. 

    With DiT loss  \cref{def:generic_dit_loss},
    we have
   \begin{align*}
    \dv{g_2}{\underline{W}}
   &=\half \sum_{j_0=1}^L \sum_{i=1}^d \dv{}{\underline{W}} c^2_{j_0,i_0}(\underline{W})
   \\
   &= \sum_{j_0=1}^L \sum_{i=1}^d \dv{}{\underline{W}} c^2_{j_0,i_0} c(\underline{W})_{j_0,i_0} \cdot 
   \dv{c(\underline{W})_{j_0,i_0}}{\underline{W}_{i_0}}
   \\
   &= \sum_{j_0=1}^L \sum_{i=1}^d \dv{}{\underline{W}} c^2_{j_0,i_0} c(\underline{W})_{j_0,i_0} \cdot 
   \dv{\left\langle f(\underline{W})_{j_0}, h(\underline{W}_{OV})_{i_0}\right\rangle}
   { \underline{W}_{i_0}}  
   \annot{By \cref{def:c}}
   \\
   & = \sum_{j_0=1}^L \sum_{i=1}^d \dv{}{\underline{W}} c^2_{j_0,i_0} c(\underline{W})_{j_0,i_0} \cdot
   \left\langle\dv
   {f(\underline{W})_{j_0}}
   {\underline{W}_{i}}, h(\underline{W}_{OV})_{i_0} \right\rangle
   \\
   &= \sum_{j_0=1}^L \sum_{i=1}^d \dv{}{\underline{W}} c^2_{j_0,i_0} c(\underline{W})_{j_0,i_0} \cdot\left\langle
   \dv{\alpha^{-1}(\underline{W})_{j_0} u(\underline{W})_{j_0}}
   {\underline{W}_{i}}, h(\underline{W}_{OV})_{i_0} \right\rangle 
   \annot{By \cref{def:f}}
   \\
   & = \sum_{j_0=1}^L \sum_{i=1}^d \dv{}{\underline{W}} c^2_{j_0,i_0} c(\underline{W})_{j_0,i_0} 
    \cdot
    \left\langle\alpha
    (\underline{W})_{j_0}^{-1} \cdot 
    \dv{u(\underline{W})_{j_0}}
    {\underline{W}_{i_0}} + 
    \dv{\alpha(\underline{W})_{j_0}^{-1}}
    {\underline{W}_{i_0}} \cdot u(\underline{W})_{j_0}, h(\underline{W}_{OV})_{i_0}\right\rangle 
   \nonumber\\
   & = \sum_{j_0=1}^L \sum_{i=1}^d \dv{}{\underline{W}} c^2_{j_0,i_0} c(\underline{W})_{j_0,i_0} \cdot
   \left\langle\alpha(\underline{W})_{j_0}^{-1} 
   \cdot 
    \dv
    {u(\underline{W})_{j_0}}{\underline{W}_{i_0}}-\alpha(\underline{W})_{j_0}^{-2} 
    \dv{ \alpha(\underline{W})_{j_0}}{\underline{W}_{i_0}} 
    \cdot u(\underline{W})_{j_0} , h(\underline{W}_{OV})_{i_0}\right\rangle.
    \annot{By chain rule}
   \end{align*}

   For each $j_0 \in[L]$, we have
   \begin{align*}
   \dv{\(\A_{j_0} \underline{W}\)} {\underline{W}_{i_0}}
   = \A_{j_0} \cdot \dv
   {\underline{W}}
   {\underline{W}_{i_0}} 
   = \left(\A_{j_0}\right) 
   [\cdot, i].
   \end{align*}
   Therefore, for each $j_0 \in[L]$, we have
    \begin{align*}
   \dv{u(\underline{W})_{j_0}}
   {\underline{W}_{i_0}}
   & =
   \dv{\exp \left(\A_{j_0} \underline{W}\right)}
   {\underline{W}_{i_0}}
   \annot{By \cref{def:u}}
   \\
   &= \exp \left(\A_{j_0} \underline{W}\right) \odot \dv{ \A_{j_0} \underline{W}}
   {\underline{W}_{i_0}}
   \annot{By entry-wise product rule}
   \\
   &=\A_{j_0}[\cdot, i] \odot u(\underline{W})_{j_0} .
   \annot{By \cref{def:u} again}
   \end{align*}
   Similarly, 
   \begin{align*}   
   \dv{ \alpha(\underline{W})_{j_0}}{\underline{W}_{i_0}}
   = & ~
   \dv{\left \langle u(\underline{W})_{j_0}, \one_L \right \rangle}
   {\underline{W}_{i_0}} 
   \annot{By \cref{def:alpha}}
   \\
   = & ~ \left\langle \A_{j_0}[\cdot, i] \odot \annot{By entry-wise product rule}u(\underline{W})_{j_0}, \one_L\right \rangle 
   \\
   = & ~ \left \langle \A_{j_0}[\cdot, i], u(\underline{W})_{j_0} \right \rangle .
   \annot{By \cref{def:u} again}
    \end{align*}

   Putting all together, we have
   \begin{align*}
   & ~ \dv{g_2(\underline{W})_{j_0, i_0}}
   {\underline{W}_{i_0}} \\
   = & ~ \[\left\langle h(\underline{W}_{OV})_{i_0}, \A_{j_0}[\cdot, i] \odot f(\underline{W})_{j_0} \right \rangle - \left \langle h(\underline{W}_{OV})_{i_0}, f(\underline{W})_{j_0} \right \rangle \cdot \left \langle \A_{j_0} [\cdot, i], f(\underline{W})_{j_0} \right \rangle\] \cdot c(\underline{W})_{j_0,i_0},
   \end{align*}
   where
   \begin{align*}
   & \left \langle h(\underline{W}_{OV})_{i_0}, \A_{j_0} [\cdot, i] \odot 
   f(\underline{W})_{j_0} \right \rangle - 
   \left \langle h(\underline{W}_{OV})_{i_0}, f(\underline{W})_{j_0} 
   \right \rangle \cdot \left \langle 
   \A_{j_0} [\cdot, i], f(\underline{W})_{j_0} 
   \right \rangle \\
   = & ~ \A_{j_0}^{\top} \left (\operatorname{\diag} \left (f({\underline{W}})_{j_0}\right)-f({\underline{W}})_{j_0} f({\underline{W}})_{j_0}^{\top}\right) h(\underline{W}_{OV})_{i_0}.
   \end{align*}
   This completes the proof.
   \end{proof}
Observe \eqref{eqn:grad_low_rank}  carefully.
We see that (I) is diagonal and (II) is low-rank.
This provides a hint for algorithmic speedup through low-rank approximation: 
If we approximate the other parts with low-rank approximation and carefully match the multiplication dimensions, 
we might formulate the computation of $\dv{g_2}{\underline{W}}$ as a chained low-rank approximation.

    Surprisingly, such an approach makes computing \eqref{eqn:grad_low_rank} as fast as in almost-linear time.
    To proceed, we further decompose \eqref{eqn:grad_low_rank}  according to the chain-rule in the next lemma, and then conduct the approximation term-by-term.

To facilitate our proof, it's convenient to introduce the following notations.
\begin{definition}[$q(\cdot)$]\label{def:q}
Define $c(\underline{W}) \in \mathbb{R}^{L \times d}$ as specified in \cref{def:c} and $h(\underline{W}_{OV}) \in \mathbb{R}^{L \times d}$ as described in \cref{def:h}.
Define $q(\underline{W}) \in \mathbb{R}^{L \times L}$ by
\begin{align*}
    q(\underline{W}) : = \underbrace{ c(\underline{W}) }_{L \times d} \underbrace{ h(\underline{W}_{OV})^\top }_{d \times L}.
\end{align*}
In addition, $q(\underline{W})_{j_0}^\top$ denotes the $j_0$-th row of $q(\underline{W})$, transposed, making it an $L \times 1$ vector.
\end{definition}

\begin{definition}[$p(\cdot)$,$p_1(\cdot)$, $p_2(\cdot)$]\label{def:p}
For each index $j_0 \in [L]$, we define $p(\underline{W})_{j_0} \in \mathbb{R}^n$ as follows:
\begin{align*}
p(\underline{W})_{j_0} := \left( \diag(f(\underline{W})_{j_0}) - f(\underline{W})_{j_0} f(\underline{W})_{j_0}^\top \right) q(\underline{W})_{j_0}.
\end{align*}
We define $p(\underline{W}) \in \mathbb{R}^{L \times L}$ such that $p(\underline{W})_{j_0}^\top$ forms the $j_0$-th row of $p(\underline{W})$.
In addition,
for every index $j_0\in [L]$, we define $p_1(\underline{W})_{j_0},p_2(\underline{W})_{j_0}\in \R^L$ as
\begin{align*}
p_1(\underline{W})_{j_0}\coloneqq \diag\left(f\left(\underline{W}\right)_{j_0}\right) q(\underline{W})_{j_0},\quad
p_2(\underline{W})_{j_0}\coloneqq f\left(\underline{W}\right)_{j_0} f\left(\underline{W}\right)_{j_0}^{\top} q(\underline{W})_{j_0},
\end{align*}
such that 
$p(\underline{W})=p_1(\underline{W})-p_2(\underline{W})$.
\end{definition}

$p(\cdot)$ allows us to express $\dv{g_2}{\underline{W}}$ in a neat form:
\begin{lemma}[]\label{lem:compute_gradient}
Define the functions $f(\underline{W}) \in \mathbb{R}^{L \times L}$, $c(\underline{W}) \in \mathbb{R}^{d \times L}$, $h(\underline{W}_{OV}) \in \mathbb{R}^{d \times L}$, $q(\underline{W}) \in \mathbb{R}^{L \times L}$, and $p(\underline{W}) \in \mathbb{R}^{L \times L}$ as specified in \cref{def:f,def:c,def:h,def:q,def:p}, respectively. 
Let $A_1, A_2 \in \mathbb{R}^{d \times L}$ be two given matrices, and define $\A = A_1^\top \otimes A_2^\top$. 
Define $g_2$ according to \hyperref[item:O1]{(O1)}, and let $g_2(\underline{W})_{j_0,i_0}$ be as described in \eqref{eqn:simplified_dg2_dw}.
It holds 
\begin{align}\label{eqn:A_1pA_2}
    \dv{g_2}{\underline{W}} = \vect\(A_1 p(\underline{W})A_2^\top\).
\end{align}
\end{lemma}

\begin{proof}

By definitions, \eqref{eqn:simplified_dg2_dw} gives
\begin{align}\label{eq:lxy_j0_i0}
    &~ \frac{\d (g_2)_{j_0,i_0}}{\d \underline{W}_{i_0}} 
    \\
    = &~  c_{j_0,i_0} \cdot (
    \underbrace{
    \langle  f(\underline{W})_{j_0} \odot \A_{j_0,i_0}, h(\underline{W}_{OV})_{i_0} \rangle
    }_{=\A_{j_0,i}^\top \diag(f(\underline{W})_{j_0}) h(\underline{W}_{OV})_{i_0}} 
    - \underbrace{
    \langle  f(\underline{W})_{j_0} , h(\underline{W}_{OV})_{i_0} \rangle \cdot \langle f(\underline{W})_{j_0}, \A_{j_0,i_0} \rangle)
    }_{=\A_{j_0,i}^\top f(\underline{W})_{j_0} f(\underline{W})_{j_0}^\top h(\underline{W}_{OV})_{i_0}}.
    \annot{By $\Braket{a\odot b,c}=a^\top\diag(b)c$ for $a,b,c\in\R^{L}$}
\end{align}

Therefore, \eqref{eq:lxy_j0_i0} becomes
\begin{align}\label{eq:rewrite_single_loss_gradient}
    \frac{\d (g_2)_{j_0,i_0}}{\d \underline{W}_{i_0}} 
    = & ~ c_{j_0,i_0} \cdot (\A_{j_0,i}^\top \diag(f(\underline{W})_{j_0}) h(\underline{W}_{OV})_{i_0} - \A_{j_0,i}^\top f(\underline{W})_{j_0} f(\underline{W})_{j_0}^\top h(\underline{W}_{OV})_{i_0}) \notag \\
    = & ~ c_{j_0,i_0} \cdot \A_{j_0,i}^\top ( \diag(f(\underline{W})_{j_0}) - f(\underline{W})_{j_0} f(\underline{W})_{j_0}^\top)h(\underline{W}_{OV})_{i_0}.
\end{align}
Then, by definitions of $q(\cdot),p(\cdot)$, we complete the proof.
\end{proof}

\subsubsection{Low-Rank Approximations of Building Blocks Part I:  \texorpdfstring{$f(\cdot),q(\cdot)$, and $c(\cdot)$}{}}
\label{sec:proof_building_block_I}

The definitions of $p$, $p_1$, $p_2$, and \cref{lem:compute_gradient} show that the DiT training  gradient $\dv{g_2}{\underline{W}}$ involves entry-wise products of $f$, $q$, and $c$. 
Therefore, if we approximate these with inner-dimension-matched low-rank approximations, computing $\dv{g_2}{\underline{W}}$ itself becomes a low-rank approximation. 
In the following sections, we present low-rank approximations for $f$, $q$, and $c$.

\begin{lemma}[Approximate $f(\cdot)$, Modified from  \cite{alman2024fast}]
\label{lemma:approx_f}
    Let $\Gamma = o(\sqrt{\log L})$ and $k_1=L^{o(1)}$.
    Let $A_1, A_2, \in \mathbb{R}^{d \times L}$, 
     $W\in\R^{d\times d}$ and
    $f(\underline{W})=D^{-1} \exp(A_1^\top  \bX A_2)$ with $D=\diag\left(\exp \left(A_1^\top W A_2\right) {\one_L}\right)$ follows 
    \cref{def:u,def:alpha,def:c,def:f}. 
    If $\max\big(\norm{A_1^{\top} W}_{\max} \leq \Gamma$,$\norm{A_2}_{\max} \big)\leq \Gamma$, then there exist two matrices $U_1, V_1 \in \mathbb{R}^{L \times k_1}$ such that $\norm{U_1 V_1^{\top} - f(\underline{W})}_{\max} \leq \epsilon / \poly (L)$.  
    In addition, it takes $L^{1+o(1)}$ time to construct $U_1$ and $V_1$.
\end{lemma}
\begin{proof}
By \cite[Theorem~3]{alman2024fast}, we complete the proof.
\end{proof}

\begin{lemma}[Approximate $c(\cdot)$]
\label{lemma:approx_c}
    Assume all numerical values are in $O(\log L)$ bits. 
    Let $d=O(\log L)$ and $c(\underline{W})\in\R^{L\times d}$ follows \cref{def:c}.
    There exist two matrices $U_1, V_1 \in \mathbb{R}^{L \times k_1}$ such that $\left\| U_1 V_1^{\top}h(W_{OV}) -Y^\top-c(\underline{W})\right\|_{\max} \leq \epsilon / \poly(L)$.
\end{lemma}

\begin{proof}[Proof of \cref{lemma:approx_c}]

    \begin{align*}
            \left\|U_1 V_1^{\top} h(W_{OV})-Y^\top-c(\underline{W})\right\|_{\max}
            & = 
            \left\|U_1 V_1^{\top} h(W_{OV}) -Y^\top-(f(\underline{W}) h(W_{OV}) - Y^\top)\right\|_{\max} 
            \annot{By \cref{def:c}}
            \\
            & = \left\|\[U_1 V_1^{\top}-f(\underline{W})\] h(W_{OV})\right\|_{\max} 
            \\
            & \leq \epsilon / \poly(L).
            \annot{By \cite[Theorem~3]{alman2024fast}}
    \end{align*}
\end{proof}

\begin{lemma}[Approximate $q(\cdot)$]
\label{lemma:approx_q}
    Let $k_2=L^{o(1)}$, $c(\cdot)\in\R^{L\times d}$ follow \cref{def:c} and 
    let $q(\underline{W})
    \coloneqq c(\underline{W})h(\underline{W}_{OV})^\sT \in \R^{L\times L}$ (follow \cref{def:q}).
    There exist two matrices $U_2, V_2 \in \mathbb{R}^{L \times k_2}$ such that $\left\|U_2 V_2^{\top}-q(\underline{W})\right\|_{\max} \leq \epsilon /\poly (L)$.
    In addition, it takes $L^{1+o(1)}$ time to construct $U_2, V_2$.
\end{lemma}
\begin{proof}[Proof of \cref{lemma:approx_q}]
Our proof is built on \cite[Lemma~D.3]{alman2024fast}.

    Let $\tilde{q}(\cdot)$ denote an approximation to $q(\cdot)$.  
    
    By \cref{lemma:approx_c}, $U_1 V_1^{\top} h(W_{OV})-Y$ approximates $c(\underline{W})$ up to accuracy $\epsilon=1/\poly(L)$.

    Thus, by setting $\tilde{q}(\underline{W}) = h(W_{OV})
    \left(U_1 V_1^{\top} h(W_{OV})-Y\right)^{\top}$, we find a low-rank form for $\tilde{q}(\cdot)$:
    \begin{align*}
        \tilde{q}(\underline{W})
        = h(W_{OV})
        \left(h(W_{OV})\right)^{\top} V_1 U_1^{\top}-h(W_{OV}) Y^{\top},
    \end{align*}
    such that
    \begin{align*}
        \|\tilde{q}(\underline{W}) -q(\underline{W})\|_{\max}
        & = \left\|h(W_{OV})\left(U_1 V_1^{\top} h(W_{OV})-Y\right)^{\top} - h(W_{OV}) Y^{\top}\right\|_{\max} 
        \\
        & \leq d\left\|h(W_{OV})\right\|_{\max}
        \left\|U_1 V_1^{\top} h(W_{OV}) - Y - c(\underline{W})\right\|_{\max} \\
        & \leq \epsilon / \poly (L).
    \end{align*}
    
    By $k_1,d=L^{o(1)}$, compute $\underbrace{\left(h(W_{OV})\right)^{\top}}_{{d\times L}} \underbrace{V_1}_{L\times k_1} \underbrace{U_1^{\top}}_{k_1\times L} $   takes only $L^{1+o(1)}$ time.
    This completes the proof.
\end{proof}

\subsubsection{Low-Rank Approximations of Building Blocks Part II: \texorpdfstring{$p(\cdot)$}{}}
\label{sec:proof_building_block_II}

Now, we use the low-rank approximations of $f,q,c$ to construct low-rank approximations for $p_1(\cdot),p_2(\cdot), p(\cdot)$.
\begin{lemma}[Approximate $p_1(\cdot)$]
\label{lemma:approx_p1}
    Let $k_1,k_2=L^{o(1)}$.  
    Suppose $U_1, V_1 \in \mathbb{R}^{L \times k_1}$ approximates  $f(\underline{W})\in\R^{L\times L}$ such that $\left\|U_1 V_1^{\top}-f(\underline{W})\right\|_{\max} \leq \epsilon /\poly(L)$, and
    $U_2, V_2 \in \mathbb{R}^{L \times k_2}$ approximates the $q(\underline{W}) \in \mathbb{R}^{L \times L}$ such that $\left\|U_2 V_2^{\top}-q(\underline{W})\right\|_{\max} \leq \epsilon /\poly(L)$. 
    Then there exist two matrices $U_3, V_3 \in \mathbb{R}^{L \times k_3}$ such that $\left\|U_3 V_3^{\top}-p_1(\underline{W})\right\|_{\max} \leq$ $\epsilon /\poly(L)$. 
    In addition, it takes $L^{1+o(1)}$ time to construct $U_3, V_3$.
    
\end{lemma}

\begin{proof}[Proof of \cref{lemma:approx_p1}]
    By tensor trick, 
    we construct $U_3$, $V_3$ as tensor products of $U_1,V_1$ and $U_2,V_2$, respectively, while preserving their low-rank structures.
    Then, we show the low-rank approximation of $p_1(\cdot)$ with bounded error by \cref{lemma:approx_f} and \cref{lemma:approx_q}. 
    
    Let $\oslash$ be {\textit{column-wise} Kronecker product} such that $A \oslash B \coloneqq [A[\cdot,1] \otimes B [\cdot,1] \mid \ldots \mid A [\cdot,k_1] \otimes B [\cdot,k_1]] \in \R^{L \times k_1k_2}$ for $A \in \R^{L \times k_1},B \in \R^{L \times k_2}$. 
    
    Let  $\tilde{f}(\underline{W}) \coloneqq U_1 V_1^\sT$ and $\tilde{q}(\underline{W}) \coloneqq U_2V_2^\sT$  denote matrix-multiplication approximations to  $f(\underline{W})$ and $q(\underline{W})$, respectively.

    For the case of presentation, let $U_3 = \overbrace{U_1}^{L\times k_1} \oslash \overbrace{U_2}^{L\times k_2}$ and 
    $V_3 = \overbrace{V_1}^{L \times k_1} \oslash \overbrace{V_2}^{L \times k_2}$. 
    It holds
    \begin{align*}
        & ~ \left\|U_3 V_3^{\top}-p_1(\underline{W})\right\|_{\max}
        \nonumber\\
        = & ~\left\|U_3 V_3^{\top}-f(\underline{W}) \odot q(\underline{W})\right\|_{\max}
        \annot{
        By $p_1(\underline{W})= f(\underline{W}) \odot q(\underline{W})$}
        \\
        = & ~\left\|\left(U_1 \oslash U_2\right)\left(V_1 \oslash V_2\right)^{\top} - f(\underline{W}) \odot q(\underline{W}) \right\|_{\max} \\
        = & ~
        \left\|\left(U_1 V_1^{\top}\right) \odot \left(U_2 V_2^{\top}\right)-f(\underline{W}) \odot q(\underline{W})\right\|_{\max}
        \\
        = & ~
        \|\tilde{f}(\underline{W}) \odot \tilde{q} (\underline{W})-f(\underline{W}) \odot q(\underline{W})\|_{\max} 
        \\
        \leq & ~
        \underbrace{\|\tilde{f}(\underline{W}) \odot \tilde{q}(\underline{W})-\tilde{f}(\underline{W}) \odot q(\underline{W})\|_{\max}
        }_{\le \epsilon/\poly(L)}
        +
        \underbrace{\|\tilde{f}(\underline{W}) \odot q(\underline{W})-f(\underline{W}) \odot q(\underline{W})\|_{\max} 
        }_{\le \epsilon/\poly(L)}
        \\
        \leq &~ \epsilon / \poly (L).
        \annot{By \cref{lemma:approx_f} and \cref{lemma:approx_q}}
    \end{align*}
    Computationally, by $k_1,k_2=L^{o(1)}$, computing $U_3$ and $V_3$ takes $L^{1+o(1)}$ time.
    This completes the proof.
\end{proof}

\begin{lemma}[Approximate $p_2(\cdot)$]
\label{lemma:approx_p2}
    Let $k_1,k_2,k_4=L^{o(1)}$. 
    Let $p_2(\underline{W})\in\R^{L\times L}$ follow \cref{def:p} such that its $j_0$-th column is $p_2(\underline{W})_{j_0}=f(\underline{W})_{j_0}f(\underline{W})_{j_0}^{\top} q(\underline{W})_{j_0}$ for each $j_0 \in [L]$. 
    Suppose $U_1, V_1 \in \mathbb{R}^{L \times k_1}$ approximates the $\mathrm{f}(\mathrm{\bX})$ such that $\left\|U_1 V_1^{\top}-f(\underline{W})\right\|_{\max} \leq \epsilon /\poly (L)$, and 
     $U_2, V_2 \in \mathbb{R}^{L \times k_2}$ approximates the $q(\underline{W}) \in \mathbb{R}^{L \times L}$ such that $\left\|U_2 V_2^{\top}-q(\underline{W})\right\|_{\max} \leq \epsilon /\poly (L)$. 
     Then there exist matrices $U_4, V_4 \in \mathbb{R}^{L \times k_4}$ such that $\left\|U_4 V_4^{\top}-p_2(\underline{})\right\|_{\max} \leq \epsilon /\poly(L)$. 
     In addition, it takes $L^{1+o(1)}$ time to construct $U_4, V_4$.
\end{lemma}
\begin{proof}[Proof of \cref{lemma:approx_p2}]
    From \cref{def:p},
    \begin{align*}
    p_2(\underline{W})_{j_0}
    \coloneqq \overbrace{f\left(\underline{W}\right)_{j_0}  \underbrace{f\left(\underline{W}\right)_{j_0}^{\top} q(\underline{W})_{j_0}}_{(I)}
    }^{(II)}.
    \end{align*}
    For (I),
    we show its low-rank approximation by observing the low-rank-preserving property of the multiplication between $f(\cdot)$ and  $q(\cdot)$ (from \cref{lemma:approx_f} and \cref{lemma:approx_q}). 
    For (II), 
    we show its low-rank approximation by the low-rank structure of  $f(\cdot)$ and (I).

    \paragraph{Part (I).}
    We define a function $r(\underline{W}): \R^{d^2} \to \R^L$ such that the $j_0$-th component $r(\underline{W})_{j_0} \coloneqq
    \left(f(\underline{W})_{j_0}\right)^{\top} q(\underline{W})_{j_0}$ for all $j_0\in[L]$. 
    Let $\tilde{r}(\underline{W})$ denote the approximation of $r(\underline{W})$ via decomposing  into $f(\cdot)$ and $q(\cdot)$:
\begin{align}
            \tilde{r}(\underline{W})_{j_0}
             & \coloneqq \left \langle \tilde{f}(\underline{W})_{j_0}, \tilde{q}(\underline{W})_{j_0}\right\rangle
             = \left(U_1 V_1^{\top}\right)[j_0, \cdot] \cdot\left[\left(U_2 V_2^{\top}\right)[j_0, \cdot]\right]^{\top} 
             \nonumber
             \\
             & = U_1[j_0, \cdot] \underbrace{V_1^{\top} }_{{k_1\times L}} \underbrace{V_2}_{{L\times k_2}}\left(U_2[j_0, \cdot]\right)^{\top},
             \label{eqn:U1V1V2U1}
\end{align}
for all $j_0\in[L]$. 
This allows us to write ${p}_2(\underline{W}) ={f}(\underline{W}) \diag({r}(\underline{W}))$ with
    $\diag(\tilde{r}(\underline{W}))$ denoting a diagonal matrix with diagonal entries being components of $\tilde{r}(\underline{W})$.

    \paragraph{Part (II).}
    With $r(\cdot)$, we approximate $p_2(\cdot)$ with $\tilde{p}_2(\underline{W}) = \tilde{f}(\underline{W}) \diag (\tilde{r}(\underline{W}))$ as follows.

    Since $\tilde{f}(\underline{W})$ has low rank representation, and $\diag(\tilde{r}(\underline{W}))$ is a diagonal matrix, 
    $\tilde{p}_2(\cdot)$ has low-rank representation by definition.
    Thus, we set $\tilde{p}_2(\underline{W}) = U_4V_4^\sT$ with $U_4 = U_1$ and $V_4 = \diag(\tilde{r}(\underline{W})) V_1$.
    Then, we bound the approximation error
\begin{align*}
            & ~ \left\|U_4 V_4^{\top}-p_2(\underline{W})\right\|_{\max} \\
            = & ~  \left\|\tilde{p}_2(\underline{W})-p_2(\underline{W})\right\|_{\max}
            \\
            = & ~
            \max _{j_0 \in[L]}\left\|
            {
            \tilde{f}(\underline{W})_{j_0} \tilde{r}(\underline{W})_{j_0}-f(\underline{W})_{j_0} r(\underline{W})_{j_0}
            }
            \right\|_{\max} 
            \\
            \leq & ~ \max _{j_0 \in[L]}\[\left\|\tilde{f}(\underline{W})_{j_0} \tilde{r}(\underline{W})_{j_0}-f(\underline{W})_{j_0} {r}(\underline{W})_{j_0}\right\|_{\max}+\left\|\tilde{f}(\underline{W})_{j_0} \tilde{r}(\underline{W})_{j_0}-f(\underline{W})_{j_0} r(\underline{W})_{j_0}\right\|_{\max} \]
            \annot{By triangle inequality}
            \\
            \leq & ~ \epsilon / \poly(L).  
\end{align*}
Computationally, computing  $V_1^{\top} V_2$ takes $L^{1+o(1)}$ time by $k_1,k_2=L^{o(1)}$.
    Once we have $V_1^{\top} V_2$ precomputed, \eqref{eqn:U1V1V2U1}
    only takes $O(k_1 k_2)$ time for each $j_0\in[L]$.
    Thus, the total time is 
    $O\left(L k_1 k_2\right)=L^{1+o(1)} $.
    Since $U_1$ and $V_1$  takes $L^{1+o(1)}$ time to construct and $V_4 = \underbrace{\diag(\tilde{r}(\underline{W}))}_{L\times L} \underbrace{V_1}_{L\times k_1}$ also takes $L^{1+o(1)}$ time, $U_4$ and $V_4$ takes $L^{1+o(1)}$ time to construct.
This completes the proof.
\end{proof}

\subsection{Proof of \texorpdfstring{\cref{thm:main_comp_back}}{}}
\label{proof:thm:main_comp_back}

\begin{proof}[Proof of \cref{thm:main_comp_back}]
    By the definitions of matrices $p(\cdot)$, $p_1(\cdot)$  and $p_2(\cdot)$ (\cref{def:p}),
    we have
    \begin{align*}
        p(\underline{W})=p_1(\underline{W})-p_2(\underline{W}).
    \end{align*}

    By \cref{lem:compute_gradient}, we have 
    \bea\label{eqn:lora_grad_proof}
             \dv{g_2}{\underline{W}} = \vect\(A_1 p(\underline{W})A_2^\top\)
             .
    \eea
    To show the existence of $L^{1+o(1)}$ algorithms for DiT backward computation \cref{prob:ADiTGC},  
    we prove fast low-rank approximations for 
    $A_1 p_1(\underline{W})A_2^\top$ and $A_1 p_2(\underline{W})A_2^\top$ as follows.

    Let $\tilde{p}_1(\underline{W}), \tilde{p_2}(\underline{W})$ denote the approximations to $p_1(\underline{W}), p_2(\underline{W})$, respectively.

    By \cref{lemma:approx_p1}, it takes $L^{1+o(1)}$ time to construct $U_3, V_3 \in \R^{L \times k_3}$ such that
    \begin{align*}
    A_1 \tilde{p}_1(\underline{W})A_2^\top=A_1 U_3V_3^\top A_2^\top. 
    \end{align*}
    Then,
    computing $\underbrace{A_1}_{d\times L} \underbrace{U_3}_{L\times k_3}\underbrace{V_3^\top }_{k_3\times L}\underbrace{A_2^\top}_{L\times d}$ takes $L^{1+o(1)}$ due to the fact that $d,k_1k_3=L^{o(1)}$.
    
    Therefore, total running time for $A_1 p_1(\underline{W})A_2^\top$ is $L\cdot L^{o(1)}= L^{1+o(1)}$.

    For the same reason (by \cref{lemma:approx_p2}), total running time for $A_1 p_2(\underline{W})A_2^\top$ is $L\cdot L^{o(1)}= L^{1+o(1)}$.

   Lastly, we have
\begin{align*}
        &~\left\|\pdv{g_2}{\underline{W}} -\tilde{G}^{(W)}\right\|_{\max}
        \\
    = &~ \left\|\vect\left(A_1 \tilde{p}(\underline{W})A_2^\top\right) -\vect\left(A_1 \tilde{p}(\underline{W})A_2^\top\right)\right\|_{\max}
        \annot{By \cref{lem:compute_gradient}}
    \\
    =&~  \left\|\left(A_1 \tilde{p}(\underline{W})A_2^\top\right) -\left(A_1 \tilde{p}(\underline{W})A_2^\top\right)\right\|_{\max} 
    \annot{By definition, $\norm{A}_{\max} \coloneqq \max_{i,j} \abs{A_{ij}}$ for any matrix $A$} 
    \\
    \leq &~ 
    \left\|\left(A_1 \[p_1(\underline{W})-\tilde{p}_1(\underline{W})\] A_2^\top\right)  \right\|_{\max}
    + \left\|\left(A_1 \[p_2(\underline{W})-\tilde{p}_2(\underline{W})\] A_2^\top\right)\right\|_{\max} 
    \annot{By \cref{def:p} and triangle inequality}
    \\
    \leq&~  
    \norm{A_1}_\infty\norm{A_2}_\infty
    \left(\left\|\left(p_1(\underline{W})-\tilde{p}_1(\underline{W})\right)\right\|_{\max} + 
    \left\|\left(p_2(\underline{W})-\tilde{p}_2(\underline{W})\right)\right\|_{\max}\right)
    \annot{By the sub-multiplicative property of $\norm{\cdot}_\infty$}
    \\
    \leq &~ 
    \epsilon / \poly(L).
    \annot{By \cref{lemma:approx_p1} and \cref{lemma:approx_p2}}
\end{align*}
Set $\epsilon = 1/ \poly(L)$.
We complete the proof.
\end{proof}

\clearpage

\clearpage
\newpage
\section*{NeurIPS Paper Checklist}

\begin{enumerate}

\item {\bf Claims}
    \item[] Question: Do the main claims made in the abstract and introduction accurately reflect the paper's contributions and scope?
    \item[] Answer: \answerYes{} %
    \item[] Justification: Our contributions and scope in \cref{sec:method} and \cref{sec:comp} are reflected by the claims in abstract and introduction.

\item {\bf Limitations}
    \item[] Question: Does the paper discuss the limitations of the work performed by the authors?
    \item[] Answer: \answerYes{} %
    \item[] Justification: We discuss the limitations in \cref{sec:conclusion}.

\item {\bf Theory Assumptions and Proofs}
    \item[] Question: For each theoretical result, does the paper provide the full set of assumptions and a complete (and correct) proof?
    \item[] Answer: \answerYes{}{} %
    \item[] Justification: Yes. We include our proofs in the appendix and have made every effort to ensure the correctness of our theoretical results.


    \item {\bf Experimental Result Reproducibility}
    \item[] Question: Does the paper fully disclose all the information needed to reproduce the main experimental results of the paper to the extent that it affects the main claims and/or conclusions of the paper (regardless of whether the code and data are provided or not)?
    \item[] Answer: \answerNA{} %
    \item[] Justification: This is a formal theory work without experiments.

\item {\bf Open access to data and code}
    \item[] Question: Does the paper provide open access to the data and code, with sufficient instructions to faithfully reproduce the main experimental results, as described in supplemental material?
    \item[] Answer: \answerNA{} %
    \item[] Justification: This is a formal theory work without experiments.

\item {\bf Experimental Setting/Details}
    \item[] Question: Does the paper specify all the training and test details (e.g., data splits, hyperparameters, how they were chosen, type of optimizer, etc.) necessary to understand the results?
    \item[] Answer: \answerNA{} %
    \item[] Justification: This is a formal theory work without experiments.

\item {\bf Experiment Statistical Significance}
    \item[] Question: Does the paper report error bars suitably and correctly defined or other appropriate information about the statistical significance of the experiments?
    \item[] Answer: \answerNA{} %
    \item[] Justification: This is a formal theory work without experiments.

\item {\bf Experiments Compute Resources}
    \item[] Question: For each experiment, does the paper provide sufficient information on the computer resources (type of compute workers, memory, time of execution) needed to reproduce the experiments?
    \item[] Answer: \answerNA{} %
    \item[] Justification: This is a formal theory work without experiments.
    
\item {\bf Code Of Ethics}
    \item[] Question: Does the research conducted in the paper conform, in every respect, with the NeurIPS Code of Ethics \url{https://neurips.cc/public/EthicsGuidelines}?
    \item[] Answer: \answerYes{} %
    \item[] Justification: Yes. We follow the code of ethics in this work.

\item {\bf Broader Impacts}
    \item[] Question: Does the paper discuss both potential positive societal impacts and negative societal impacts of the work performed?
    \item[] Answer: \answerYes{} %
    \item[] Justification: 
    This theoretical work aims to shed light on the foundations of diffusion generative models and is not
anticipated to have negative social impacts.
    
\item {\bf Safeguards}
    \item[] Question: Does the paper describe safeguards that have been put in place for responsible release of data or models that have a high risk for misuse (e.g., pretrained language models, image generators, or scraped datasets)?
    \item[] Answer: \answerNA{} %
    \item[] Justification: This is a formal theory work without experiments.

\item {\bf Licenses for existing assets}
    \item[] Question: Are the creators or original owners of assets (e.g., code, data, models), used in the paper, properly credited and are the license and terms of use explicitly mentioned and properly respected?
    \item[] Answer: \answerNA{} %
    \item[] Justification: This is a formal theory work without experiments.

\item {\bf New Assets}
    \item[] Question: Are new assets introduced in the paper well documented and is the documentation provided alongside the assets?
    \item[] Answer: \answerNA{} %
    \item[] Justification: This is a formal theory work without experiments.

\item {\bf Crowdsourcing and Research with Human Subjects}
    \item[] Question: For crowdsourcing experiments and research with human subjects, does the paper include the full text of instructions given to participants and screenshots, if applicable, as well as details about compensation (if any)? 
    \item[] Answer: \answerNA{} %
    \item[] Justification: This is a formal theory work without experiments.

\item {\bf Institutional Review Board (IRB) Approvals or Equivalent for Research with Human Subjects}
    \item[] Question: Does the paper describe potential risks incurred by study participants, whether such risks were disclosed to the subjects, and whether Institutional Review Board (IRB) approvals (or an equivalent approval/review based on the requirements of your country or institution) were obtained?
    \item[] Answer: \answerNA{} %
    \item[] Justification: This is a formal theory work without experiments.

\end{enumerate}

\end{document}